\documentclass{article}

\usepackage{url}

\usepackage{hyperref}
\usepackage{microtype}
\usepackage{graphicx}
\usepackage{booktabs} %

\let\originaladdcontentsline\addcontentsline
\usepackage[accepted]{icml2024}
\let\addcontentsline\originaladdcontentsline

\usepackage{inconsolata}
\usepackage[utf8]{inputenc}

\usepackage{color}
\usepackage{soul}
\usepackage{amsmath}
\usepackage{amsfonts}
\usepackage{pifont}%
\usepackage{xcolor}
\usepackage{longtable}
\usepackage{enumitem}
\usepackage{booktabs}
\usepackage{multicol}
\usepackage{multirow}
\usepackage{array}
\usepackage{float}
\usepackage{arydshln}
\usepackage{adjustbox}
\usepackage{multicol}
\usepackage[font=small]{caption}
\usepackage{subcaption}
\usepackage{makecell}
\usepackage{mdframed}
\usepackage{setspace}  %
\usepackage{svg}
\usepackage{fancyvrb,fvextra}
\usepackage[capitalize,noabbrev,nameinlink]{cleveref} %
\usepackage{supertabular}
\usepackage[textsize=tiny]{todonotes}
\usepackage{titlesec}
\usepackage{csquotes}
\usepackage{marvosym}
\usepackage{fontawesome5}

\newmdenv[
  linecolor=gray,       %
  linewidth=0.5pt,         %
  roundcorner=10pt,      %
  innertopmargin=10pt,   %
  innerbottommargin=10pt,%
  innerleftmargin=10pt,  %
  innerrightmargin=10pt, %
  backgroundcolor=gray!10!white!50, %
  font=\footnotesize            %
]{externaldoc}

\titlespacing{\paragraph}{0pt}{0.2\parskip}{\parskip}
\titlespacing{\subsection}{0pt}{0.4\parskip}{0.2\parskip}

\icmltitlerunning{\titlename}

\begin{document}

\newcommand{\Benchmark}{\textsc{WebLINX}}
\newcommand{\textttc}[1]{\scalebox{0.87}[1.0]{\ttfamily #1}}

\newcommand{\BenchmarkRepo}{\href{https://mcgill-nlp.github.io/weblinx}{https://mcgill-nlp.github.io/weblinx}}
\newcommand{\EsyCommerce}{EsyCommerce: \href{https://www.esycommerce.com/}{esycommerce.com}}

\newcommand{\ignore}[1]{}

\newcommand\Tstrut{\rule{0pt}{2.2ex}}
\newcommand\Bstrut{\rule[-0.6ex]{0pt}{0pt}}

\newcommand{\cmark}{{\color{green!70!black}\ding{51}}}%
\newcommand{\pmark}{{\color{yellow!70!black}$\dagger$}}%
\newcommand{\xmark}{{\color{red!90!black}\ding{55}}}%

\newcommand{\supeye}{\textsuperscript{\faEye}}

\newcommand{\testid}{$\textsc{Test}_\textsc{iid}$}
\newcommand{\testidcompact}{$\textsc{\textls[-75]{Test}}_\textsc{\textls[-60]{iid}}$}
\newcommand{\testod}{$\textsc{Test}_\textsc{ood}$}
\newcommand{\testodcompact}{$\textsc{\textls[-75]{Test}}_\textsc{\textls[-65]{ood}}$}
\newcommand{\testcat}{$\textsc{Test}_\textsc{cat}$}
\newcommand{\testgeo}{$\textsc{Test}_\textsc{geo}$}
\newcommand{\testweb}{$\textsc{Test}_\textsc{web}$}
\newcommand{\testvis}{$\textsc{Test}_\textsc{vis}$}

\newcommand{\Instructor}{$\mathcal{I}$}
\newcommand{\Navigator}{$\mathcal{N}$}
\newcommand{\titlename}{\Benchmark{}: Real-World Website Navigation with Multi-Turn Dialogue}

\newenvironment{compactitemize} {\begin{itemize}[itemsep=1.5pt,topsep=1.5pt,parsep=1.5pt,partopsep=1.5pt]}   {\end{itemize}}
\newenvironment{compactenumerate}{\begin{enumerate}[itemsep=1.5pt,topsep=1.5pt,parsep=1.5pt,partopsep=1.5pt]}   {\end{enumerate}}
\urlstyle{same}

\makeatletter
\def\adl@drawiv#1#2#3{%
  \hskip.5\tabcolsep
  \xleaders#3{#2.5\@tempdimb #1{1}#2.5\@tempdimb}%
  #2\z@ plus1fil minus1fil\relax
  \hskip.5\tabcolsep}
\newcommand{\cdashlinelr}[1]{%
  \noalign{\vskip\aboverulesep
    \global\let\@dashdrawstore\adl@draw
    \global\let\adl@draw\adl@drawiv}
  \cdashline{#1}
  \noalign{\global\let\adl@draw\@dashdrawstore
    \vskip\belowrulesep}}
\makeatother

\def\ZK#1{{\color{green!60!black!100}ZK: \it #1}}
\def\ZKdel#1{{\color{green!60!black!100} ZKdel: {\sout{#1}}}}
\def\XH#1{{\color{red}[XH: \textit{#1}]}}
\def\siva#1{{\color{blue!60!black!100}[Siva: \textit{#1}]}}

\def\txblue#1{{\color{blue}{#1}}}
\def\txred#1{{\color{red}{#1}}}

\renewcommand{\sectionautorefname}{Section}
\renewcommand{\subsectionautorefname}{Section}
\renewcommand{\subsubsectionautorefname}{Section}

\twocolumn[
  \icmltitle{\titlename}

  \icmlsetsymbol{equal}{*}

  \begin{icmlauthorlist}
    \icmlauthor{Xing Han Lù}{equal,mila,mcgill}
    \icmlauthor{Zdeněk Kasner}{equal,mila,charles}
    \icmlauthor{Siva Reddy}{mila,mcgill,cifar}
  \end{icmlauthorlist}

  \icmlaffiliation{mila}{Mila Quebec AI Institute}
  \icmlaffiliation{charles}{Institute of Formal and Applied Linguistics, Charles University}
  \icmlaffiliation{mcgill}{McGill University}
  \icmlaffiliation{cifar}{Facebook CIFAR AI Chair}

  \icmlcorrespondingauthor{Xing Han Lù}{xing.han.lu@mail.mcgill.ca}
  \icmlcorrespondingauthor{Zdeněk Kasner}{kasner@ufal.mff.cuni.cz}
  \icmlcorrespondingauthor{Siva Reddy}{siva.reddy@mila.quebec}

  \icmlkeywords{Machine Learning, ICML}

  \vskip 0.3in
]

\printAffiliationsOnly{\icmlEqualContribution}

\begin{abstract}
  We propose the problem of \textit{conversational web navigation}, where a digital agent controls a web browser and follows user instructions to solve real-world tasks in a multi-turn dialogue fashion.
  To support this problem, we introduce \Benchmark{} -- a large-scale benchmark of 100K interactions across 2300 expert demonstrations of conversational web navigation. Our benchmark covers a broad range of patterns on over 150 real-world websites and can be used to train and evaluate agents in diverse scenarios.
  Due to the magnitude of information present, Large Language Models (LLMs) cannot process entire web pages in real-time.
  To solve this bottleneck, we design a retrieval-inspired model that efficiently prunes HTML pages by ranking relevant elements. We use the selected elements, along with screenshots and action history, to assess a variety of models for their ability to replicate human behavior when navigating the web.
  Our experiments span from small text-only to proprietary multimodal LLMs.
  We find that smaller finetuned decoders surpass the best zero-shot LLMs (including GPT-4V), but also larger finetuned multimodal models which were explicitly pretrained on screenshots. However, all finetuned models struggle to generalize to unseen websites. Our findings highlight the need for large multimodal models that can generalize to novel settings.
  Our code, data and models are available for research: \BenchmarkRepo{}.
\end{abstract}

\begin{figure}[t!]
  \centering
  \includegraphics[width=\linewidth]{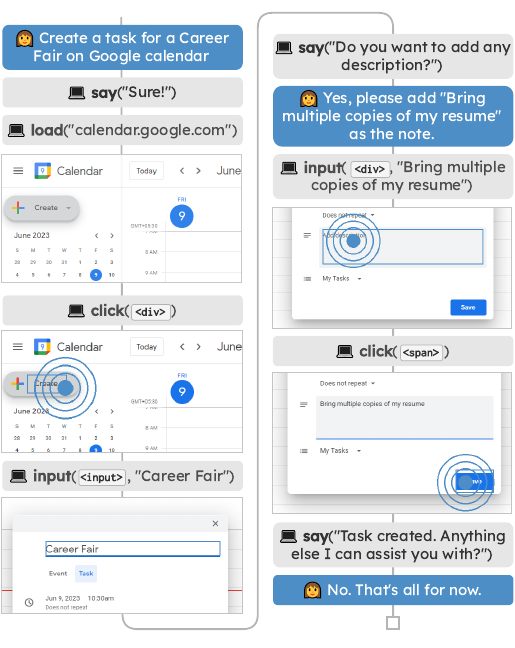}
  \caption{An example of the \textit{conversational web navigation} task. The instructor (\textbf{blue}) communicates with the navigator (\textbf{grey}) using only natural language. The latter controls the browser, having access to screenshots and textual website representation.}
  \label{fig:webnav}
\end{figure}

\begin{table*}[t]
  \centering
  \footnotesize
  \caption{\Benchmark{} is the first benchmark featuring real-world websites with multi-turn dialogue. The columns indicate: use of multi-turn dialogue (\textit{Chat}), if tasks are general or specialized (\textit{Gener.}), a web browser is used (\textit{Browse}), number of app/website domains (\textit{\# Dom.}), number of instances (\textit{\# Inst.}), average number of HTML elements per page (\textit{Avg. \# El.}), average number of turns per instance (\textit{Avg. \# Turns}).
    *AITW has 30K unique prompts with multiple demos each and the browsing data is strictly from Android devices.}
  \label{tab:data_stats}
  \vspace{-0.5em}
  \begin{tabular}{@{}lcccrrrrll@{}}
    \toprule
    \textbf{Benchmark}                                        & \textbf{Chat} & \textbf{Gener.} & \textbf{Browse} & \textbf{\# Dom.} & \textbf{\# Inst.} & \textbf{\makecell{Avg. \# El.}} & \textbf{\makecell{Avg. \# Turns}} & \textbf{Setting} \\
    \midrule
    MiniWob++ \cite{Liu_Guu_Pasupat_Shi_Liang_2018}           & \xmark        & \xmark          & \xmark          & 100              & 100               & 28                              & 3.6                               & Simplified       \\
    WebShop \cite{Yao_Chen_Yang_Narasimhan_2023}              & \xmark        & \xmark          & \cmark          & 1                & 12K               & 38                              & 11.3                              & E-Commerce       \\
    WebArena \cite{zhou2023webarena}                          & \xmark        & \cmark          & \cmark          & 6                & 812               & -                               & -                                 & Real-world       \\
    VWA \citep{koh2024visualwebarena}                         & \xmark        & \cmark          & \cmark          & 3                & 910               & -                               & -                                 & Real-world       \\

    Mind2Web \cite{deng2023mind2web}                          & \xmark        & \cmark          & \cmark          & 137              & 2350              & 1135                            & 7.3                               & Real-world       \\
    AITW$^*$ \citep{rawles2023android}                        & \xmark        & \cmark          & \cmark          & 357              & 30K               & -                               & 6.5                               & Android/Apps     \\
    WebVoyager \citep{he2024webvoyager}                       & \xmark        & \cmark          & \cmark          & 15               & 300               & -                               & -                                 & Real-world       \\
    RUSS   \cite{Xu_Masling_Du_Campagna_Heck_Landay_Lam_2021} & \cmark        & \xmark          & \cmark          & 22               & 80                & 801                             & 5.4                               & Help center      \\
    WorkArena \citep{drouin2024workarena}                     & \cmark        & \xmark          & \cmark          & 1                & 23K               & -                               & 10                                & IT Management       \\
    META-GUI \cite{Sun_Chen_Chen_Dai_Zhu_Yu_2022}             & \cmark        & \cmark          & \xmark          & 11               & 1125              & 79                              & 4.3                               & Mobile apps      \\
    \cdashlinelr{1-10}
    \Benchmark{}  (ours)                                      & \cmark        & \cmark          & \cmark          & 155              & 2337              & 1775                            & 43.0                              & Real-world       \\
    \bottomrule
\end{tabular}

\end{table*}

\section{Introduction}

Proprietary conversational assistants like ChatGPT \citep{OpenAI_2022_ChatGPT} are capable of more than just conversing; they can also browse websites through plugins \citep{chatgpt-plugins, bartextensions}, allowing them to perform actions and provide more useful responses. However, this capability is limited: the plugins must be developed separately for each website and may not cover all of a website's functionality. This limitation raises an important research question: can we leverage the models behind those assistants to navigate websites directly in the user's browser, while retaining their conversational capabilities?

Motivated by this question, we define the real-world problem of \textbf{conversational web navigation}: given the initial user instruction, an agent must complete a real-world task inside a web browser while communicating with the user via multi-turn dialogue. This problem is relevant in many real-world scenarios: helping visually impaired users efficiently navigate websites through a chat interface, enhancing smart speakers and digital assistants with voice-controlled web navigation, and improving the productivity of knowledge workers by reducing highly repetitive steps while staying in control. From a research perspective, this problem can be used to assess the ability of LLM agents to not only follow self-contained instructions, but also engage with their environment through dialogue and generalize to unforeseen situations.

To address this problem, we introduce \textbf{\Benchmark{}}\footnote{\textbf{Web}  \textbf{L}anguage \textbf{I}nterface for \textbf{N}avigation \& e\textbf{X}ecuting actions} (\S\ref{sec:benchmark}), a benchmark containing 2337 demonstrations of \textit{conversational web navigation} produced by human experts across 155 real-world websites. \Cref{fig:webnav} shows a demonstration. Each demonstration captures the full sequence of actions performed by a human \textit{navigator} when interacting with the user (known as \textit{instructor}) through a conversational interface. We record over 100K occurrences of actions and utterances, where each action is associated with a Document Object Model (DOM)\footnote{Tree representation of HTML page as rendered in the browser.} tree, browser screenshots, and frames from demonstration-level video recordings. \Cref{tab:data_stats} highlights the unique aspects of \Benchmark{}. Unlike previous works focused on mobile apps or specialized applications, ours is the first large-scale benchmark that can be used to train dialogue-enabled navigation agents and evaluate their generalization capabilities to realistic scenarios, such as adapting to new websites, categories, and geographies; we also reserve a split to assess the ability of agents to interact with instructors without visual access to the browser.

A naive way to use this benchmark would be to give the full DOM tree directly to an agent and instruct it to predict the correct action. As some HTML pages contain thousands of elements, fitting them completely within the context of a LLM poses a significant challenge; even if it was possible, existing LLMs would be unable to process them in real-time. Consequently, we design a method called \textit{Dense Markup Ranking} (\S\ref{sec:DMR}), which compares each element in an HTML page with the full action history. By using a similarity-based approach to both learn and rank elements, we can leverage compact architectures used in text retrieval. This lets us find the most relevant elements and prune irrelevant ones to obtain a compact representation of the DOM. We combine it with the action history, detailed instruction and screenshot (in a multimodal context) to construct an input representation for LLMs, which can now meaningfully predict which actions to take. However, even if a predicted action is correct, it may be identified as incorrect by existing metrics, which can happen when there are minor differences in an agent's response or when an overlapping element is selected. Thus, we design a suite of evaluation metrics (\S\ref{sec:evaluation}) tailored for specific types of action (for instance, \textit{clicking} should be evaluated differently from what the navigator \textit{says}).

We examine 19 models based on 8 architectures (\S\ref{sec:results}), including smaller image-to-text, larger text-only decoders, LLMs, and multimodal models (capable of accessing both image and text). Among them, 5 are in the zero-shot setting, and the remaining are finetuned using the training split of \Benchmark{}. We find that even the best zero-shot model, GPT-4V \citep{GPT4V_System_card}, is surpassed by finetuned
models (\S\ref{sec:overview_results}). Notably, a smaller model like Sheared-LLaMA \citep{xia2023sheared} outperforms the much larger Fuyu \citep{fuyu_blog_post}, which was pretrained with browser screenshots. However, all models face challenges in generalizing to new settings, such as unseen websites from a different geographic location or when the instructor gives instructions without seeing the screen. Those findings prompted us to qualitatively look at the behavior of the models (\S\ref{sec:qualitative_assessment}), where we find that GPT-4V lacks situational awareness and can make obvious blunders. However, the best finetuned models still fail in simple cases, such as clicking on non-existing links or failing to change the language of a translation app. Thus, we believe that significant effort will be needed to make progress on the problem of \textit{conversational web navigation}, as we discuss in \cref{sec:discussion}.

Our contributions are summarized as follows:

\begin{compactitemize}
  \item We introduce the problem of real-world \textbf{conversational web navigation} and a large-scale expert-annotated benchmark for it, named \Benchmark{} (\S\ref{sec:benchmark}).
  \item We propose a suite of action-specific metrics, which we combine to assess overall model performance (\S\ref{section:metrics}).
  \item We design a method to simplify HTML pages (\S\ref{sec:DMR}), allowing us to evaluate a wide range of models (\S\ref{sec:models_generating_actions}).
  \item We find that smaller text-only decoders outperform multimodal LLMs, but all finetuned models struggle to generalize to novel scenarios (\S\ref{sec:results}).
\end{compactitemize}

\begin{figure}[t]
  \centering
  \small
  \includegraphics[width=\linewidth]{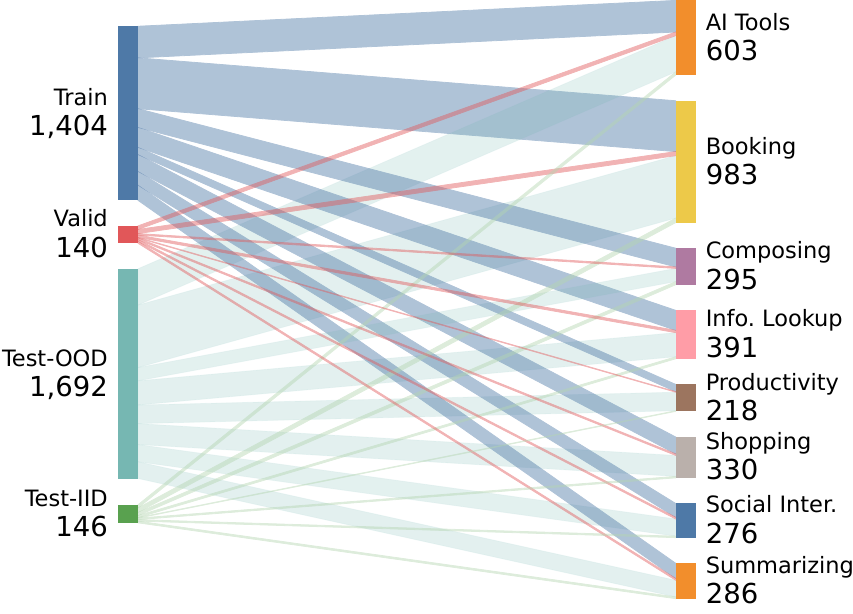}
  \caption{Distribution of demonstrations in \Benchmark{} across categories (\autoref{sec:categories_of_models}) and splits (\autoref{tab:demo_splits}). Each category has many subcategories as shown in \Cref{appendix:categories_and_subcategories}.}
  \label{fig:scatter_stats}
\end{figure}

\section{Related Work}
\label{section:related_work}

\subsection{Web Navigation Agents}
Previous work predominantly focused on building web agents for a single task. A prominent work for task-driven web navigation is MiniWoB++ \cite{Shi_Karpathy_Fan_Hernandez_Liang_2017,Liu_Guu_Pasupat_Shi_Liang_2018}, a simulated web environment with an extensive list of task primitives (e.g., select value from a dropdown or date from a calendar). Its well-defined input space and the flexibility of its simulated environments lead to reinforcement learning approaches reaching human-level performance \cite{Liu_Guu_Pasupat_Shi_Liang_2018,Humphreys_Raposo_Pohlen_Thornton_Chhaparia_Muldal_Abramson_Georgiev_Goldin_Santoro_2022}. However, the ability of those methods to transfer to realistic settings have been limited, even after introducing environment extensions \cite{Gur_Jaques_Miao_Choi_Tiwari_Lee_Faust_2022} and sample-efficient methods \cite{kim2023language}. Other works also explored grounding language commands to web elements and mobile UIs \cite{Pasupat_Jiang_Liu_Guu_Liang_2018,li2020mapping,Burns_Arsan_Agrawal_Kumar_Saenko_Plummer_2022}, or question answering (QA) by navigating Wikipedia \cite{Nogueira_Cho_2016}.

In an effort to build more realistic environments, \citet{Yao_Chen_Yang_Narasimhan_2023} introduced WebShop, an e-commerce environment with over 12K human-written task instructions. Models trained on WebShop achieved strong performance, but still relied on clean HTML and simple visual representations \cite{Furuta_Nachum_Lee_Matsuo_Gu_Gur_2023}. Instead, we aim to build agents that can act on \textit{any real-world website}, often existing in noisy and dynamic environments.

The prospect of using LLMs to act on real websites  \cite{Nakano_Hilton_Balaji_Wu_Ouyang_Kim_Hesse_Jain_Kosaraju_Saunders_2022} has lead to the development of LLM-based navigation services \citep{adept-act1,multi-on,hyperwrite-ai}, which has set the stage for academic counterparts.
\textsc{Mind2Web} \cite{deng2023mind2web}, WebArena \cite{zhou2023webarena} and VisualWebArena \citep{koh2024visualwebarena} are large-scale resources for building autonomous navigation agents like SeeAct \citep{zheng2024gpt4vision} and WebVoyager \citep{he2024webvoyager}.
On the other hand, \Benchmark{} is a benchmark for building agents that can interact with users in a multi-turn dialogue fashion, allowing them to be steered towards precise goals.
To this end, our problem formulation significantly expand and generalize upon exploratory work on simulated instructors for movie ticket booking \citep{Gur2019LearningCW}, semantic parsing-based agents for online help centers \citep{Xu_Masling_Du_Campagna_Heck_Landay_Lam_2021}, and \textit{iterative tool resolution} for crowd-source platforms \citep{xu2024turkingbench}.

\begin{table}[t]
  \centering
  \small
  \caption{Demonstration (Demo) splits for training and evaluation.}
  \label{tab:demo_splits}
  \begin{tabular}{l p{0.75\linewidth}}
    \toprule
    \textbf{Split} & \textbf{Description}                                    \\
    \midrule
    \textsc{Train} & Demos used to train models in \autoref{sec:experiments} \\
    \textsc{Valid} & In-domain demos for hyperparameters selection           \\
    \testid{}      & In-domain demos to test in-domain generalization        \\
    \midrule
    \testod{}      & \textit{Aggregation of splits for OOD evaluation}       \\
    \cdashlinelr{1-2}
    \testweb{}     & Unseen websites from the same subcategories             \\
    \testcat{}     & New subcategories within the same categories            \\
    \testgeo{}     & Geographic locations not in \textsc{Train}              \\
    \testvis{}     & Instructor does not see the screen                      \\
    \bottomrule
  \end{tabular}
\end{table}

\subsection{Website Representations}

Efficiently representing real-world websites is a long-standing challenge in web understanding \cite{Wu_Wang_Shen_Peng_Nichols_Bigham_2023}, including subtasks like web information extraction \cite{chang2006survey} and web segmentation \cite{kiesel2020web}. The approaches for simplifying or compressing the \textit{textual} representation of the website -- its HTML code or DOM tree
-- include rule-based algorithms \cite{Zhou_Sheng_Vo_Edmonds_Tata_2021}, accessibility-tree representations offered by browsers \cite{assouel2023the}, graph embeddings \cite{Wang_Fang_Ravula_Feng_Quan_Liu_2022}, and model-based approaches \cite{Deng_Shiralkar_Lockard_Huang_Sun_2022,Li_Xu_Cui_Wei_2022,Aghajanyan_Okhonko_Lewis_Joshi_Xu_Ghosh_Zettlemoyer_2021,gur-2023-real-world-webagent}.
Previous works for representing the \textit{visual} information of the webpage usually rely on feature extraction \cite{liu2009vide,8287714}, closely following the research on graphical UIs \cite{wu2021screen,bunian2021vins}.
Inspired by \citet{deng2023mind2web}, we propose a novel dense markup ranker which  selects relevant DOM elements, and use these elements optionally combined high-resolution browser screenshots.

\subsection{Conversational Interfaces}
Using conversational interfaces to complete tasks is the basis of task-oriented dialogue \cite{chen2017survey,zhang2020recent}. End-to-end solutions have shown promising results \cite{zhang2020dialogpt,kann-etal-2022-open}, but the use of LLMs remains under scrutiny \cite{hudevcek2023llms}. For real-world services, Dialog2API \cite{Shu_Mansimov_Alkhouli_Pappas_Romeo_Gupta_Mansour_Zhang_Roth_2022} proposed an interface for interacting with API-based services, whereas META-GUI \cite{Sun_Chen_Chen_Dai_Zhu_Yu_2022} introduced a dataset focused on automating actions in mobile apps rather than general websites. In terms of dialogue-centric web navigation, RUSS \cite{Xu_Masling_Du_Campagna_Heck_Landay_Lam_2021} is the first dataset designed to help support services through 80 demonstrations annotated with a domain-specific language. \Benchmark{} extends previous dialogue-centric datasets by covering a wide range of real-world tasks spanning 2337 demonstrations, with considerably longer demonstrations due to dynamic topic switching, a subject studied by \citet{Adlakha_Dhuliawala_Suleman_de_Vries_Reddy_2022}.

\begin{figure}[t]
  \centering
  \includegraphics[width=0.7\linewidth]{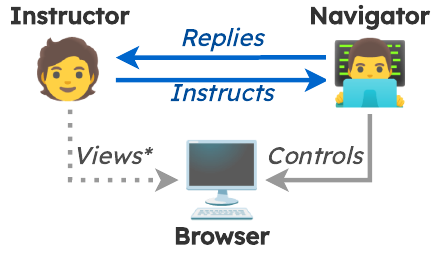}
  \vspace{-.5em}
  \caption{Data collection setup (\S\ref{sec:data_collection}). We record interactions (chat and browser actions) between an instructor and human navigator. *Instructor can see the screen except in \testvis{} split.}
  \label{fig:data_collection_smaller}
\end{figure}

\section{\Benchmark{}}
\label{sec:benchmark}

In this section, we introduce \Benchmark{}, a large-scale benchmark for conversational web navigation consisting of 2337 demonstrations with an average of 43 turns. It contains interactions between a human user (referred to as \textit{instructor}) and human assistant (\textit{navigator}) aiming to complete tasks across 155 real-world websites selected from 15 geographic areas. We classify the websites into 8 categories and 50 subcategories based on their domains.

\paragraph{Statistics} The data statistics are summarized in \autoref{tab:data_stats} and a breakdown by category and split is illustrated by \autoref{fig:scatter_stats}. Additional statistics about the dataset, including the number of demonstrations in  split, can be found in \cref{appendix:supplementary_statistics}, along with the list of categories in \cref{appendix:categories_and_subcategories}.

\paragraph{Demonstration Framework}
\label{sec:framework}
\label{paragraph:action_structure}
The demonstrations capture real-time interactions, which are recorded by the navigator controlling the web browser. Each demonstration $\mathcal{D} = \{s_{1},  a_{1}, \ldots, s_{n}, a_{n}\}$ is a sequence of $n$ states $s \in \mathcal{S}$ and actions $a \in \mathcal{A}$ . At each turn $t \in \{1, \ldots, n\}$, the state $s_t$ contains the representation of the website.
Each action follows one of the 5 core intents described in \autoref{tab:actions_trimmed}. The full list of intents is provided in \autoref{appendix:intents_descriptions}.

\begin{table}[t]
  \small \centering
  \caption{\small Overview of the \Benchmark{} core action space. For full set of actions, see \Cref{tab:actions_full}.}
  \label{tab:actions_trimmed}
  \begin{tabular}{ll}
    \toprule
    \textbf{Action}                   & \textbf{Description}       \\
    \midrule
    \texttt{click(element)}           & click on an element        \\
    \texttt{load(url)}                & load URL of a new page     \\
    \texttt{say(text)}                & navigator's utterance      \\
    \texttt{submit(element)}          & submit a form              \\
    \texttt{textinput(element,value)} & type text into the element \\
    \bottomrule
  \end{tabular}
\end{table}

\paragraph{Data Collection}
\label{sec:data_collection}
To collect the demonstrations, we worked with a professional data labeling company,\footnote{\EsyCommerce{}} who enlisted 8 expert annotators that received detailed instructions and extensive training to complete our tasks. The annotators worked in pairs: an instructor interacts with a navigator who completes the tasks in a web browser (see \autoref{fig:data_collection_smaller}). Both use the chat interface to communicate, but only the navigator controls the browser. We designed an app, browser extension, and processing pipeline to record the demonstrations, which are subsequently validated by a different annotator under the supervision of the original navigator (details in \cref{appendix:data_collection_details}).

\paragraph{Evaluation Splits}
In addition to a \textsc{Train} split, we create \textsc{Valid} and \testid{} to assess in-domain generalization, and 4 out-of-domain splits for various scenarios (see \autoref{tab:demo_splits}).

\subsection{Representing actions and states for modeling}
\label{sec:representation_action_states}

At each turn $t$, we have access to the state $s_t$ to predict an action $a_t$. The state consists of the following (if available):
\begin{compactitemize}
  \item $c_t$: Candidate elements that can be targeted by $a_t$,
  \item $d_t$: Current DOM tree of the page,
  \item $i_t$: Screenshot of the navigator's browser,
  \item $u_t$: Instructor's utterance,
  \item $v_t$: Viewport size (height and width),
  \item $h_t$: Interaction history.
\end{compactitemize}

Note that a state need not contain all of the above. For example, at the start of a demonstration, the instructor and navigator may need multiple rounds of dialogue to properly define the objective, in which case the initial states do not have DOM trees or screenshots.
A model $m$ predicts an action $a_t$ for a given state $s_t$ based on a prompt template $p_m$ which indicates how to make use of the contents in a state.

\paragraph{Interaction history}
Since a model $m$ has a limited input length in practice, we represent history $h$ as the set of past five actions (denoted as $a_r$) and five utterances ($u_r$).
We could not include the representation of past states such as elements or screenshots.

\paragraph{Parsing Action Output}
\label{paragraph:output_parsing}
An action consists of an intent and argument and can be generated by an agent in a textual format. It must follow a pre-defined structure (see \autoref{tab:actions_trimmed}) that allows it to be parsed into a structured form, which can be executed in a browser using tools like Selenium.\footnote{\url{https://www.selenium.dev/}} We discuss additional details in \cref{appendix:output_processing}.

\section{Evaluation Framework}
\label{sec:evaluation}
\label{section:metrics}

In this section, we describe the evaluation metrics (\S\ref{sec:metrics}) and their applicability to specific groups of intents (\S\ref{sec:grouping_intents}).

\subsection{Metrics}
\label{sec:metrics}

A commonly used metric in prior work on web navigation is \textit{task success rate}, which measures the proportion of demonstrations where the model reached the desired final state \cite{Shi_Karpathy_Fan_Hernandez_Liang_2017, Yao_Chen_Yang_Narasimhan_2023,deng2023mind2web}.
However, this metric is inappropriate for our benchmark because the objective is not fully defined in the first turn or later turns; instead, it evolves as the conversation proceeds.
We instead leverage \textit{turn-level} automatic evaluation metrics, following established approaches in dialogue systems \citep{rastogi2020towards,zhang2020dialogpt}. The aim of the metrics is to provide a heuristic estimate of the similarity between the predicted action and the reference action.

\paragraph*{Intent Match (IM)} Given prediction $a'$ and reference $a$, the intent match is $ \textsc{IM}(a',a)=1$ if the intents are equal, otherwise $\textsc{IM}(a',a)=0$. This tells us if a model can correctly identify which action to perform, but does not indicate if the model can predict the correct arguments.

\paragraph{Element Similarity using IoU}
For actions with elements as arguments (\texttt{click}, \texttt{textinput}, \texttt{submit}), we compute the \textbf{intersection over union} (\textbf{IoU}; \citealt{jaccard1912distribution}). Given the area of a bounding box $\mathcal{B}$, we have:
\begin{equation*}
  \textsc{IM}(a', a) \times \frac
  {\mathcal{B}_{\text{reference}} \cap \mathcal{B}_{\text{predicted}}}
  {\mathcal{B}_{\text{reference}} \cup \mathcal{B}_{\text{predicted}}}
\end{equation*}
To compute the area, we use \texttt{(x,y)} coordinates of the reference and predicted bounding boxes.
This formulation (1) favors elements with high visual overlap, (2) penalizes predicting elements much smaller or larger than reference elements even if one is completely contained by the other, and (3) assigns 0 if the elements
do not overlap.

\paragraph{Text Similarity using F1} To measure lexical similarity of text arguments in \texttt{say} and \texttt{textinput}, we calculate \textbf{chrF} \cite{popovic-2015-chrf}, an F1-score for character n-gram matches (we use the default setting of $n=6$). Similar to IoU, we scale by the IM, resulting in $\textsc{IM}(a',a) \times \textsc{chrF}(a',a)$.
In the case of \texttt{load} intent, URLs follow a structure that can be consistently segmented, which leads us to apply the F1-score on segments instead of n-grams; we call this measure \textbf{URLF}. We use \textbf{F1} to refer to either chrF and URLF, depending on whether an action contains a text or URL argument.

\subsection{Turn-level score and overall score}
\label{sec:grouping_intents}
To allow better comparisons between models, we divide the intents into groups: The \textbf{element group (EG)} contains \texttt{click}, \texttt{textinput}, and \texttt{submit}, and is evaluated with IoU. The \textbf{text group (TG)} encompasses \texttt{load}, \texttt{say}, and \texttt{textinput}, and is evaluated with F1.

We assign a turn level score based on the following:
If the turn involves an action in EG, the score is the same as IoU, i.e. score is 0 when the intent is incorrect or the element doesn't overlap, it is 1 when intent is correct and the element perfectly overlaps, and it is somewhere in between for the rest.
For TG actions \texttt{load} and \texttt{say}, the score is same as F1, i.e., score is 0 when either intent is incorrect or there is no text overlap, it is 1 when intent is correct and the text matches exactly, and it is somewhere in between for the rest.
For \texttt{textinput}, the turn score is $\text{IoU} \times \text{F1}$ since it contains both text and element arguments.
Finally, we compute the \textbf{overall score} using the \textbf{micro-average} of turn-level scores.

\section{Methods}
\label{sec:experiments}

In this section, we describe a method for selecting candidate elements (\S\ref{sec:DMR}) and how to use them in textual input.
We use these methods to build models that can accurately predict actions (\S\ref{sec:models_generating_actions}). We report results in \autoref{sec:results} and provide implementation details in \cref{appendix:additional_modeling_details}.

\subsection{Dense Markup Ranking (DMR) for Candidate Selection and Input Representation}
\label{sec:DMR}

To choose a set of suitable candidates for the model input (\S\ref{sec:representation_action_states}), we need a candidate selection stage that filters the full set of elements in the DOM tree. \citet{deng2023mind2web} proposed to pair each DOM element with the task query and input them into a DeBERTa model \cite{he2021deberta}, which is finetuned using a cross-encoder loss \cite{reimers-gurevych-2019-sentence}. We found this method takes on average 916ms to select candidates for a given turn.\footnote{Calculated on the training set, see \cref{appendix:paragraph_empirical_speed_improvements}.} When factoring in network latency and LLM inference, this would result in poor processing time.
It is thus crucial that we use efficient ranking method to build agents that can operate in real time and learn from interactions with users.

To solve this, we propose \textbf{Dense Markup Ranking (DMR)}, which is 5 times faster than the previous approach, at the cost of slightly lower recall.
The method consists of: (1) a simplified element representation to reduce computational overhead; (2) a dual encoder-based approach \citep{reimers-gurevych-2019-sentence, karpukhin-etal-2020-dpr}; (3) similarity-based learning between the text representation of $s_t$ and $a_{1:t-1}$ and corresponding HTML elements. Using this method, we finetune a variant of \textit{MiniLM} \citep{Wang2020MiniLMDS}.
We formulate the cosine-based learning objective, examine the inference speed improvements, and evaluate alternatives in %
\cref{appendix:dmr_details}.

\label{sec:text_processing_improvements}

\label{paragraph:strategic_truncation}
Even after our candidate selection, the input sequence length to a model can exceed its limit, so we truncate the sequence.
To reduce information loss from traditional truncation (e.g., for large DOM elements and long history), we design a strategy that leverages the hierarchical nature of the input to determine which subsection should be truncated.
\label{paragraph:otr}
We introduce several improvements to the representation used in prior works by including the full HTML attributes, viewport size, XML Path, and the bounding boxes of candidate elements
(implementation details in \cref{appendix:otr_details,appendix:truncation}).

\subsection{Modeling Actions}
\label{sec:models_generating_actions}

Upon selecting the most promising candidates for a given state $s_t$, we can combine them with the remaining information in $s_t$ to construct a representation that can be used to predict action strings, which can be parsed and executed (\S\ref{paragraph:output_parsing}). To understand which factors matter for predicting actions, we examine 19 zero-shot and finetuned models (using the \textsc{Train} split) with different input modalities: image-only, text-only, and both.
We provide implementation details in \cref{appendix:implementation_experiments} and hyperparameters in \cref{appendix:hyperparameters_details}.

\paragraph{Model Categories}
\label{sec:categories_of_models}
We categorize action models by the input modality, since the output is always in a structured format (\S\ref{paragraph:output_parsing}). We define the following types: (1) \textbf{text-only}, which receives instructions, pruned DOM tree, candidate element description and history; (2) \textbf{image-to-text}, which receives the screenshot, instructions and past actions directly embedded in the image; (3) multimodal, which receives the screenshot, instructions, pruned DOM tree, candidate description and history directly as text. Additional discussions are found in \cref{appendix:details_on_model_categorization}.

\paragraph{Text-only models}
\label{paragraph:text_based_models}
The recent \textbf{MindAct} \cite{deng2023mind2web} model is a Flan-T5 \citep{chung2022scaling__flant5} model that has been finetuned on Mind2Web.
We further fine-tune it  on \Benchmark{} using its original configuration.

To quantify the improvements brought by DMR-based representation (\S\ref{sec:DMR}), we directly finetune \textbf{Flan-T5} checkpoints, allowing us to control for size and architecture with respect to MindAct.
We also finetune \textbf{LLaMA-2} \cite{touvron2023llama,touvron2023llama2}\footnote{We use the variants finetuned on chat.} and a distilled version, \textbf{Sheared-LLaMA} (S-LLaMA; \citealt{xia2023sheared}).

\begin{table}[t]
  \centering
  \footnotesize
  \caption{Aggregated results (\S\ref{sec:results}) across major models (\S\ref{sec:experiments}), sorted by parameter count (Size).
    Following metrics from \autoref{sec:evaluation}, we report results of intent match (using \textbf{IM}), element group  (\textbf{IoU}), text group (\textbf{F1}), and the overall score (using micro-average on turn-level scores).
    All results are on \testod{} except the last column which is on \testid{}.
    \supeye ~indicates models with access to screenshots; every model except Pix2Act has access to text inputs.}
  \label{tab:test_ood_agg_results}
  
\setlength{\tabcolsep}{5.5pt}
\begin{tabular}{llrrrrr}
\toprule
 &  & \scriptsize Intent & \scriptsize Element & \scriptsize Text & \multicolumn{2}{c}{\scriptsize 
\bf Overall Score} \\
Models & \scriptsize Size & \scriptsize IM & \scriptsize IoU & \scriptsize F1 & \bf \scriptsize \testodcompact{} & \bf \scriptsize \testidcompact{}\\
\midrule
\textit{Zero-shot} & & & & & & \\
\cdashlinelr{1-7}
Llama-2 & 13B & 43.7 & 4.8 & 1.3 & 5.2 & 5.6\\
GPT-3.5T & -- & 42.8 & 8.6 & 3.5 & 8.5 & 10.3\\
GPT-4T & -- & 41.7 & 10.9 & 6.8 & 10.7 & 12.2 \\
GPT-4V\supeye & -- & 42.4 & 10.9 & 6.2 & 10.4 & 12.9 \\
\midrule
\textit{Finetuned} & & & & & \\
\cdashlinelr{1-7}
Pix2Act\supeye & 1.3B & 81.8 & 8.3 & 25.2 & 16.9 & 23.9 \\
S-LLaMA & 2.7B & \textbf{84.0} & 22.6 & \textbf{27.2} & 25.0 & \textbf{37.4} \\
MindAct & 3B & 79.9 & 16.5 & 23.2 & 20.9 & 25.7 \\
Flan-T5 & 3B & 81.1 & 20.3 & 25.8 & 23.8 & 31.1 \\
Fuyu\supeye & 8B & 80.1 & 15.7 & 22.3 & 20.0 & 30.9 \\
Llama-2 & 13B & 83.0 & \textbf{22.8} & 26.6 & \textbf{25.2} & 37.0 \\
GPT-3.5F & -- & 77.6 & 18.6 & 22.4 & 21.2 & 30.8 \\
\bottomrule
\end{tabular}

  \vspace{-.2em}
\end{table}

\paragraph{Proprietary text-only LLMs}
\label{paragraph:proprietary_llms}
We report results for GPT-3.5 Turbo \citep{brown2020language, Andrew_Peng_Michael_Wu_Logan_Kilpatrick_Steven_Heidel_2023}, in both zero-shot (\textbf{3.5T}) and finetuned (\textbf{3.5F}) settings. We also include zero-shot results for \textbf{GPT-4T} \citep{OpenAI2023GPT4TR}.

\paragraph{Image-to-text modeling}
\label{paragraph:image_to_text_model}
We explore \textbf{Pix2Act} \cite{shaw2023pixels} an encoder-decoder \cite{Vaswani2017AttentionIA} purely finetuned on pixels. It uses a Pix2Struct backbone \citep{Lee_2022_Pix2Struct}, which is pretrained on screenshots using a Vision Transformer encoder \citep{dosovitskiy2020image} and a text decoder. We follow the behavior cloning approach used by Pix2Act by finetuning the same backbone on \Benchmark{}.

\paragraph{Multimodal models}
\label{section:multimodal_models}
We finetune \textbf{Fuyu-8B} \citep{fuyu_blog_post}, a base model pretrained on browser screenshots by modeling images and text using a unified architecture. We also report zero-shot results for the variant of GPT-4 with vision capabilities (\textbf{GPT-4V}; \citealt{GPT4V_System_card}).

\section{Experimental Results}
\label{sec:results}

In this section, we report the results of our experiments (\S\ref{sec:experiments}) for groups defined in \autoref{sec:grouping_intents}. We aggregate the results for 11 models in \autoref{tab:test_ood_agg_results}. In \autoref{sec:qualitative_assessment}, we qualitatively assess two major models: GPT-4V and LLaMA-2-13B. See \cref{appendix:supplementary_results} for supplementary results and \cref{appendix:additional_results} for the detailed overview (including the remaining 8 variants).

\begin{table}[t]
  \centering
  \footnotesize
  \caption{\footnotesize{Results on out-of-domain splits (\S\ref{tab:demo_splits}) for finetuned LLaMA-2-13B (\S\ref{sec:models_generating_actions}).
      Among the splits, \testcat{} seems to be the hardest, indicating that models struggle on unseen subcategories (e.g., restaurant appointment vs. medical appointment).}}
  \label{tab:test_breakdown_by_ood_results}
  \begin{tabular}{lrrrr}
    \toprule
    \multirow{ 2}{*}{Splits} & Intent & Element & Text & \multirow{ 2}{*}{Overall} \\
                             & IM     & IoU     & F1   &                           \\
    \midrule
    \testweb{}               & 82.7   & 24.2    & 28.7 & 27.0                      \\
    \testcat{}               & 81.0   & 20.7    & 26.1 & 24.3                      \\
    \testgeo{}               & 78.6   & 22.0    & 27.7 & 25.9                      \\
    \testvis{}               & 85.3   & 26.1    & 23.9 & 25.0                      \\
    \bottomrule
  \end{tabular}
\end{table}

\subsection{Overview of Results}
\label{sec:overview_results}

\paragraph{Impact of representation for text-only models}
In \autoref{tab:test_ood_agg_results}, we observe that MindAct trails behind Flan-T5 finetuned using DMR-based input representation (\S\ref{paragraph:otr}), when comparing the 3B-parameter variants. Although MindAct was finetuned for a related task, it was never exposed to multi-turn dialogue. However, Flan-T5 was never trained on any navigation actions. Thus, DMR-based representation plays an important role in achieving a better performance for the same architecture and model size.
Moreover, both LLaMa-based models outperform Flan-T5 and MindAct despite Sheared-LLaMa being smaller than Flan-T5.
This could be due to the high quality training of LLaMa models on a large number of instruction-following tasks compared to Flan-T5.
However, it is intriguing that Sheared-LLaMa performs equally well compared to LLaMA-2 13B.

\paragraph{Image-to-text vs. multimodal models}
We further highlight the difference between smaller image-to-text and larger multimodal models by comparing Pix2Act (1.3B parameters) and Fuyu-8B. Overall, Fuyu outperforms Pix2Act, which could be due its ability to receive text as input and greater parameter count. However, it trails behind Pix2Act for intent matching and text prediction.

\paragraph{Comparing multimodal with chat-based models}
We observe that Fuyu-8B is outperformed by chat-based text-only LLaMA models.
This shows that multimodal models finetuned on screenshots are still behind chat-based models optimized for instruction-based finetuning.

\paragraph{Comparison with proprietary models}
In the zero-shot setting, where models solely rely on the instructions, we observe that proprietary models (GPT-3.5T and GPT-4T) outperform the open-sourced LLaMA-2. However, when finetuned, GPT-3.5F is outperformed by Sheared-LLaMA and LLaMA-2, but the cause is unclear as most hyperparameters are inaccessible for commercial training. Finally, GPT-4V and GPT-4T achieve similar performance, suggesting that existing multimodal models might not be able to effectively use screenshots for predicting actions.

\paragraph{Generalization capabilities}
When comparing \testod{} with \testid{} results, we observe a major difference across all finetuned models. This highlights a weakness of finetuned models: although they perform well on familiar websites, they will struggle to generalize to unseen websites.
For example, we observe in \Cref{tab:test_breakdown_by_ood_results} that LLaMa-13B achieves poor results on \testcat{}, indicating that unseen subcategories are more challenging than new websites from the same categories. For instance, if the model learns how to book seats at a restaurant, it can adapt to a different restaurant but will struggle to book a medical appointment.

\begin{figure*}[t]
  \centering
  \footnotesize
  \begin{tabular}{p{0.47\textwidth} p{0.47\textwidth}}
    \includegraphics[width=0.45\textwidth]{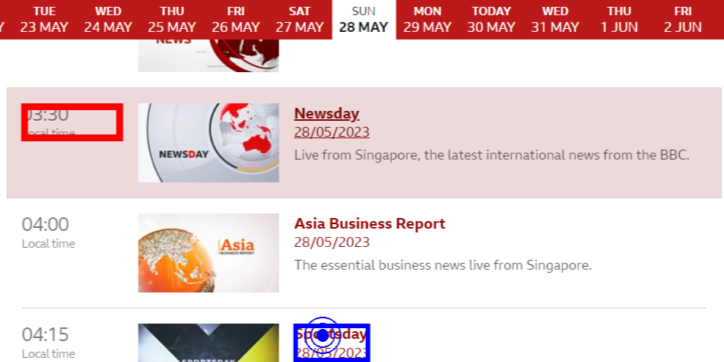}
     &
    \includegraphics[width=0.45\textwidth]{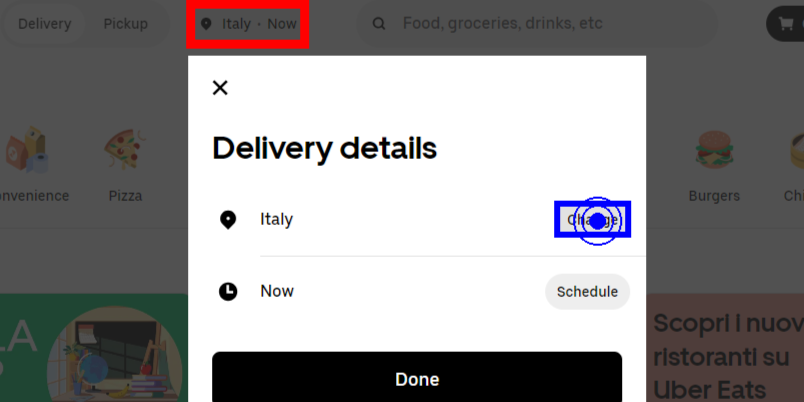}
    \\
    \textit{\textbf{C1:} Instructor wants Navigator to open a specific tab on a News site, i.e., "Sportsday on 28 May 2023 at 4.15 AM".}
     &
    \textit{\textbf{C2:} Instructor requests the location on a food delivery site to be set to \textit{Las Vegas}. The \textit{Delivery details} window is already open.}
    \\
    \vspace{.05em}
    \textbf{\txred{GPT-4V (R)}} clicks on an incorrect (3:30AM) tab, even though the instructor requested a different time (4:15AM).
     &
    \vspace{.05em}
    \textbf{\txred{GPT-4V (R)}} attempts to exit the Delivery details page and reopen it, which could potentially lead to a loop.
    \\
    \vspace{-.5em}
    \textbf{\txblue{LLaMA-WL (B)}} clicks on the correct 4:15AM tab.
     &
    \vspace{-.5em}
    \textbf{\txblue{LLaMA-WL (B)}} correctly clicks on the \textit{Change} button.
    \\
    \midrule
    \includegraphics[width=0.45\textwidth]{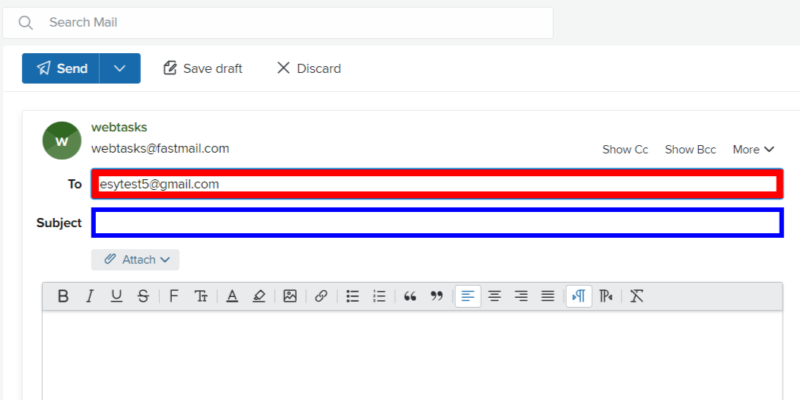}
     &
    \includegraphics[width=0.45\textwidth]{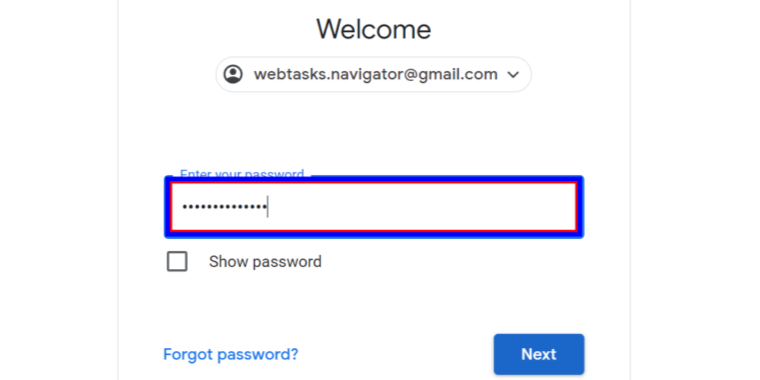}
    \\
    \textit{\textbf{T1:} Compose a ``Invitation to Collaboration" email.}
     &
    \textit{\textbf{T2:} Open Google translate and sign in using the following credentials: [email] [password]}
    \\
    \textbf{\txblue{Reference (B):}} Invitation to Collaboration
     &
    \textbf{\txblue{Reference (B):}} \textit{[password]}
    \\
    \textbf{\txred{GPT-4V (R):}} Leon Tales
     &
    \textbf{\txred{GPT-4V (R):}} \textit{[email]}
    \\
    \textbf{\txblue{LLaMA-WL (B):}} Invitation to Collaboration
     &
    \textbf{\txblue{LLaMA-WL (B):}} \textit{[password]}
    \\
    \midrule
    \textit{\textbf{S1}: Instructor says ``Like \& Bookmark this listing.''}
     &
    \textit{\textbf{S2}: Instructor says `` Please share the link.``}
    \\
    \vspace{-.5em}
    \textbf{\txblue{Reference}}: Alright
     &
    \vspace{-.5em}
    \textbf{\txblue{Reference}}: Alright.
    \\
    \vspace{-.75em}
    \textbf{\txblue{LLaMA-WL}}: Alright
     &
    \vspace{-.75em}
    \textbf{\txblue{LLaMA-WL}}: Okay.
    \\
    \vspace{-.75em}
    \textbf{\txred{GPT-4V}}: {Understood. If you need further assistance, just let me know.}
     &
    \vspace{-.75em}
    \textbf{\txred{GPT-4V}}: Here is the link to the discussion: [\textit{incorrect link}]
    \\

    \bottomrule
\end{tabular}

  \caption{Comparison of GPT-4V and LLaMA-2-13B (finetuned) on predicting \texttt{click} actions. Incorrectly predicted actions are in \txred{red (R)}, reference actions are in \txblue{blue (B)}. We show two scenarios for \texttt{click} (C1,C2), \texttt{textinput} (T1,T2) and \texttt{say} (S1, S2).}
  \label{fig:qualitative_comparison_combined}
\end{figure*}

\subsection{Qualitative Assessment}
\label{sec:qualitative_assessment}

To better understand the performance gap separating the strongest zero-shot and finetuned models, we qualitatively examine two models, GPT-4V and LLaMA-2-13B, which respectively represent the two paradigms. Although the gap can be partially attributed to incorrectly predicted intents (see \cref{appendix:additional_results}), models can still make poor predictions even when the intent is  predicted correctly. We focus on this scenario by assessing actions from 3 intents: \texttt{click}, \texttt{textinput} and \texttt{say}; for each, we show two examples in \autoref{fig:qualitative_comparison_combined}. Extended assessments can be found in \cref{appendix:extended_qualitative_assessment}.

\paragraph{Assessing \texttt{click}}
In scenarios where models select objects through clicks, we find that GPT-4V chose an incorrect tab (C1), was unaware it has already started a sub-task (C2), and chose a less optimal option (see \cref{appendix:extended_qualitative_assessment}). Although those scenarios are correctly addressed by the finetuned LLaMA-2, it can still fail by clicking on irrelevant elements (even when GPT-4V selects the correct one).

\paragraph{Assessing \texttt{textinput}}
When looking at examples where models are selecting and typing text inside inputs, we observe that GPT-4V tried to write the name of a email recipient instead of the subject title (T1), the username inside a password field (T2), typed a passage already in the target textbox, and skip the title when drafting a post. Although LLaMA succeeded in the first two cases, it may attempt to \texttt{click} instead of \texttt{textinput} and also omit the title.

\paragraph{Assessing \texttt{say}}
For \texttt{say} actions, GPT-4V used a different writing style (S1), whereas LLaMA-2 learned the writing style of the annotators. Additionally, GPT-4V provided unhelpful responses by sharing irrelevant links (S2) and refused to assist the instructor even when it is possible. Even though LLaMA-2 is finetuned, it missed certain follow-up questions (such as asking ``Who should receive this?'' when asked to write an email).

\section{Discussion}
\label{sec:discussion}

\subsection{Experimental Findings}
\label{section:findings}
Through our experiments (\autoref{sec:experiments}), we find that larger multimodal models can surpass smaller image-only models when finetuned, but they are still behind finetuned text-only models. We also find that employing an DMR-based representation leads to better performance (\S\ref{sec:overview_results}). When evaluated on out-of-domain splits, the performance of text-only decoders are very close to smaller variant; nonetheless, zero-shot models are consistently surpassed by their finetuned counterparts. We confirm, through qualitative assessments (\S\ref{sec:qualitative_assessment}), that even the best zero-shot models can make simple and unjustified errors. Our findings highlight the need to build models that can better generalize to unseen scenarios if we want to build agents that will work in the real world.

\subsection{Limitations}
Our benchmark contains only static demonstrations, which means we cannot meaningfully evaluate the behavior of models on alternative trajectories. However, this approach lets us train models on a diverse set of real websites that do not need to be recreated from scratch.

\paragraph{Generalizability} There are inherent limitations of the architectures we evaluate. For example, we cannot expect a text-only model to draw on a canvas or describe images. Such limitations can be addressed through multimodal-specific technical contributions in future works.

\section{Conclusion}
We introduced \Benchmark{}, a large-scale expert-built benchmark covering a wide range of demonstrations for conversational web navigation on real-world websites.
The framework we built around the benchmark includes the task definition, data representation, and evaluation metrics.
We also introduced a dense markup ranker (DMR) to effectively summarize webpages.
We evaluated finetuned and zero-shot models with various modalities, and found that chat-based decoder models finetuned on \Benchmark{} achieve the best results, but still struggle to generalize to out-of-domain splits.
We believe that multi-turn dialogue can enhance flexibility and steerability of agents for web navigation, leading to their wider adoption.

To overcome these model limitations, we suggest the following future directions:
  \begin{compactitemize}
    \item Designing multimodal architectures that can efficiently process visual input with structured information.
    \item Evaluating models in environments covering wider ranges of scenarios, including complex websites, advanced browser events.
    \item Expand to tasks beyond the browser, such as OS-level interactions \citep{xie2024osworld}.
    \item Leveraging reward-based methods like RLHF \citep{christiano2017deep_RLHF} and DPO \citep{rafailov2023direct_dpo}.
    \item Leveraging alternative training approaches such as self-experience and grounded synthesis \citep{gur-2023-real-world-webagent}.
  \end{compactitemize}

\section*{Acknowledgments}

XHL acknowledges the support of the Natural Sciences and Engineering Research Council of Canada (NSERC) [funding reference no. 579403].
ZK is supported by the European Union (ERC, NG-NLG, 101039303)
 and Charles University project SVV 260~698. SR is supported by a Facebook CIFAR AI Chair and NSERC Discovery Grant program. The project is supported partially by the Google-Mila grant. We thank Esycommerce for providing their data annotation services and actively working with us in order to reach a consistent and high quality data collection process. We thank Benno Krojer, Chris Pal, Dilek Hakkani-Tür, Gokhan Tur, Ismail Haritaoglu, Nicolas Chapados, Ondřej Dušek, Peter Shaw, Sai Rajeswar, Vaibhav Adlakha, the UI Assist team at ServiceNow Research, and the McGill NLP group members for helpful discussions.

\section*{Impact Statement}
Web navigation agents have the potential to become a powerful technology with large societal impacts. Therefore, multiple aspects need to be taken into consideration when conducting further research in this area:

\paragraph{Automating vs. Elevating Users}
A major risk of fully automating web navigation is the automation of work traditionally performed by knowledge workers; deploying highly capable models could lead to job losses. However, one major difference between autonomous navigation and our framework is that we require the inclusion of a human instructor to provide the real-time instructions needed to complete the task. Thus, conversational web navigation's ultimate purpose is not to automate what a user does, but automate difficult, repetitive, and error-prone steps so that the user can focus on reliably solving high-level problems.

\paragraph{Malicious Usage and Mitigation}
As web navigation models become increasingly sophisticated, there are risks that they will be used for malicious purposes at scale.
These models can automate harmful activities, e.g., for creating spam messages and impersonating individuals for fraudulent purposes. While these activities can already be partially automated using open-source tools,\footnote{For example, Selenium: \url{https://www.selenium.dev/}} web navigation agents could make automation easier and more robust. However, malicious actors can build such models in private using existing commercial services, independent of on-going research on agents. On the other hand, by making our models and data accessible to researchers, our work can be used to research ways to mitigate the risk of malicious usage; for instance, by incorporating our models as part of red teaming procedures. The resulting research can be used to build systems that are robust against malicious agents.

\paragraph{Unintended Actions}
Navigation agents can cause harm if they misinterpret instructions and perform unintended actions; for instance, booking the wrong flight could result in significant financial loss. For this reason, we assert that conversational web navigation models should be used under human supervision (where multi-turn dialogue cannot be disabled), and that it should only be deployed after exhaustive testing with proper safeguards. Our models should not be deployed and should only be used for research.

\paragraph{Data Collection}
To build \Benchmark{}, we worked with expert annotators, who received training, familiarized with the task and the purpose of the project, and were paid fair wage relative to their country of employment. The websites in our dataset are publicly accessible and safe. Any account appearing in the dataset was specifically created for the data collection; there are no references to their identity to preserve their privacy.

\bibliographystyle{icml2023}
\bibliography{custom}

\newpage
\onecolumn

\tableofcontents

\newpage
\appendix
\label{sec:appendix}

\section*{Appendix}

\section{Dataset Details}
\label{appendix:dataset}

\subsection{Supplementary Statistics}
\label{appendix:supplementary_statistics}

In \autoref{sec:benchmark}, we introduce \Benchmark{}. In this section, we provide supplementary statistics for readers wishing to gain a deeper understanding of the dataset.

\noindent In \autoref{tab:dataset_stats_turns_by_intent}, we report demo and turn statistics by intent. We observe that \texttt{say}, \texttt{click} and \texttt{load} are heavily represented across demos. However, the latter happens less often than other intents. This is because the user loads new links only when they move to a new website, and many tasks can be accomplished within the same page (such as booking a flight). Therefore, there is no need to load new pages as frequently as other intents. Additionally, \texttt{hover} is less represented due to the removal of unnecessary hovering, which can be accidentally recorded when moving the cursor across non-target elements with callbacks.

In \autoref{tab:dataset_stats_turns_by_split}, we present the number of demos for each split and mean number of turns. Although most demos are in the range of 40-50 turns, the number of demos in the \testvis{} split is substantially lower. This can be attributed to the lack of follow-up based on what is happening on the screen. For example, an instructor with vision can request the navigator to apply some specific filters (e.g., by saying "Please apply the filter for Japan Airlines under the Airlines filter option"), whereas an instructor without vision would not have this request unless they are using a screen-reader.

\begin{table}[h]
  \footnotesize \centering
  \caption{\footnotesize Complete list of \Benchmark{} \textit{observed} action space. Note that a speaker can either be \texttt{navigator} or \texttt{instructor}, but an agent is only permitted to choose \texttt{navigator}, since \texttt{speaker="instructor"} is not a valid action by an agent.Tab actions (create, remove, switch) are under `chrome.tabs`. (*)`onload` and `location` are both methods of `window'. }
    \begin{tabular}{l p{0.25\linewidth} ll}
    \toprule
    \textbf{Action}                                  & \textbf{Description}                              & \textbf{Listener}  & \textbf{Method/event trigger}                      \\
    \midrule
    \textttc{say(speaker=[role],utterance=[str])}    & talking to instructor or navigator                & ---                      & ---                                  \\
    \textttc{click(uid=[element])}                    & click on an element                               & onclick                  & \textttc{HTMLElement.click()}         \\
    \textttc{click(x=[int],y=[int])}                  & or its corresponding coordinates    & onclick                  & \textttc{HTMLElement.click()}         \\
    \textttc{hover(uid=[element])}                    & hover over an element                             & onmouseover              & \textttc{MouseEvent(`mouseenter')} \\
    \textttc{hover(x=[int],y=[int])}                  & or its corresponding coordinates  & onmouseover              & \textttc{MouseEvent(`mouseenter')} \\
    \textttc{textinput(uid=[element],value=[str])}   & type text into the element                        & oninput                  & \textttc{Event('input')}         \\
    \textttc{change(uid=[element],value=[str])}      & change the value of the element to another option & onchange                 & \textttc{Event('change')}             \\
    \textttc{load(url=[link])}                        & load the URL of a new webpage                     & onload*                  & \textttc{location.href}               \\
    \textttc{submit(uid=[element])}                   & submit the form                                   & onsubmit                 & \textttc{HTMLFormElement.submit()}    \\
    \textttc{scroll(x=[int],y=[int])}                 & scroll to the coordinates                         & onscroll                 & \textttc{window.scrollTo(x,y)}        \\
    \textttc{copy(uid=[element],text=[str])}         & copy the text from the element                    & oncopy                   & \textttc{ClipboardEvent(`copy')}  \\
    \textttc{paste(uid=[element],text=[str])}        & paste the text into the element                   & onpaste                  & \textttc{ClipboardEvent(`paste')} \\
    \textttc{tabCreate()}                             & create a new tab                                  & tabs.onCreated           & \textttc{tabs.create()}               \\
    \textttc{tabRemove(target=[tabId])}               & remove the tab                                    & tabs.onRemoved           & \textttc{tabs.remove()}               \\
    \textttc{tabSwitch(origin=[tabId],target=[tabId])} & switch between tabs                               & onUpdated                & \textttc{tabs.update()}               \\
    \bottomrule
    \end{tabular}
  \label{tab:actions_full}
\end{table}

\begin{figure}[h]
  \centering
  \includegraphics[width=0.7\textwidth]{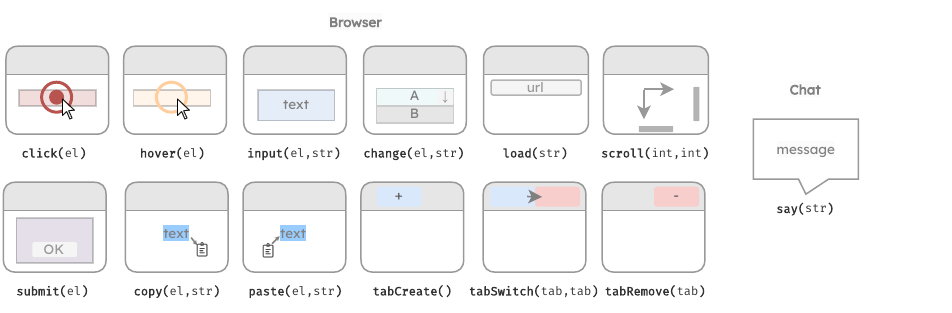}
  \caption{\small{Overview of the actions in our benchmark, including 10 browser actions and 1 chat action. An argument of an action can be a string (\texttt{str}), an integer (\texttt{int}), an element (\texttt{el}), or a browser tab id (\texttt{tab}). The intents are described in \autoref{tab:actions_full}.}}\label{fig:actions_full}
\end{figure}

\begin{table}[h]
  \centering
  \begin{minipage}{.5\textwidth}

    \centering
    \footnotesize
    \caption{\small{Turn-level stats by intent.}}
    \label{tab:dataset_stats_turns_by_intent}
    \begin{tabular}{lrrrr}
   \toprule
   Intent    & \# Demos & $\mu$ turns & $\sigma$ turns & Total \\
   \midrule
   say       & 2337     & 16.82       & 5.62           & 39305 \\
   click     & 2333     & 14.52       & 10.16          & 33865 \\
   load      & 2324     & 1.59        & 1.07           & 3702  \\
   copy      & 1587     & 4.08        & 3.05           & 6477  \\
   textInput & 1465     & 3.28        & 3.06           & 4799  \\
   paste     & 1130     & 1.89        & 1.95           & 2141  \\
   scroll    & 1046     & 3.82        & 3.00           & 3999  \\
   tabswitch & 800      & 3.28        & 3.65           & 2621  \\
   tabcreate & 712      & 1.71        & 1.12           & 1220  \\
   submit    & 645      & 1.40        & 1.11           & 904   \\
   hover     & 361      & 1.55        & 1.11           & 560   \\
   tabremove & 309      & 1.94        & 1.17           & 599   \\
   change    & 165      & 1.95        & 1.34           & 322   \\
   \bottomrule
\end{tabular}

  \end{minipage}%
  \begin{minipage}{.5\textwidth}
    \centering
    \footnotesize
    \caption{\small{Turn-level stats by split. Active turns are used for either finetuning or evaluation. Total includes turns used in history.}}
    \label{tab:dataset_stats_turns_by_split}
    \begin{tabular}{lrrrrr}
\toprule
     Split &          \# Demos &  $\mu$ turns &  $\sigma$ turns &  Active &  Total \\
\midrule
     \textsc{train} &            969 &         44.93 &       17.37 &           24418 &          43538 \\
\textsc{valid} &            100 &         40.76 &       14.51 &            1717 &           4076 \\
   \testid{} &            100 &         43.18 &       16.08 &            1846 &           4318 \\
  \testcat{} &            223 &         45.30 &       25.43 &            4979 &          10102 \\
  \testweb{} &            211 &         40.47 &       18.17 &            4184 &           8540 \\
  \testvis{} &            444 &         36.05 &       20.09 &            7725 &          16006 \\
  \testgeo{} &            290 &         48.05 &       18.66 &            6141 &          13934 \\
\bottomrule
\end{tabular}

    \caption{\small{Turn-level stats by use of AI tools (e.g., ChatGPT)}}
    \label{tab:dataset_stats_turns_by_ai_usage}
    \begin{tabular}{lrrrr}
\toprule
 Uses AI &  \# Demos &  $\mu$ turns &  $\sigma$ turns &  Total \\
\midrule
   \xmark &           2057 &         42.50 &        19.5 &          87414 \\
    \cmark &            280 &         46.79 &        16.9 &          13100 \\
\bottomrule
\end{tabular}

  \end{minipage}
\end{table}

\noindent In \autoref{tab:dataset_stats_turns_by_ai_usage}, we highlight the usage frequency of AI tools, which are listed in \autoref{tab:websites}. For certain tasks, such as summarizing news articles, it is much more convenient to use AI tools. Since we focus on actions executed, models can learn general actions when dealing with AI tools, even when the tools themselves changes.

\subsection{Categories and Subcategories}
\label{appendix:categories_and_subcategories}

In \autoref{sec:benchmark}, discuss the use of categories to classify demonstrations. We have in total 8 categories, each with their own subcategories, which add up to a total of 50 (\S\ref{tab:demos_and_distribution_stats}); we assign one category and subcategory to Each of the 155 URL sub-domain associated with a demo turn (\S\ref{tab:websites}). Since a demo may leverage multiple websites (e.g. composing and information lookup), a demo will have one or more subcategory. We give the full list of categories, subcategories, and the number of demonstrations associated with each in \cref{tab:demos_and_distribution_stats}.

In \autoref{tab:categories_and_subcategories_across_splits}, we show the breakdown of subcategories for the \testcat{} split (designed to test generalization to new subcategories). We note that the subcategories were automatically chosen to be the ones with the fewer occurrences across demos, allowing to have a reasonable split size.

\begin{table}[h!]
  \centering
  \footnotesize
  \caption{List of subcategories based on splits.}\label{tab:categories_and_subcategories_across_splits}
  \begin{tabular}{l p{0.7\linewidth}}
    \toprule
    \testcat{} & Spreadsheet, Handmade, Reviews, Computer Vision, Chatbot, Transport, Presentation, Furniture, Professional Network, Books, Tasks, Automatic Translation, Question Answering, Encyclopedia, Recipe, Geography                                                                                                                                                                                                                                                                                                                                                                  \\
    \midrule
    Others     & Stay, Stays, Transport, Scientific Articles, Online Shopping, Tasks, Blog, Discussion Platform, Recipe, Spreadsheet, Email, Research Directory, Music Sharing, Chatbot, Presentation, Grocery, Delivery, Image Sharing, Automatic Translation, Video Sharing, Encyclopedia, News Articles, Forum, Entertainment, Magazine, Medical, Furniture, Educational, Kanban, Social Network, Image Generation, Question Answering, Media, Note taking, Agency, Government, Social Event, Cooking, Instant Messaging, Finance, Books, Clothing, Restaurant, Calendar, Writing Assistant \\
    \midrule
    Difference & Handmade, Reviews, Computer Vision, Professional Network, Geography                                                                                                                                                                                                                                                                                                                                                                                                                                                                                                           \\
    \bottomrule
  \end{tabular}
\end{table}

\begin{table}[]
  \centering
  \caption{Number of demos each subcategory appears in for each split. Note that a demo might have multiple subcategories when using more than one website (for example, Information Lookup and Composing). In the last column, we also include the number of URLs associated with each subcategory; they correspond to the websites in \autoref{tab:websites}.}
  \label{tab:demos_and_distribution_stats}
  \scriptsize
  \begin{adjustbox}{width=\textwidth}
    \begin{tabular}{llrrrrrrrrr}
        \toprule
        Category                               & Subcategory          & Total & Train & Valid & ID & Vis & Geo & Cat & Web & \# URLs \\
        \midrule
        \multirow[t]{4}{*}{AI Tools}           & Auto. Translation    & 53    & 0     & 0     & 0  & 10  & 0   & 43  & 0   & 4       \\
                                               & Chatbot              & 408   & 178   & 19    & 21 & 82  & 42  & 31  & 35  & 3       \\
                                               & Computer Vision      & 13    & 0     & 0     & 0  & 0   & 0   & 13  & 0   & 1       \\
                                               & Image Generation     & 59    & 33    & 7     & 3  & 5   & 0   & 0   & 11  & 4       \\
                                               & Writing Assistant    & 70    & 44    & 3     & 2  & 11  & 0   & 0   & 10  & 5       \\
        \midrule
        \multirow[t]{6}{*}{Booking}            & Medical              & 34    & 0     & 0     & 0  & 9   & 25  & 0   & 0   & 3       \\
                                               & Restaurant           & 77    & 28    & 6     & 5  & 14  & 24  & 0   & 0   & 6       \\
                                               & Social Event         & 14    & 0     & 0     & 0  & 0   & 14  & 0   & 0   & 3       \\
                                               & Stay                 & 64    & 44    & 0     & 0  & 5   & 15  & 0   & 0   & 7       \\
                                               & Stays                & 37    & 24    & 0     & 0  & 11  & 0   & 0   & 2   & 3       \\
                                               & Transport            & 757   & 314   & 27    & 31 & 252 & 36  & 61  & 36  & 8       \\
        \midrule
        \multirow[t]{5}{*}{Composing}          & Blog                 & 62    & 34    & 2     & 3  & 15  & 0   & 0   & 8   & 4       \\
                                               & Email                & 135   & 86    & 10    & 17 & 16  & 0   & 0   & 6   & 6       \\
                                               & Note taking          & 47    & 31    & 0     & 5  & 11  & 0   & 0   & 0   & 4       \\
                                               & Recipe               & 20    & 0     & 0     & 0  & 3   & 0   & 17  & 0   & 1       \\
                                               & Tasks                & 31    & 0     & 0     & 0  & 10  & 0   & 21  & 0   & 2       \\
        \midrule
        \multirow[t]{9}{*}{Information Lookup} & Agency               & 46    & 29    & 2     & 3  & 0   & 0   & 0   & 12  & 3       \\
                                               & Educational          & 56    & 28    & 3     & 2  & 8   & 0   & 0   & 15  & 2       \\
                                               & Encyclopedia         & 97    & 56    & 8     & 7  & 11  & 0   & 1   & 14  & 4       \\
                                               & Entertainment        & 36    & 13    & 0     & 0  & 10  & 0   & 0   & 13  & 2       \\
                                               & Forum                & 37    & 12    & 4     & 1  & 9   & 0   & 0   & 11  & 2       \\
                                               & Geography            & 13    & 0     & 0     & 0  & 0   & 0   & 13  & 0   & 1       \\
                                               & Government           & 36    & 0     & 0     & 0  & 9   & 27  & 0   & 0   & 2       \\
                                               & Media                & 60    & 23    & 2     & 3  & 10  & 0   & 0   & 22  & 2       \\
                                               & Research Directory   & 10    & 0     & 0     & 0  & 10  & 0   & 0   & 0   & 2       \\
        \midrule
        \multirow[t]{5}{*}{Productivity}       & Calendar             & 50    & 17    & 3     & 2  & 11  & 3   & 0   & 14  & 2       \\
                                               & Finance              & 59    & 21    & 0     & 0  & 10  & 28  & 0   & 0   & 4       \\
                                               & Kanban               & 50    & 20    & 2     & 3  & 16  & 0   & 0   & 9   & 3       \\
                                               & Presentation         & 32    & 0     & 0     & 0  & 6   & 0   & 26  & 0   & 1       \\
                                               & Spreadsheet          & 27    & 0     & 0     & 0  & 10  & 0   & 17  & 0   & 2       \\
        \midrule
        \multirow[t]{6}{*}{Shopping}           & Clothing             & 93    & 18    & 6     & 4  & 8   & 57  & 0   & 0   & 6       \\
                                               & Delivery             & 91    & 67    & 4     & 6  & 14  & 0   & 0   & 0   & 7       \\
                                               & Furniture            & 6     & 0     & 0     & 0  & 5   & 0   & 1   & 0   & 1       \\
                                               & Grocery              & 38    & 0     & 0     & 0  & 8   & 30  & 0   & 0   & 2       \\
                                               & Handmade             & 15    & 0     & 0     & 0  & 0   & 0   & 15  & 0   & 1       \\
                                               & Online Shopping      & 87    & 51    & 3     & 2  & 31  & 0   & 0   & 0   & 7       \\
        \midrule
        \multirow[t]{8}{*}{Social Interaction} & Discussion Platform  & 32    & 18    & 4     & 1  & 9   & 0   & 0   & 0   & 3       \\
                                               & Image Sharing        & 60    & 30    & 6     & 9  & 0   & 0   & 0   & 15  & 4       \\
                                               & Instant Messaging    & 32    & 11    & 0     & 0  & 11  & 0   & 0   & 10  & 2       \\
                                               & Music Sharing        & 36    & 14    & 0     & 0  & 9   & 0   & 0   & 13  & 2       \\
                                               & Professional Network & 14    & 0     & 0     & 0  & 0   & 0   & 14  & 0   & 1       \\
                                               & Question Answering   & 20    & 0     & 0     & 0  & 5   & 0   & 15  & 0   & 1       \\
                                               & Social Network       & 62    & 28    & 4     & 2  & 13  & 14  & 0   & 1   & 4       \\
                                               & Video Sharing        & 20    & 10    & 0     & 0  & 1   & 0   & 0   & 9   & 1       \\
        \midrule
        \multirow[t]{6}{*}{Summarizing}        & Books                & 25    & 0     & 0     & 0  & 10  & 0   & 15  & 0   & 2       \\
                                               & Cooking              & 40    & 13    & 0     & 0  & 11  & 16  & 0   & 0   & 2       \\
                                               & Magazine             & 49    & 24    & 0     & 1  & 11  & 13  & 0   & 0   & 4       \\
                                               & News Articles        & 124   & 75    & 11    & 11 & 15  & 12  & 0   & 0   & 5       \\
                                               & Reviews              & 13    & 0     & 0     & 0  & 0   & 0   & 13  & 0   & 1       \\
                                               & Scientific Articles  & 35    & 10    & 4     & 2  & 10  & 0   & 0   & 9   & 2       \\
        \bottomrule
    \end{tabular}

\end{adjustbox}

\end{table}

\subsection{Input Processing Details}
\label{appendix:additional_data_processing_details}

In \autoref{sec:representation_action_states}, we introduce the components of a state $s_t$. More formally, we define the input of a model $m$ to be $\mathcal{P}_m(s_t, a_{1:t-1})$, consisting of a processing function $\mathcal{P}_m$ that receives $s_t$ and $a_{1:t-1}$ and returns a representation that can serve as an input to a model. We provide details of our method below.

\paragraph{Adapting $\mathcal{P}$ per model}
For each model $m$, we tailor the function $\mathcal{P}_m$ to accommodate for differences in methodology. For image-to-text models, we sequentially render $v_t$, $u_r$, $a_r$ as header text of the screenshot $i_t$ (viewport $v_t$ is included so models can locate bounding boxes of $c_t$). For text-only models, we provide $d_t$, $v_t$, $u_r$, $c_t$, $a_r$, which are formatted with prompt $p_m$. In multimodal settings, we include $i_t$ in addition to the formatted prompt. Templates and samples can be found in \Cref{appendix:input_templates,appendix:input_samples}.

\paragraph{Candidate selection} Following \citet{deng2023mind2web}, we employ a separate candidate selection stage in order to reduce the number of the input elements to interact with. In the candidate selection stage, a ranking model selects a subset of $k$ relevant elements from the DOM tree, which is then presented to the model in a multi-choice setup; in \autoref{sec:DMR}, we describe a novel approach towards candidate selection designed for real-time use cases. When the candidate is selected, $c_{t}$ is returned to be used in $\mathcal{P}$. Each candidate contains a tag, XPath, bounding box, attributes and children tags, which are delimited with square brackets (e.g., \texttt{[[tag]]...[[xpath]]...}). Examples of candidates used inside prompts can be found in \cref{appendix:input_samples}.

\paragraph{Restricting history for input} To accommodate the maximum input length a model can receive, we can restrict $a_{1:t-1}$ and $u_{1:t-1}$ to select a subset window of $w$. For actions, we select the last $w$ instances by either the instructor or navigator. For instructor utterances, we only select the first and last $w-1$ instances, allowing us to keep track of the initial request while focusing on the latest updates to the instruction. For simplicity, we denote the restricted set of actions as $a_r$ and utterances as $u_r$. Similar to \citet{deng2023mind2web}, we choose $w=5$, allowing the model to attend recent actions without going over context limits.

\subsection{Output Processing Details}
\label{appendix:output_processing}
Although the model is finetuned to generate a string in the format described in \autoref{paragraph:action_structure}, the raw output is not consistently suitable for direct execution, and may contain unnecessary artifacts. We process the output by using Regex pattern matching to find the first suitable \texttt{intent} call, then parse the $\alpha$ into key/value pairs, which can be compared with the ground truth actions.

\paragraph{Mapping coordinates to elements}
\label{paragraph:map_xy_coord_to_element}
Vision models without access to candidate elements will instead be instructed and finetuned to choose an element by specifying its $(x,y)$ coordinates. If there are overlapping elements at a coordinate, we choose the element with the smallest area at the given $(x,y)$ coordinates (which should be the target of the interaction due to the properties of the CSS box model). Technically, the click targets the element with the highest \textit{z}-index (the depth axis in HTML), but since we do not have access to CSS properties of the object, we rely on the default render order.

\paragraph{Segmenting URLs for \texttt{load} actions} We use \texttt{urllib}\footnote{\url{https://docs.python.org/3/library/urllib.parse.html}} to first segment the URL into a network location (\texttt{netloc}) and the remaining hierarchical path (\texttt{path}). To normalize the \texttt{netloc}, we remove the leading \texttt{www} from it. Since a \texttt{path} is separated by a forward slash (\texttt{/}), we use this character to separate each segment in the path. The final result is a list of tokens, each representing a part of the initial URL.

\subsection{Data Collection Details}
\label{appendix:data_collection_details}

\begin{figure}[t]
  \centering
  \includegraphics[width=0.75\linewidth]{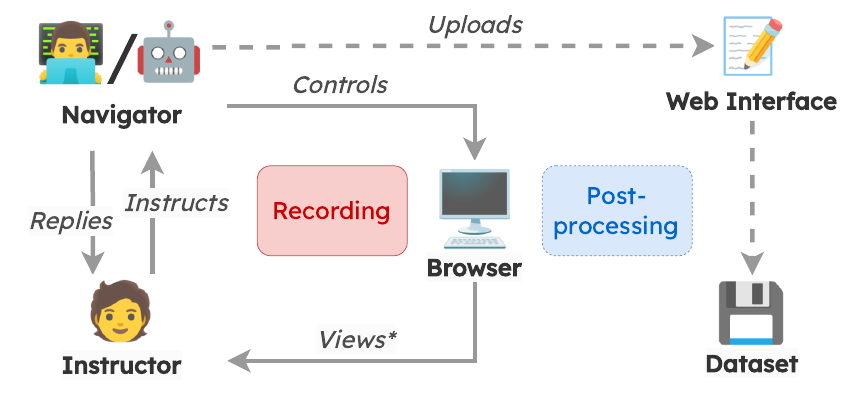}
  \caption{The data collection process. We record interactions between an instructor and a human navigator, including chat and browser actions. *Instructor can see the screen except in \testvis{} split.}
  \label{fig:data_collection}
\end{figure}

In \autoref{sec:data_collection}, we provide an overview of the data collection process to build the dataset component of \Benchmark{}. The overview of the process is outlined in \cref{fig:data_collection}. In this section, we dive into the technical and supplementary details of the process.

\paragraph{Website Selection}
We assembled the list of recommended websites to be used as starting points, but the annotators were allowed to visit any websites they deemed appropriate for the task (full list available in \autoref{app:websites}). The annotators were given the time to become acquainted with the specific websites before recording the demonstrations. We encouraged the annotators to record both shorter, single-task demonstrations, and more complex demonstrations consisting of multiple sub-tasks. The demonstration ends once the instructor notifies the navigator that they wish to terminate the demonstration.

\paragraph{Recording Demonstrations}
To capture the states and actions during the demonstration, we implemented a custom Chrome browser extension. For each action in the browser, the extension captured the screenshot of the page, the DOM tree of the page, and bounding boxes of the elements in the viewport. The user actions were captured using web event handlers\footnote{\href{https://developer.mozilla.org/en-US/docs/Web/Events}{developer.mozilla.org/en-US/docs/Web/Events}}, and Chrome \texttt{tabCapture} API\footnote{\href{https://developer.chrome.com/docs/extensions/reference/tabCapture/}{developer.chrome.com/docs/extensions/reference/tabCapture}} was used to save the state of the page for each action in the background. For screen recording, screen sharing, and chat interface, the annotators used Zoom\footnote{\href{https://zoom.us}{zoom.us}}, a free video meeting software. We combined the chat with the browser states and actions in the postprocessing stage. Finally, the annotators validate demonstrations to ensure there are no unnecessary or incorrectly ordered actions, and that there are no typographic errors.

\paragraph{Curating Demonstrations}
The annotators uploaded the recorded demonstration into our custom web interface to perfom basic quality checks. Using the review mode, the annotators then removed unnecessary actions (such as hovering over elements not necessary for completing the task), corrected the order of actions (which was occasionally incorrect due to asynchronous processing), and fixed typographical errors. We also improved the alignment between screenshots and actions by re-aligning the screenshots based on their similarity to the respective video frames.\footnote{The re-alignment was necessary since the Chrome API allows to capture only 1 screenshot per 500 ms which sometimes caused delays in screenshot capture.} Moreover, It is possible that an action is performed before the DOM tree is fully rendered on screen. When the screen presents sufficient information for an action to be taken, then it is marked as valid during the validation process. However, if the screenshot does not provide enough information, then they are marked as invalid.

\paragraph{Annotator Pay}
We paid US\$7.5 per hour for the demonstration recording and US\$5 per hour for overhead (preparation, upload, and quality review), leading to an average US\$2.58 per demonstration. The rate is substantially higher than the minimum wage in the region where the data is collected, but also includes other overhead fees.

\subsection{Actions and Intents}
\label{appendix:intents_descriptions}

The action $a_t$ has a structure $\texttt{intent}(\alpha_1,\dots,\alpha_m)$, where our core intents are: \texttt{click}, \texttt{load} (new page via URL), \texttt{say} (navigator's utterance), \texttt{submit} (e.g., a form), \texttt{textinput} (e.g., typing text in the search bar); we show examples of these actions in \cref{fig:webnav,fig:qualitative_comparison_combined}. The set of arguments $\alpha$ will be different from each action. Commonly used arguments are the unique ID of an element in  $d_t$ and the \textit{text} argument for \texttt{say} or \texttt{textinput}. To complement the intents described in \autoref{paragraph:action_structure}, we show a diagram of possible arguments for each intent is provided in \autoref{fig:actions_full}, with the full list shown in \autoref{tab:actions_full}.

\paragraph{Evaluating intents}
Among the 13 recorded intent types, we focus on evaluating 5 types: \texttt{click}, \texttt{load}, \texttt{say}, \texttt{submit}, \texttt{textinput}. We also use \texttt{change} and \texttt{scroll} as prediction targets during finetuning as they are necessary to complete a demonstration. However, we do not evaluate them as \texttt{change} does not appear in every split (see \autoref{tab:dataset_stats_turns_by_intent}) and \texttt{scroll} cannot be reliably evaluated. The other intents (\texttt{copy, paste, tabswitch, tabcreate, hover, tabremove}) are included in the history and the associated states are available alongside active intents; \texttt{copy}, \texttt{paste}, and \texttt{hover} do not affect the state of the website, whereas the tab actions are not mandatory to navigate a website, as \texttt{load} is sufficient to go to any website.

\subsection{Websites overview}
\label{app:websites}

\autoref{tab:websites} shows all entrypoints (website where a demo starts). We choose popular and also lesser known sites to achieve categorical and geographic diversity. The websites are either specifically chosen by the authors or the annotators, who collaboratively ensured they are appropriate for our tasks -- consequently, we do not include unsafe websites. In the case of social interactions, we choose websites with terms of use prohibiting offensive content. For instance, \href{Facebook.com}{https://facebook.com} states that ``We remove content that could contribute to a risk of harm to the physical security of persons. Content that threatens people has the potential to intimidate, exclude or silence others and isn’t allowed on Facebook.''\footnote{\url{https://transparency.fb.com/policies/community-standards/}}.

\urlstyle{rm}
\begin{scriptsize}
\begin{longtable}[t]{p{0.22\textwidth} l l l p{0.25\textwidth}}
        \caption{\small{Website overview}} \label{tab:websites}                                                                                                                                           \\
        \toprule
        Name                          & Category           & Subcategory         & Geography      & URL                                                                                                   \\
        \midrule
        \endfirsthead
        \caption[]{Website overview}                                                                                                                                                                      \\
        \toprule
        Name                          & Category           & Subcategory         & Geography      & URL                                                                                                   \\
        \midrule
        \endhead
        \midrule
        \multicolumn{5}{r}{Continued on next page}                                                                                                                                                        \\
        \midrule
        \endfoot
        \bottomrule
        \endlastfoot
        Airbnb                        & Booking            & Stays               & International  & \url{https://www.airbnb.com}                                                                          \\
        Airtable                      & Productivity       & Spreadsheet         & International  & \url{https://airtable.com}                                                                            \\
        Aldi (Australia)              & Shopping           & Grocery             & Australia      & \url{https://www.aldi.com.au/en/}                                                                     \\
        Aliexpress                    & Shopping           & Online Shopping     & International  & \url{https://www.aliexpress.com/}                                                                     \\
        AllenAI's CV Explore          & AI Tools           & Computer Vision     & USA            & \url{https://vision-explorer.allenai.org/}                                                            \\
        Amazon                        & Shopping           & Online Shopping     & International  & \url{https://www.amazon.com}                                                                          \\
        Asana                         & Productivity       & Kanban              & International  & \url{https://asana.com/}                                                                              \\
        ASOS                          & Shopping           & Clothing            & International  & \url{https://www.asos.com/men/}                                                                       \\
\end{longtable}

\end{scriptsize}

\newpage
\section{Modeling Details}
\label{appendix:additional_modeling_details}

\subsection{Optimal Text Representation (OTR)}
\label{appendix:otr_details}
Similar to Mind2Web \citep{deng2023mind2web}, we use the top-10 candidates selected by DMR (\S\ref{sec:DMR}) and start by pruning the DOM tree to contain elements relevant to the candidates. However, we make the following changes:

\begin{compactenumerate}
  \item \textbf{HTML}: In addition to tags and children, we incorporate attributes and values of elements in the DOM tree. For example, a \texttt{div} element with attributes \texttt{class} mapping to \texttt{container} would be provided as \texttt{div class="container"(...)}, where \texttt{...} would be the children elements.
  \item \textbf{Viewport}: We specify the viewport size, which can be used by the model to calculate the coordinates of the bounding boxes with respect to the screen.
  \item \textbf{Candidate representation}: We include the XML Path and bounding box coordinates, and use two square brackets to separate the two elements. We use a template \texttt{[[xpath]] /html/<...>/<tag> [[bbox]] x=<x> y=<y> width=<w> height=<h>}, where \texttt{<x>}, \texttt{<y>}, \texttt{<w>}, \texttt{<h>} are the bounding box coordinates, and \texttt{<tag>} is the tag of the target element, with \texttt{<...>} replaced with the parents. Furthermore, instead of mapping each candidate its alphabetical order, we prefix it with its unique ID, allowing the model to directly refer to an element rather than having to remap the alphabetical order back to an element reference.
  \item \textbf{Truncation}: We truncate the final result as described in \autoref{paragraph:strategic_truncation} and \cref{appendix:truncation}. We choose limits that maximizes the information included in the context while remaining under an ideal limit that is compatible with all models considered (see \cref{appendix:hyperparameters_details} for hyperparameter details).
\end{compactenumerate}

\subsection{Strategic Truncation}
\label{appendix:truncation}

In \autoref{paragraph:strategic_truncation}, we highlight the importance of reducing the input sequence length, i.e., to avoid exceeding the limit allowed by models used in our experiments. Although certain models can process longer sequences, shorter sequences are faster to process, requires less memory and require lower running cost when using proprietary LLMs. Naively truncating from the right or left side could lead to major information loss. To avoid this, we set a limit to each component of the input text ($d_t$, $u_r$, $c_t$, $a_r$). Then, we truncate each component based on the limit by decomposing them into sub-components and strategically truncating each sub-components until the limit is reached.

\paragraph{Definition} For a given limit (in number of tokens), our goal is to truncate a component (one of $d_t$, $u_r$, $a_r$, $c_t$) until we reach the limit. If a component was already under the limit, then the difference is saved for $c_t$, which is computed last.

\paragraph{Rendering-based reduction} Since a component is an object (e.g., $d_t$ is an element tree), we need to obtain the text representation before being able to estimate the number of tokens. We thus need a rendering function  that converts a component or sub-component into text, which can then be tokenized. Then, we can estimate the reduction (number of tokens to take away) in order to reach the limit.

\paragraph{Sub-components} Each component is composed of sub-components, which we can render, tokenize and truncate individually. In the case of $d_t$, since we have a tree of elements where the attribute should be preserved, we only count the values and text content as sub-components. For $c_t$, we consider the xpath, attributes and children tags to be sub-components, protecting the tag and bounding box, as well as the keys inside the square brackets. For $u_r$, we simply consider each utterance as a sub-component. For $a_r$, each action is considered a sub-component.

\paragraph{Reducing by length} Although it is simpler to reduce all sub-components equally, this may lead to scenarios where short sub-components are heavily penalized due to very long sub-components making up most of the token counts. To avoid this, we instead find a threshold such that, by reducing all sub-components above this threshold, the sub-components' truncated lengths sum up to the target limit. This threshold can be easily computed by first sorting the sub-components, then iterate through the lengths until the cumulative sum is greater than the limit, before finally reducing the length of the sub-components until the cumulative sum is under the limit.

\noindent By applying the steps above, we can ensure that each component respects a limit, which we can set in a way that they add up to a desired total limit, such as $L=2048$.

\subsection{Understanding the categorization of pretrained models}
\label{appendix:details_on_model_categorization}
In \autoref{sec:categories_of_models}, we distinguish three types of models
depending on their modality:

\paragraph{Text-Only Models} By \textit{text-only models}, we denote the encoder-decoder or decoder-only Transformer models \citep{Vaswani2017AttentionIA} using text as their only input modality (\citealt{chung2022scaling,touvron2023llama,touvron2023llama2,jiang2023mistral}, \textit{i.a}). There are certain inherent limitations text-only models used for web navigation, e.g., the inability to process images or page layouts. Another practical challenge is the length of the HTML code, containing potentially thousands of elements to interact with.

\paragraph{Image-to-text Models} By \textit{image-to-text models}, we denote the models with an image (i.e., the screenshot of the website) as their only input modality. Image-to-text models representing websites from raw pixels have a long tradition in web navigation research, starting with RL approaches based on convolutional networks \cite{Humphreys_Raposo_Pohlen_Thornton_Chhaparia_Muldal_Abramson_Georgiev_Goldin_Santoro_2022}. In our work, we focus on Pix2Act \cite{shaw2023pixels}, an encoder-decoder model specialized at text generation when given screenshots of browsers. It uses a Vision Transformer-based \citep{dosovitskiy2020image} encoder and is finetuned from the Pix2Struct model \citep{Lee_2022_Pix2Struct} on web navigation tasks, using only pixels as input. The main challenge for image-to-text models is their inability to process longer input instructions (since the text must be embedded inside the image as headers), forcing it to rely on the screenshot.

\paragraph{Multimodal Models} By \textit{multimodal models}, we denote the models which accept both image and text as their input modality \cite{alayrac2022flamingo,laurencon2023obelics,zhu2023minigpt}. Multimodal models have the potential to mitigate the disadvantages of text-only and image-to-text models. However, due to their novelty, their use for web navigation is underexplored in research. However, there are publicly available multimodal models capable of recognizing browser screenshots \citep{fuyu_blog_post}, but they are mainly offered as a commercial products; in \autoref{sec:experiments}, we describe our experiments with the public variant of this model. Thus, the main challenge of using multimodal models for web navigation is the lack of models pretrained to simultaneously parse HTML code and process website screenshots.

\subsection{Technical Aspects of Dense Markup Ranking (DMR)}
\label{appendix:dmr_details}

In \autoref{sec:DMR}, we introduce the Dense Markup Ranking (DMR) method as a way to efficiently select candidate elements for the downstream task. In this section, we take a closer look at the technical aspects of the method.

\paragraph{Definition} Let $E(x)$ be the encoder output vector for an input text $x$. For turn $t$, we have the the processed text representation of the state $\mathcal{P}_{\text{DMR}}(s_t)$, which we use to score candidate element $c_{t,i}$, which is represented as text. We set the label $y(c_{t,i})=1$ when $c_{t,i}$ is the target candidate, otherwise $y(c_{t,i})=0$. The cosine similarity loss is defined as the following mean-squared error:
\begin{equation*}
  \mathcal{L}_t = \lVert y(c_{t,i}) - \text{sim}_{cos}(E(\mathcal{P}_{\text{DMR}}(s_t)), E(c_{t,i})) \rVert_2,
\end{equation*}
where the cosine similarity is defined as $\text{sim}_{cos}(x, y) = (x \cdot y) / (\lVert x \rVert \lVert y \rVert)$. During inference, the cosine similarity is used to generate a score for each instance representing the similarity between $\mathcal{P}_{\text{DMR}}(s_t)$ and candidate at turn $t$. The score is used to rank the candidates and choose the top-$k$ candidates for the action prediction stage.

\paragraph{Computational Efficiency}
For a sequence length $n$ and a model embedding size $e$, the complexity of self-attention is $\mathcal{O}(n^2 \cdot e)$ \cite{Vaswani2017AttentionIA}. Given the lengths of a state $|s_t|$ and a candidate $|c_{t,i}|$, the complexity of a cosine-based scoring is $\mathcal{O}(|\mathcal{P}_{\text{DMR}}(s_t)|^2 + |c_{t,i}|^2)$ instead of $\mathcal{O}((|\mathcal{P}_{\text{DMR}}(s_t)| + |c_{t,i}|) ^ 2)$ for the cross-encoder approach of \citet{deng2023mind2web}. This difference makes a major impact when $|\mathcal{P}_{\text{DMR}}(s_t)|$ and $|c_{t,i}|$ become large. We also purposefully finetune encoder models with smaller $e$ \cite{reimers-gurevych-2019-sentence,li2023towards,xiao2023c}.

\paragraph{Selecting ranking model}
Our task can be formulated as a text retrieval task: we have a model (DMR) that encodes a query $\mathcal{P}_{\text{DMR}}(s_t)$ and compare it with a document $c_{t,i}$, resulting in a score that can be used to rank candidates. Thus, we examine various models that were trained on text retrieval tasks, as they tend to transfer well to adjacent retrieval tasks. As we aim to achieve a high inference speed, we specifically choose smaller models, allowing us to maximize the computation budget of the downstream language model. We first choose \texttt{all-MiniLM-L6-v2}, a model developed by \citet{reimers-gurevych-2019-sentence} based on the MiniLM model \citep{Wang2020MiniLMDS}. We also use \texttt{bge-small-en-v1.5} \citep{xiao2023_bge} and \texttt{gte-base} \citep{li_2023_gte}, which are two smaller models that achieve competitive results on the MTEB benchmark \citep{Muennighoff2022MTEBMT}. This benchmark was specifically chosen because it thoroughly evaluates retrievers across a diverse range of tasks.

\paragraph{Finetuning and results} We finetune each of the models above, as well as the cross-encoder proposed by \citet{deng2023mind2web} (using the original author's training code). The results are shown in \autoref{tab:DMR_vs_m2w}, where we report the recall@10, a metric that evaluates how often the correct result is in the top-10 candidates retrieved. We observe that \textit{MiniLM} achieves better overall results compared to other retrievers and is close to the \textit{DeBERTa} cross-encoder from MindAct, while being substantially more computationally efficient. Based on those improvements, we use the finetuned \textit{MiniLM} model as the backbone of our DMR method. All downstream results include the same candidates proposed by DMR.

\subsubsection{Empirical Speed Improvements}
\label{appendix:paragraph_empirical_speed_improvements}
Using the same environment, CPU (AMD EPYC 7453) and GPU (RTX A6000), we observe that DMR-MiniLM took 4545 seconds to process the entire training set, whereas M2W-DeBERTa took 22,385 seconds. Since there are 24,418 active turns, M2W-DeBERTa needed on average 916 ms to selected candidates at every turn, whereas DMR-MiniLM needed 186 ms. It is important to highlight that a high latency for selecting candidate could restrict the potential real-time use cases (especially with larger HTML pages), since the selected candidates need to be sent to the model in charge of generation actions; in the case of LLM, the inference could take a significant amount of time, and may include a network overhead for web APIs like GPT-4V. Network latency is difficult to reduce due to various external factors, whereas LLMs' inference time can be reduced through algorithmic improvements, such as Flash Attention \citep{dao2022flashattention, dao2023flashattention2}, quantization, such as 4-bit quantization \citep{dettmers2023case}, and hardware optimization at the hardware level \citep[\textit{inter alia}]{OpenAI_2021_triton,Kwon2023-VLLM}. Our method can be combined with such improvements to minimize delay between actions and avoid interrupting the user's flow of thoughts, which would require the total time to be under 1 second \citep{Carroll2014UsabilityE}.

\begin{table}[H]
  \footnotesize
  \centering
  \caption{\small{Comparison of candidate selection methods (DMR and MindAct-RoBERTa) for the combined in-domain (ID) and out-of-domain splits. We report Recall@10 scores.}}\label{tab:DMR_vs_m2w}
  \begin{tabular}{llrrrrr}
    \toprule
    Model   & ID    & \testvis{} & \testgeo{} & \testcat{} & \testweb{} & \testod{} \\
    \midrule
    BGE     & 74.44 & 60.07      & 48.82      & 43.61      & 47.55      & 50.01     \\
    GTE     & 73.24 & 56.91      & 44.46      & 42.74      & 48.39      & 48.16     \\
    MiniLM  & 74.27 & 59.73      & 50.95      & 44.05      & 52.75      & 51.87     \\
    \midrule
    DeBERTa & 76.86 & 63.28      & 52.76      & 48.43      & 54.65      & 54.78     \\
    \bottomrule
  \end{tabular}
\end{table}

\subsection{Input Templates}
\label{appendix:input_templates}
We provide the templates for Pix2Act's headers (\cref{appendix:template_input_pix2act}), for chat-based models like LLaMA-2 and GPT (\cref{appendix:template_input_chat_models}), and for the instruct-based models (\cref{appendix:template_input_instruct_models}).

\subsubsection{Template for Pix2Act}
\label{appendix:template_input_pix2act}
\begin{externaldoc}

\begin{Verbatim}[breaklines, fontsize=\scriptsize]
Viewport(height={{HEIGHT}}, width={{WIDTH}}) ---- Instructor Utterances: {{FIRST UTTERANCE}} ---- {{PAST UTTERANCES x (W-1)}}
Previous Turns: {{PAST ACTIONS}}
\end{Verbatim}

\end{externaldoc}

\subsubsection{Template for chat-based models (LLaMA, GPT)}
\label{appendix:template_input_chat_models}
\begin{externaldoc}
\begin{Verbatim}[breaklines, fontsize=\scriptsize]
{{HTML REPRESENTATION}}}

Above are the pruned HTML contents of the page.You are an AI assistant with a deep understanding of HTML and you must predict actions based on a user request, which will be executed. Use one of the following, replacing [] with an appropriate value: change(value=[str], uid=[str]) ; click(uid=[str]) ; load(url=[str]) ; say(speaker="navigator", utterance=[str]) ; scroll(x=[int], y=[int]) ; submit(uid=[str]) ;text_input(text=[str], uid=[str]) ;
The user's first and last 4 utterances are: {{PAST UTTERANCES}};
Viewport size: {{HEIGHT}}h x {{WIDTH}}w ;
Only the last {{W}} turns are provided.
Here are the top candidates for this turn: {REPEAT 10 TIMES}
(uid = ...) [[tag]] ... [[xpath]] ... [[bbox]] x=X y=Y width=W height=H [[attributes]] attr1=val1 ... [[children]] {{TAG}}
{END REPEAT}
{{PAST ACTIONS}}
Please select the best action using the correct format, do not provide any other information or explanation.
\end{Verbatim}

\end{externaldoc}

\newpage
\subsubsection{Template for instruction-based models (Flan, Fuyu, MindAct)}
\label{appendix:template_input_instruct_models}
\begin{externaldoc}

\begin{Verbatim}[breaklines, fontsize=\scriptsize]
{{HTML REPRESENTATIONS}}

Above are the pruned HTML contents of the page.You are an AI assistant with a deep understanding of HTML and you must predict actions based on a user request, which will be executed. Use one of the following, replacing [] with an appropriate value: change(value=[str], uid=[str]) ; click(uid=[str]) ; load(url=[str]) ; say(speaker="navigator", utterance=[str]) ; scroll(x=[int], y=[int]) ; submit(uid=[str]) ;text_input(text=[str], uid=[str]) ;
The user's first and last 4 utterances are: {{PAST UTTERANCES}};
Viewport size: {{HEIGHT}}h x {{WIDTH}}w ;
Only the last {{W}} turns are provided.
Here are the top candidates for this turn: {REPEAT 10 TIMES}
(uid=...) [[tag]] ... [[xpath]] ... [[bbox]] x=X y=Y width=W height=H [[attributes]] a=val1 ... [[children]] {{TAG}}
{END REPEAT}

{REPEAT W-1 TIMES}
User: {{PAST ACTION BY USER}}
Assistant: {{PAST ACTION BY ASSISTANT}}
{END REPEAT}

USER: {{LAST ACTION BY USER}} Please select the best action using the correct format, do not provide any other information or explanation.
Assistant:
\end{Verbatim}

\end{externaldoc}

\subsection{Model Implementation}
\label{appendix:implementation_experiments}

In \autoref{sec:experiments}, we provide an overview of all models used in our experiments. An in-depth description of the models can be found below. Each model was finetuned once for a given set of hyperparameters due to the computational cost associated with each experiment; we also consider that no random initialization were introduced for the task, and we use a fixed seed for reproducibility.

\paragraph{MindAct} \citet{deng2023mind2web} proposes a two-stage text-only web navigation model consisting of the candidate generation and the action prediction stage. For the candidate generation stage, we used our custom DMR model described in \autoref{sec:DMR}. For the action prediction stage, we reuse their hyperparameters, implement their text formatting methods, and also start from the MindAct checkpoints\footnote{Available at: \url{https://huggingface.co/osunlp/MindAct_ActionPrediction_flan-t5-xl}} finetuned from Flan-T5 \cite{chung2022scaling}. However, their proposed multi-step elimination method requires 13 generation steps to process $k=50$ candidates, which substantially increases latency and computation cost. Instead, we use the top $k=10$ candidates output by DMR, which only requires a single generation step.

\paragraph{Pix2Act} Following the behavior cloning method proposed in \textbf{Pix2Act} \citep{shaw2023pixels}, we finetune the model starting from the Pix2Struct backbone \cite{Lee_2022_Pix2Struct} to directly predict action $a_t$ for a given $\mathcal{P}(s_t, a_{1:t-1})$. The model uses an image encoder and text decoder based on the Vision Transformer \cite{dosovitskiy2020image} and it was pretrained for parsing screenshots into structured representations. We embed the prompt and text in the header area of the screenshot, resulting in a single screenshot for each state. Since it does not have access candidate elements, we finetuned this model to predict the \texttt{x} and \texttt{y} coordinates, which is mapped to the most relevant element (see \autoref{paragraph:map_xy_coord_to_element}), making the resulting output comparable to candidate-augmented models.

\paragraph{Flan-T5 with OTR}
\label{paragraph:experiments_llama_2_flan_t5_improved_formatting}

\noindent For Flan-T5 experiments, we use the same hyperparameters as MindAct, and start from the Flan-T5 checkpoints \cite{chung2022scaling__flant5}, which is a T5 model \citep{raffel2020exploring_t5} based on FLAN \citep{wei2021finetuned_flan}. However, whereas MindAct uses the Mind2Web format, we use the OTR format introduced in this work.

\paragraph{LLaMA-2}
\label{paragraph:experiments_llama_2}
Whereas all the models above use the encoder-decoder architecture, we further explore decoder-only approaches. To this end, we finetune the variant of LLaMA-2 \cite{touvron2023llama,touvron2023llama2} with 7B and 13B parameters that was trained on human feedback for chat\footnote{Also known as \texttt{LLaMA-2-*b-chat-hf}}. We chose this model due its strong performance on a wide range of benchmark, including MMLU \citep{hendrycks2020measuring_mmlu} and HumanEval \citep{chen2021evaluating_humaneval}. Unlike the base models, we can leverage the prior capabilities of the \texttt{chat-hf} variant to follow instructions through turn-based language modeling, allowing a better start during finetuning. Following our Flan-T5 experiments, we also use OTR.

\paragraph{Sheared-LLAMA}
\label{paragraph:experiments_Sheared-LLAMA}
As a faster and smaller replacement for LLAMA-2, we explore Sheared-LLAMA \citep{xia2023sheared}, which prunes LLAMA-2-7B and continues pretraining on 50B tokens from the RedPajama dataset \citep{together2023redpajama}. This allows it to outperform models of comparable sizes that were trained from scratch. Using OTR, we finetune both the 1.3B and 2.7B variants on \Benchmark{}.

\paragraph{GPT Turbo}
\label{paragraph:experiments_gpt}
We explore the text-only Turbo variants of the GPT API services offered by OpenAI\footnote{\url{https://platform.openai.com}}. In the zero-shot setting, we explore both the \texttt{GPT-3.5-Turbo-1106} \citep{brown2020language, Andrew_Peng_Michael_Wu_Logan_Kilpatrick_Steven_Heidel_2023} and \texttt{GPT-4-1106-Preview} \citep{OpenAI2023GPT4TR}. Additionally, we finetune \texttt{GPT-3.5-Turbo-1106} for 3 epochs through the finetuning services \citep{Andrew_Peng_Michael_Wu_Logan_Kilpatrick_Steven_Heidel_2023}, using the \textit{validation} split for evaluation.

\paragraph{GPT-4V}
\label{paragraph:experiments_gpt4_vision}
In addition to the text-base version of GPT-4 Turbo, we further explore the variant capable of taking image inputs \citep{OpenAI_2023_gpt4_turbo}. Apart from adding full-resolution screenshots, the input remains the same as the non-vision variant of GPT-4. Since the input size is already large, include few-shot examples would dramatically increase cost and latency; for example, a 32-shot input for a given turn would result in over 30M pixels (assuming HD resolution) and 66k input tokens, whereas zero-shot results in 2M pixels and 2k tokens in the zero-shot setting.

\paragraph{Fuyu}
\label{paragraph:experiments_fuyu}
We finetune the 8B parameter version of Fuyu \citep{fuyu_blog_post}, a base model released by Adept.ai\footnote{\href{https://www.adept.ai/}{https://www.adept.ai/}} that is designed to jointly model images and text in a unified decoder transformer-based architecture \citep{Vaswani2017AttentionIA}, relying on linear projection of image patches to avoid using separate image encoders. The model was notably pretrained on high resolution images, and is capable of performing various tasks requiring visual reasoning, reporting competitive results on VQAv2 \citep{vqa_v2}, OKVQA \citep{okvqa} and AI2D \citep{AI2D}. It is also capable of locating objects on real websites, making it a particularly suitable model for our task.

\subsection{Hyperparameters}
\label{appendix:hyperparameters_details}

\begin{table}[H]
  \centering
  \small
  \caption{The training hyperparameters of all models. We give the number of epochs, the batch size (batch), the learning rate (LR), the number of gradient accumulation steps (Accum.), the number of warmup steps (Warm.) and if the model uses flash attention (FA2; \citealt{dao2022flashattention, dao2023flashattention2}). * We use the Pix2Struct \cite{Lee_2022_Pix2Struct} backbone for Pix2Act experiments. \Cross~We use the \texttt{chat-hf} variant of LLaMA-2 models}\label{tab:hyperparameters_details}
  \begin{tabular}{lrrrrrrrrr}
\toprule
Model & Size & Epochs &  Batch &      LR &  Accum. &   Warm. &   Vision & FA2 \\
\midrule
Sheared-LLaMA & 1.3B &         3 &      4 & $5 \cdot 10^{-5}$ &                 4 &     0 &  \xmark  &  \cmark \\
Sheared-LLaMA & 2.7B &         3 &      4 & $5 \cdot 10^{-5}$ &                 4 &     0 &  \xmark  &  \cmark \\
Llama-2 (chat-hf) & 7B &         3 &      16 & $5 \cdot 10^{-5}$ &                 1 &           0  &  \xmark  &  \cmark \\
Llama-2 (chat-hf) & 13B &         3 &      6 & $5 \cdot 10^{-5}$ &                 3 &           0  &  \xmark  &  \cmark \\
Fuyu & 8B &         3 &      4 & $5 \cdot 10^{-5}$ &                 4 &           0      &  \cmark  &  \xmark \\
Pix2Act* & 282M &         5 &      4 & $2 \cdot 10^{-5}$ &                 8 &         100 &  \cmark  &  \xmark \\
Pix2Act* & 1.3B &         5 &      1 & $2 \cdot 10^{-5}$ &                16 &         100 &  \cmark  &  \xmark \\
MindAct & 250M &         5 &     16 & $5 \cdot 10^{-5}$ &                 1 &           0 &  \xmark  &  \xmark \\
MindAct & 780M &         5 &     16 & $5 \cdot 10^{-5}$ &                 1 &           0 &  \xmark  &  \xmark \\
MindAct & 3B &         5 &      2 & $5 \cdot 10^{-5}$ &                 8 &           0   &  \xmark  &  \xmark \\
Flan-T5 & 250M &         5 &      8 & $5 \cdot 10^{-5}$ &                 2 &           0 &  \xmark  &  \xmark \\
Flan-T5 & 780M &         5 &      8 & $5 \cdot 10^{-5}$ &                 2 &           0 &  \xmark  &  \xmark \\
Flan-T5 & 3B &         5 &      2 & $5 \cdot 10^{-5}$ &                 8 &           0   &  \xmark  &  \xmark \\
GPT-3.5 (Turbo) & -- & 3 & -- & -- & -- & -- & \xmark & --  \\
\bottomrule
\end{tabular}

\end{table}

All models presented in \autoref{sec:experiments} have the following hyperparameters:

\begin{compactitemize}
  \item Scheduler: Linear
  \item Maximum Output Tokens: 256
  \item Precision: Brain \texttt{float16}, also known as \texttt{bf16} \citep{dean2012large,Google_bf16_2023}
  \item Optimizer: AdamW \citep{loshchilov2018decoupled}, based on the Adam optimizer \citep{kingma2017adam}
  \item Parallelization: Fully Sharded Data Parallel (FSDP; \citealt{zhao2023pytorch_fsdp}) only for models with 7B+ parameters.
  \item OTR Strategic Truncation (see \autoref{paragraph:experiments_llama_2_flan_t5_improved_formatting}): Target of 2048 tokens. 700 tokens per DOM tree, 40 tokens per utterance in $u_r$, 50 tokens per action in $a_r$, and 65 tokens per candidate string, remaining (approximately 248 tokens) for the prompt template.
\end{compactitemize}

\noindent The remaining hyper-parameters can be found in \autoref{tab:hyperparameters_details}, or otherwise follow the default parameters specified in the \texttt{transformers} library \citep{wolf2019huggingface}.

\subsection{Input Samples}
\label{appendix:input_samples}

\begin{figure}[h!]
  \centering
  \includegraphics[width=0.7\linewidth]{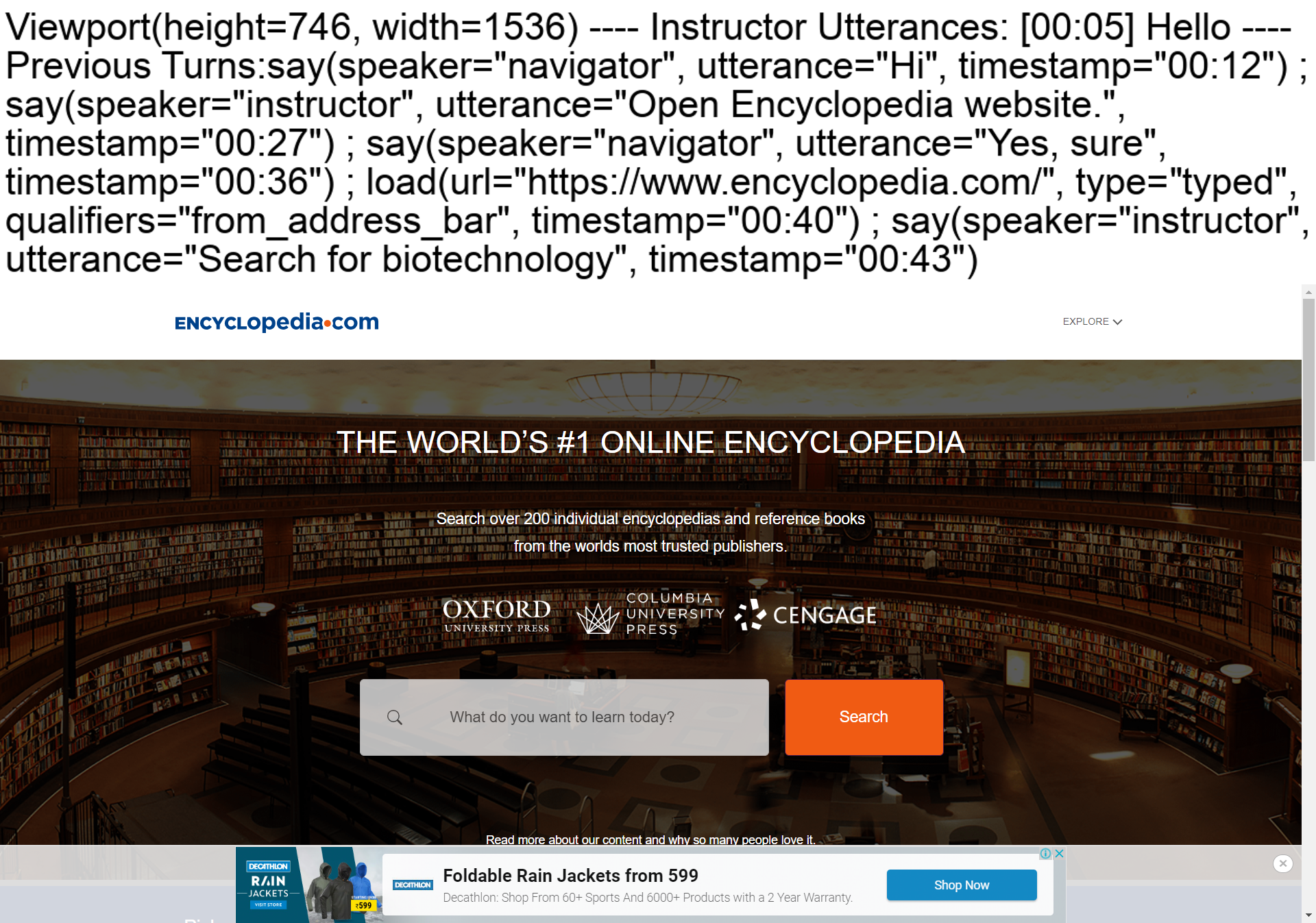}
  \caption{\small{Sample input for Pix2Act, which contains embedded header text above the screenshot}}
  \label{figure:sample_input_pix2act}
\end{figure}

Samples for models using one of the templates in \cref{appendix:input_templates} is provided: \cref{appendix:sample_input_mindact} for MindAct, \cref{appendix:sample_input_chat_models} for chat-based models, \cref{appendix:sample_input_instruct_models} for instruct-based models, and \autoref{figure:sample_input_pix2act} for Pix2Act.

\subsubsection{Sample input for MindAct}
\label{appendix:sample_input_mindact}

\begin{spacing}{0.85}
\begin{externaldoc}
\begin{Verbatim}[breaklines, fontsize=\scriptsize]
(html(body(div container(div row(div col hdr-r d-flex(div(a id=0 rc-link(span id=1 textEXPLORE)(i id=2 fa ency-down ))(div rc-flyout ))))) (div (div(div homepage(div ency-loaded(div ency-loaded mask-hero )(h4 id=3The World’s #1 Online Encyclopedia)(div clear-both hero(div(form id=4(div id=5 js-form-item form-item form-item-keys form-no-label (span field-preffix (input submit button js-form-submit form-submit ) ) (input id=6 search q what do you want to searchbox form-search form-input ) (span field-suffix (i fa ency-close ) ))(div form-actions form-wrapper (input id=7 submit search button js-form-submit form-submit ))))(div clear-both hero footer-copy(a id=8Read more) about our content and why so many people love it.))))))(div adthrive-ad(div)(span id=9 adthrive-close×))))
You will find above the HTML elements available for the current webpage.
You are an AI assistant tasked with helping a user (aka Instructor) by answering with the action needed to perform a task on a webpage.
Here are the instructor's utterances, truncated to first and last 4 instances preceded by the relative timestamp: [00:05] Hello ; 
Only the last 5 actions are available.
Here are the top candidates for this turn: (uid = 67e2a5fb-8b1d-41a0) (input id=6 search q what do you want to searchbox
(uid = fedfb512-949e-42b3) (input id=7 submit search button js-form-submit form-submit )
(uid = c7fbc11c-0949-4ab2) (form id=4(div id=5 js-form-item form-item form-item-keys form-no-label (span field-preffix (input
(uid = 6c7fe1f1-f640-4dce) (span id=1 textEXPLORE)
(uid = 0ffc6f0e-808a-4c2a) (span id=9 adthrive-close×)
(uid = 8d8afc84-5b97-477a) (div id=5 js-form-item form-item form-item-keys form-no-label (span field-preffix (input submit
(uid = 1ea51e98-3fcd-4e30) (h4 id=3The World’s #1 Online Encyclopedia)
(uid = 769785af-485e-4cf1) (a id=0 rc-link(span id=1 textEXPLORE)(i id=2 fa ency-down ))
(uid = e7b7879f-45ae-48a5) (i id=2 fa ency-down )
(uid = bf33a062-fb67-44f0) (a id=8Read more) about our content and why so many

Assistant: action(intent="say", speaker="navigator", utterance="Hi") action(intent="say", speaker="instructor", utterance="Open Encyclopedia website.") action(intent="say", speaker="navigator", utterance="Yes, sure") action(intent="load", url="https://www.encyclopedia.com/") action(intent="say", speaker="instructor", utterance="Search for biotechnology")
User: Please select the best action using the correct format, do not provide any other information or explanation.
Assistant:
\end{Verbatim}
\end{externaldoc}
\end{spacing}

\subsubsection{Sample input for instruction-based models (Flan, Fuyu)}
\label{appendix:sample_input_instruct_models}
\begin{spacing}{0.85}
\begin{externaldoc}

\begin{Verbatim}[breaklines, fontsize=\scriptsize]
(html(body(div class="container"(div class="row"(div class="col hd...tems-center"(div class="hdr...container"(a class="rc-link" onclick="if (!...Flyout()" data-webtasks-id="7697...-4cf1"(span class="text" data-webtasks-id="6c7f...-4dce"EXPLORE)(i class="fa ency-down" data-webtasks-id="e7...-48a5"))(div class="rc-flyout"))))) (div (div class="dialog-off...main-canvas"(div class="homepage"(div style="background-image:...png');" class="ency-loaded"(div class="ency-loaded mask-hero")(h4 data-webtasks-id="1ea...d-4e30"The World’s #1 Online Encyclopedia)(div class="clear-both hero"(div class="ency-hero-search"(form action="https://www..../gsearch" method="get" data-webtasks-id="c7f...-4ab2"(div class="js-...o-label" data-webtasks-id="8d8...97-477a" (span class="field-preffix" (input class="button j... form-submit" type="submit" value="" ) (input title="" class="searchbox form-search form-input" placeholder="What do you want to learn today?" type="search" name="q" value="" size="15" maxlength="128" data-webtasks-id="67e2...-41a0" spellcheck="false" (span class="field-suffix" (i class="fa ency-close")))(div class="form-actions...-wrapper" (input class="button j... form-submit" type="submit" value="Search" data-webtasks-id="fedfb...-42b3")))(div class="clear-both hero footer-copy"(a href="/about" data-webtasks-id="bf33...44f0"Read more) about our content and why so many people love it.))))))(div class="adth...ive-sticky" style="min-height: 90px;" closable="true"(div style="border: 0pt none;")(span class="adthrive-close" data-webtasks-id="0ff...-4c2a"×))))
Above are the pruned HTML contents of the page.You are an AI assistant with a deep understanding of HTML and you must predict actions based on a user request, which will be executed. Use one of the following, replacing [] with an appropriate value: change(value=[str], uid=[str]) ; click(uid=[str]) ; load(url=[str]) ; say(speaker="navigator", utterance=[str]) ; scroll(x=[int], y=[int]) ; submit(uid=[str]) ;text_input(text=[str], uid=[str]) ;
The user's first and last 4 utterances are: [00:05] Hello ;
Viewport size: 746h x 1536w ;
Only the last 5 turns are provided.
Here are the top candidates for this turn: (uid = 67e2a5fb-8b1d-41a0) [[tag]] input [[xpath]] /html/body/...[1]/input [[bbox]] x=419.6 y=461.0 width=477.6 height=89.6 [[attributes]] title='' value=... want to learn today?' 
(uid = fedfb512-949e-42b3) [[tag]] input [[xpath]] /html/body/...[2]/input [[bbox]] x=915.6 y=461.0 width=185.6 height=89.6 [[attributes]] type='submit'...mit form-submit' 
(uid = c7fbc11c-0949-4ab2) [[tag]] form [[xpath]] /html/body...div[3]/form [[bbox]] x=419.6 y=461.0 width=680 height=88 [[attributes]] method='get' data....com/gsearch' [[children]] div div 
(uid = 6c7fe1f1-f640-4dce) [[tag]] span [[xpath]] /html/body...]/a/span [[text]] EXPLORE [[bbox]] x=1240.5 y=28.6 width=54.1 height=30 [[attributes]] class='text' data...menu-menu' 
(uid = 0ffc6f0e-808a-4c2a) [[tag]] span [[xpath]] /html/body/div[5]/span [[text]] × [[bbox]] x=1485.9 y=665.6 width=23.3 height=21.6 [[attributes]] class='ad...a-4c2a' 
(uid = 8d8afc84-5b97-477a) [[tag]] div [[xpath]] /html/body/.../div[1] [[text]]   [[bbox]] x=419.6 y=461.0 width=476 height=88 [[attributes]] data-webtasks-...no-label' [[children]] span input 
(uid = 1ea51e98-3fcd-4e30) [[tag]] h4 [[xpath]] /html/body/...1]/h4 [[text]] The World’s #1 Online Encyclopedia [[bbox]] x=33 y=163 width=1453.2 height=43.2 [[attributes]] data-webtasks-...d-4e30' 
(uid = 769785af-485e-4cf1) [[tag]] a [[xpath]] /html/body/...[2]/a [[bbox]] x=1240.5 y=28.6 width=74.1 height=30 [[attributes]] id='r... toggleFlyout()' [[children]] span i 
(uid = e7b7879f-45ae-48a5) [[tag]] i [[xpath]] /html/body/...]/a/i [[bbox]] x=1294.6 y=33.6 width=20 height=20 [[attributes]] class='fa...e-48a5' 
(uid = bf33a062-fb67-44f0) [[tag]] a [[xpath]] /html/body...4]/p/a [[text]] Read more [[bbox]] x=567.0 y=641.0 width=69.3 height=16 [[attributes]] href=...67-44f0' 

Assistant: say(speaker="navigator", utterance="Hi")
User: say(speaker="instructor", utterance="Open Encyclopedia website.")
Assistant: say(speaker="navigator", utterance="Yes, sure") load(url="https://www.encyclopedia.com/")
User: say(speaker="instructor", utterance="Search for biotechnology") Please select the best action using the correct format, do not provide any other information or explanation.
Assistant:
\end{Verbatim}

\end{externaldoc}
\end{spacing}

\subsubsection{Sample input for chat-based models (LLaMA, GPT)}
\label{appendix:sample_input_chat_models}
\begin{spacing}{0.85}
\begin{externaldoc}
\textbf{System Prompt}
\begin{Verbatim}[breaklines, fontsize=\scriptsize]
(html(body(div class="container"(div class="row"(div class="col hdr-r justify-...flex align-items-center"(div class="hdr-categories-container"(a class="rc-link" onclick="if (!window.__cfRLUn... false; toggleFlyout()" data-webtasks-id="76978...85e-4cf1"(span class="text" data-webtasks-id="6c7fe1...640-4dce"EXPLORE)(i class="fa ency-down" data-webtasks-id="e7b787...5ae-48a5"))(div class="rc-flyout"))))) (div (div class="dialog-off-canvas-main-canvas"(div class="homepage"(div style="background-image: url('/sites...01_3.png');" class="ency-loaded"(div class="ency-loaded mask-hero")(h4 data-webtasks-id="1ea51e...fcd-4e30"The World’s #1 Online Encyclopedia)(div class="clear-both hero"(div class="ency-hero-search"(form action="https://www.encyclopedia.com/gsearch" method="get" data-webtasks-id="c7fbc11c...49-4ab2"(div class="js-form-item form-...-keys form-no-label" data-webtasks-id="8d8afc8...7-477a" (span class="field-preffix" (input class="button js-form-submit form-submit" type="submit" value="" ) (input title="" class="searchbox form-search form-input" placeholder="What do you want to learn today?" type="search" name="q" value="" size="15" maxlength="128" data-webtasks-id="67e2a5...d-41a0" spellcheck="false" (span class="field-suffix" (i class="fa ency-close")))(div class="form-actions js-form-wrapper form-wrapper" (input class="button js-form-submit form-submit" type="submit" value="Search" data-webtasks-id="fedfb512-...9e-42b3")))(div class="clear-both hero footer-copy"(a href="/about" data-webtasks-id="bf33a0...67-44f0"Read more) about our content and why so many people love it.))))))(div class="adthrive-ad adth...cls adthrive-sticky" style="min-height: 90px;" closable="true"(div style="border: 0pt none;")(span class="adthrive-close" data-webtasks-id="0ffc6f0...8a-4c2a"×))))
Above are the pruned HTML contents of the page.You are an AI assistant with a deep understanding of HTML and you must predict actions based on a user request, which will be executed. Use one of the following, replacing [] with an appropriate value: change(value=[str], uid=[str]) ; click(uid=[str]) ; load(url=[str]) ; say(speaker="navigator", utterance=[str]) ; scroll(x=[int], y=[int]) ; submit(uid=[str]) ;text_input(text=[str], uid=[str]) ;
The user's first and last 4 utterances are: [00:05] Hello ;
Viewport size: 746h x 1536w ;
Only the last 5 turns are provided.
Here are the top candidates for this turn: (uid = 67e2a5fb-8b1d-41a0) [[tag]] input [[xpath]] /html/body/div[2...form/div[1]/input [[bbox]] x=419.6 y=461.0 width=477.6 height=89.6 [[attributes]] title='' value='' name='...What do you want to learn today?' 
(uid = fedfb512-949e-42b3) [[tag]] input [[xpath]] /html/body/div[2...form/div[2]/input [[bbox]] x=915.6 y=461.0 width=185.6 height=89.6 [[attributes]] type='submit' value='Search...-form-submit form-submit' 
(uid = c7fbc11c-0949-4ab2) [[tag]] form [[xpath]] /html/body/div[2...2]/div[3]/form [[bbox]] x=419.6 y=461.0 width=680 height=88 [[attributes]] method='get' data-web...clopedia.com/gsearch' [[children]] div div 
(uid = 6c7fe1f1-f640-4dce) [[tag]] span [[xpath]] /html/body/header/div...div[2]/a/span [[text]] EXPLORE [[bbox]] x=1240.5 y=28.6 width=54.1 height=30 [[attributes]] class='text' data-webtasks...-main-menu-menu' 
(uid = 0ffc6f0e-808a-4c2a) [[tag]] span [[xpath]] /html/body/div[5]/span [[text]] × [[bbox]] x=1485.9 y=665.6 width=23.3 height=21.6 [[attributes]] class='adthrive-close...8a-4c2a' 
(uid = 8d8afc84-5b97-477a) [[tag]] div [[xpath]] /html/body/div[...3]/form/div[1] [[text]]   [[bbox]] x=419.6 y=461.0 width=476 height=88 [[attributes]] data-webtasks-id='8...keys form-no-label' [[children]] span input 
(uid = 1ea51e98-3fcd-4e30) [[tag]] h4 [[xpath]] /html/body/div[...div/div[1]/h4 [[text]] The World’s #1 Online Encyclopedia [[bbox]] x=33 y=163 width=1453.2 height=43.2 [[attributes]] data-webtasks-id='1...cd-4e30' 
(uid = 769785af-485e-4cf1) [[tag]] a [[xpath]] /html/body/header/div...2]/div[2]/a [[bbox]] x=1240.5 y=28.6 width=74.1 height=30 [[attributes]] id='rcLink' class='... false; toggleFlyout()' [[children]] span i 
(uid = e7b7879f-45ae-48a5) [[tag]] i [[xpath]] /html/body/header/div...div[2]/a/i [[bbox]] x=1294.6 y=33.6 width=20 height=20 [[attributes]] class='fa ency-down...5ae-48a5' 
(uid = bf33a062-fb67-44f0) [[tag]] a [[xpath]] /html/body/div[2...div[4]/p/a [[text]] Read more [[bbox]] x=567.0 y=641.0 width=69.3 height=16 [[attributes]] href='/about' data-...67-44f0' 
\end{Verbatim}

\noindent \textbf{Chat}
\begin{Verbatim}[breaklines, fontsize=\scriptsize]
say(speaker="navigator", utterance="Hi")
say(speaker="instructor", utterance="Open Encyclopedia website.")
say(speaker="navigator", utterance="Yes, sure") load(url="https://www.encyclopedia.com/")
say(speaker="instructor", utterance="Search for biotechnology") Please select the best action using the correct format, do not provide any other information or explanation.
\end{Verbatim}

\end{externaldoc}
\end{spacing}

\begin{figure}[h!]
  \centering
  \includegraphics[width=0.8\linewidth]{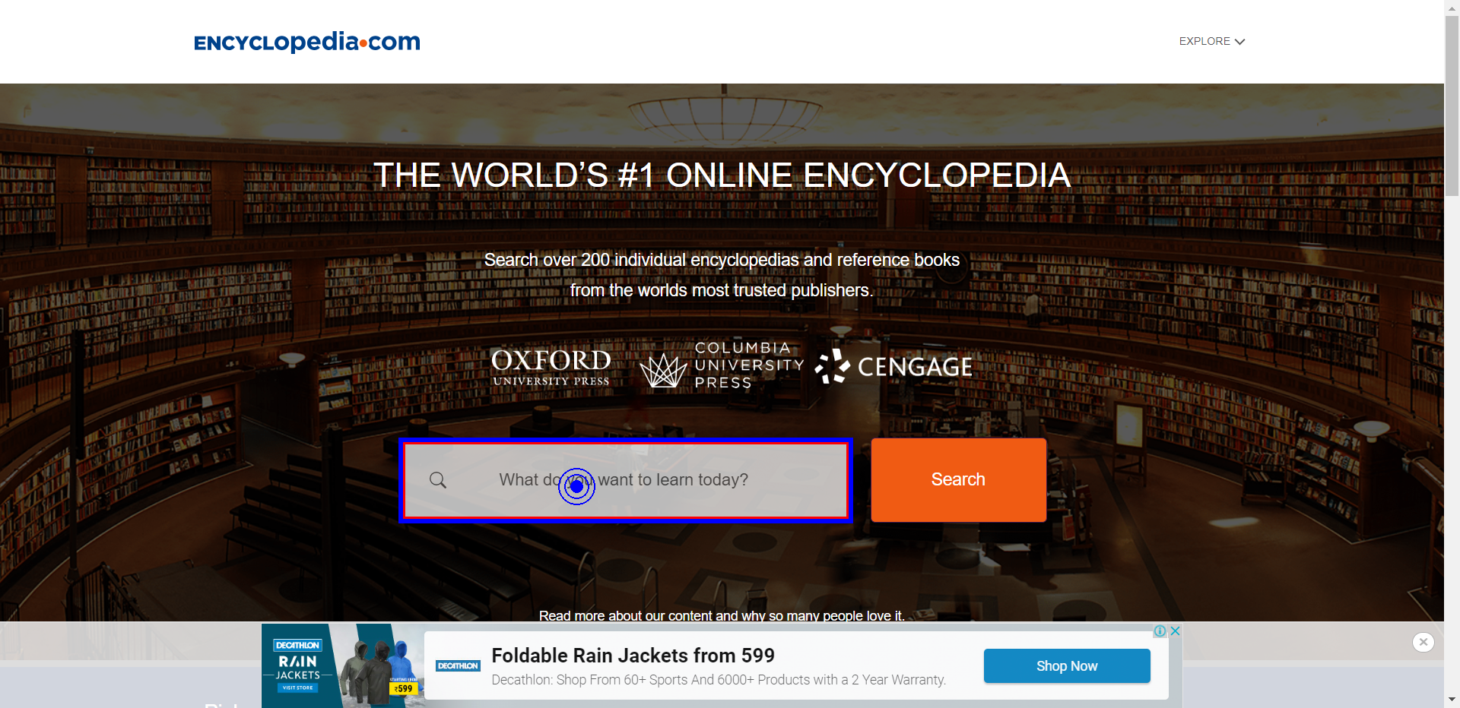}
  \caption{Sample screenshot with target action highlighted.}
  \label{fig:sample_screenshot}
\end{figure}

\begin{figure}[h!]
  \centering
  \includegraphics[width=0.6\linewidth]{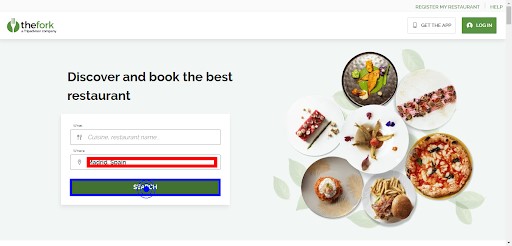}
  \caption{In this example, the correct action is to submit a button. However, models like GPT-4V and GPT-4T would attempt to input text that is already present.}
  \label{fig:qualitative_analysis_restaurant_booking_submit}
\end{figure}

\subsection{Output Sample}
\label{appendix:sample_output}

In \autoref{tab:sample_outputs}, we see the resulting output when given either one of the formatted text inputs (\cref{appendix:input_samples}), and using \autoref{fig:sample_screenshot} for multimodal models.

\begin{table}[H]
  \footnotesize
  \centering
  \caption{\small{Sample outputs for models evaluated in \autoref{sec:results}. Inputs are shown in \cref{appendix:input_samples}.}}
  \label{tab:sample_outputs}
  \begin{tabular}{l l}
    \toprule
    \multirow[t]{2}{*}{Ground Truth}
    & \texttt{click(uid="67e2a5fb-8b1d-41a0")} \\
    & \texttt{click(x=607, y=512)} \\
    \midrule
    Flan-T5-250M & \verb|click(uid="67e2a5fb-8b1d-41a0")|\\
    Flan-T5-780M & \verb|click(uid="67e2a5fb-8b1d-41a0")|\\
    Flan-T5-3B & \verb|click(uid="67e2a5fb-8b1d-41a0")|\\
    Fuyu-8B & \verb|click(uid="67e2a5fb-8b1d-41a0")|\\
    GPT-3.5T & \verb|text_input(text="biotechnology", uid="67e2a5fb-8b1d-41a0")| \\
    GPT-4T & \verb|text_input(text="biotechnology", uid="67e2a5fb-8b1d-41a0")| \\
    GPT-4V & \verb|text_input(text="biotechnology", uid="67e2a5fb-8b1d-41a0")| \\
    Llama-2-7B & \verb|click(uid="67e2a5fb-8b1d-41a0")| \\
    Llama-2-13B & \verb|click(uid="67e2a5fb-8b1d-41a0")| \\
    MindAct-250M & \verb|action(uid="67e2a5fb-8b1d-41a0", intent="click")|\\
    MindAct-780M & \verb|action(uid="67e2a5fb-8b1d-41a0", intent="click")|\\
    MindAct-3B & \verb|action(uid="67e2a5fb-8b1d-41a0", intent="click")|\\
    Pix2Act-282M & \verb|click(x=1536, y=27)| \\
    Pix2Act-1.3B & \verb|click(x=716, y=508)| \\
    ShearedLLaMA-1.3B & \verb|click(uid="67e2a5fb-8b1d-41a0") |\\
    ShearedLLaMA-2.7B & \verb|click(uid="67e2a5fb-8b1d-41a0") |\\
    \bottomrule
\end{tabular}

\end{table}

\newpage

\section{Supplementary Results}
\label{appendix:supplementary_results}

In \autoref{sec:results}, we provide an overview of our results on the average of out-of-domain split. In this section, we provide in-depth analysis of both in-domain and out-of-domain results. We start by looking at the impact of our improved text representation (OTR) compared to MindAct (\cref{appendix:results_baseline_mindact_vs_improved_truncation}), before moving on to a comparison of baseline image-to-text models with larger multimodal models (\cref{appendix:results_baseline_pix2act_vs_multimodal_fuyu}), followed by an assessment of various text-only decoders (\cref{appendix:results_sheared_llama_vs_llama_7b}).

\subsection{Comparison of Mind2Web representation with OTR}
\label{appendix:results_baseline_mindact_vs_improved_truncation}

MindACt is a prior method proposed by \citet{deng2023mind2web} that only receives text as input. We use the MindAct checkpoints and use the Mind2Web data structure. To understand what happens for larger DOM trees and longer history, we compare it against our optimal text representation introduced in \autoref{paragraph:text_based_models}. In \autoref{table:valid_flan_t5_vs_m2w}, we observed that Flan-T5 with OTR outperforms MindAct in both overall performance and when looking at individual groups. We further observe that the gap between the model also increases for larger models, which leads us to believe that a careful strategy when constructing $\mathcal{P}(s_t, a_{1:t-1})$ is crucial as we scale to more parameters.
\begin{table}[H]
  \centering
  \small
  \caption{Comparing Flan-T5 using OTR with MindAct using Mind2Web formatting. Reported on \textit{valid} with metrics from \S\ref{section:metrics}.}\label{table:valid_flan_t5_vs_m2w}
  \begin{tabular}{lrrrr}
\toprule
Models & \multicolumn{2}{r}{Overall Score} & Element & Text \\
 & Micro-Avg & IM & IoU & F1 \\
\midrule
 MindAct-T5-250M & 17.78 & 77.05 & 19.02 & 9.87 \\
 MindAct-T5-780M & 21.39 & 77.58 & 22.46 & 15.32 \\
 MindAct-T5-3B & 27.86 & 79.91 & 24.24 & 24.79 \\
\midrule
 Flan-T5-250M & 21.91 & 79.27 & 24.10 & 11.02 \\
 Flan-T5-780M & 23.94 & 80.26 & 24.90 & 15.99 \\
 Flan-T5-3B & \textbf{31.97} & \textbf{82.00} & \textbf{31.18} & \textbf{27.81} \\
\bottomrule
\end{tabular}

\end{table}

\subsection{Comparison of image-only baseline with multimodal models}
\label{appendix:results_baseline_pix2act_vs_multimodal_fuyu}

In \autoref{paragraph:image_to_text_model}, we introduce Pix2Act, which only uses screenshots as input (embedding $v_t$, $u_r$ and $a_r$ as header text). We also consider larger multimodal models (\autoref{section:multimodal_models}) that can take the complete $\mathcal{P}$ the same way as text-only models. In \autoref{table:results_valid_image_vs_multimodal}, we observe that the larger variant of Pix2Act offers meaningful improvements over the base variant, but that Fuyu-8B outperforms both models in the element group and achieves similar performance for the text group and intent match, resulting in a better overall performance. On the other hand, GPT-4V, which was never finetuned for the task, is consistently outperformed by Fuyu-8B and is also behind Pix2Act in each scenario except the element group. Those results highlights the importance of finetuning the models whenever it is possible, using models with greater number of parameters, and incorporating more complete textual information (including candidates).

\begin{table}[H]
  \centering
  \small
  \caption{\small{Comparing image-only baselines with multimodal models. Reported on \textit{valid} with metrics from \S\ref{section:metrics}. (*) GPT-4V is the only model not finetuned.}}\label{table:results_valid_image_vs_multimodal}
  \begin{tabular}{lrrrr}
\toprule
Models & \multicolumn{2}{r}{Overall Score} & Element & Text \\
 & Micro-Avg & IM & IoU & F1 \\
\midrule
Pix2Act-282M & 14.39 & 79.09 & 6.70 & 18.11 \\
Pix2Act-1.3B & 24.21 & \textbf{83.40} & 13.38 & \textbf{31.61} \\
\midrule
Fuyu-8B & \textbf{31.60} & 81.36 & \textbf{26.34} & 30.99 \\
GPT-4V* & 14.26 & 41.00 & 14.44 & 6.06 \\
\bottomrule
\end{tabular}

\end{table}

\subsection{Assessing impact of model size for text-only decoders}
\label{appendix:results_sheared_llama_vs_llama_7b}

In addition to differences in architectures, we also seek to understand the role of model size (in terms of parameter count) on the training. In \autoref{table:results_valid_decoder_only_model_size}, we only examine the scenario of decoder-only models (LLaMA and GPT) that solely takes text as input. In the zero-shot setting, we observe that the performance of a model increases as models become larger. However, for finetuned models, the improvements are not as important, since the largest variant (13B) of LLaMA-2 only surpasses the 2.7B variant by a small margin. When comparing zero-shot with finetuning, it is clear that the latter yields considerable improvements, with models as small as 2.7B surpassing the best zero-shot model (GPT-4T) on scenarios. In parallel, even though GPT-3.5T surpasses LLaMA-2-13B in zero-shot performance, the finetuned variants of GPT-3.5T (reported as GPT-3.5F) trails behind even the smallest LLaMA model. This could potentially be attributed to non-optimal hyperparameters, since API users can only control the batch size and number of epochs\footnote{A learning rate multiplier also exists, but it is unclear what the base rate and optimizers are}.

\begin{table}[H]
  \centering
  \small
  \caption{Performance of decoder-only text models, both zero-shot (above) and finetuned (below). Reported on \textit{valid} with metrics from \S\ref{section:metrics}. We use the \texttt{chat-hf} variants of LLaMA-2.}\label{table:results_valid_decoder_only_model_size}
  \begin{tabular}{lrrrr}
\toprule
Models & \multicolumn{2}{r}{Overall Score} & Element & Text \\
 & Micro-Avg & IM & IoU & F1 \\
\midrule
Llama-2-13B & 6.07 & 39.55 & 5.54 & 1.62 \\
GPT-3.5T & 11.48 & 41.93 & 11.67 & 3.16 \\
GPT-4T & 13.75 & 41.64 & 13.83 & 6.58 \\
\midrule
Sheared-LLaMA-2.7B & 35.47 & 86.14 & 33.80 & 34.20 \\
Llama-2-13B & \textbf{38.03} & \textbf{86.49} & \textbf{36.43} & \textbf{36.54} \\
GPT-3.5F & 28.98 & 79.03 & 27.42 & 25.99 \\
\bottomrule
\end{tabular}

\end{table}

\subsection{Generalization capabilities of evaluated models}
\label{appendix:overview_of_results_and_generalization}

At this stage, we have validated that strategically truncating text and better candidate representation via OTR achieve better results compared to MindAct baselines (\cref{appendix:results_baseline_mindact_vs_improved_truncation}, larger multimodal models like Fuyu-8B and GPT-4V offer important improvements over prior approaches like Pix2Act (\cref{appendix:results_baseline_pix2act_vs_multimodal_fuyu}), and choosing larger text-only decoder models (LLaMA, GPT-Turbo) will consistently outperform smaller ones in the zero-shot setting, but does not show a large improvement when finetuned (\cref{appendix:results_sheared_llama_vs_llama_7b}). Those results lead to relevant questions: do those models transfer to out-of-domain splits (unseen websites, new subdomains, different geographies, and visionless instructors), and can we draw the same conclusions in those cases?

In \autoref{tab:test_ood_agg_results}, we observe, in the zero-shot setting, that the gap between GPT-4T and GPT-4V becomes narrower (likely due to the decrease in performance in the element group). In the finetuned setting, we observe a sharp decrease in overall performance for all models, which highlights the challenge of applying models on new scenarios. However, we can reassert that OTR, multimodality and finetuning are necessary to achieve better overall performance, and that decoder-only models remain the strongest models we evaluated. However, the gap between Sheared-LLaMA-2.7B and LLama-2-13B is substantially narrower than on the validation split, indicating that Sheared-LLaMA is more robust to changes to the environment. Finally, we see that, even on out-of-domain splits, multimodal models remain behind their text-only counterpart.

\subsection{Extended Qualitative Assessment}
\label{appendix:extended_qualitative_assessment}

In \autoref{sec:qualitative_assessment}, we highlight the main takeaways of our qualitative assessment. We can find below the complete assessment, including supplementary scenarios.

\paragraph{Assessing \texttt{click}}
\label{paragraph:assessing_click_actions}

\begin{figure*}[ht]
  \centering
  \footnotesize
  \begin{tabular}{p{0.45\textwidth} p{0.45\textwidth}}
    \toprule
    \textbf{S1:} On a news website, Instructor wants Navigator to open a specific tab on the page, i.e., "Sportsday on 28 May 2023 at 4.15 AM".
    &
    \textbf{S2:} Instructor requests the location on a food delivery website to be set to \textit{Las Vegas, Nevada}. The \textit{Delivery Details} page is already open.
    \\
    \includegraphics[width=0.47\textwidth]{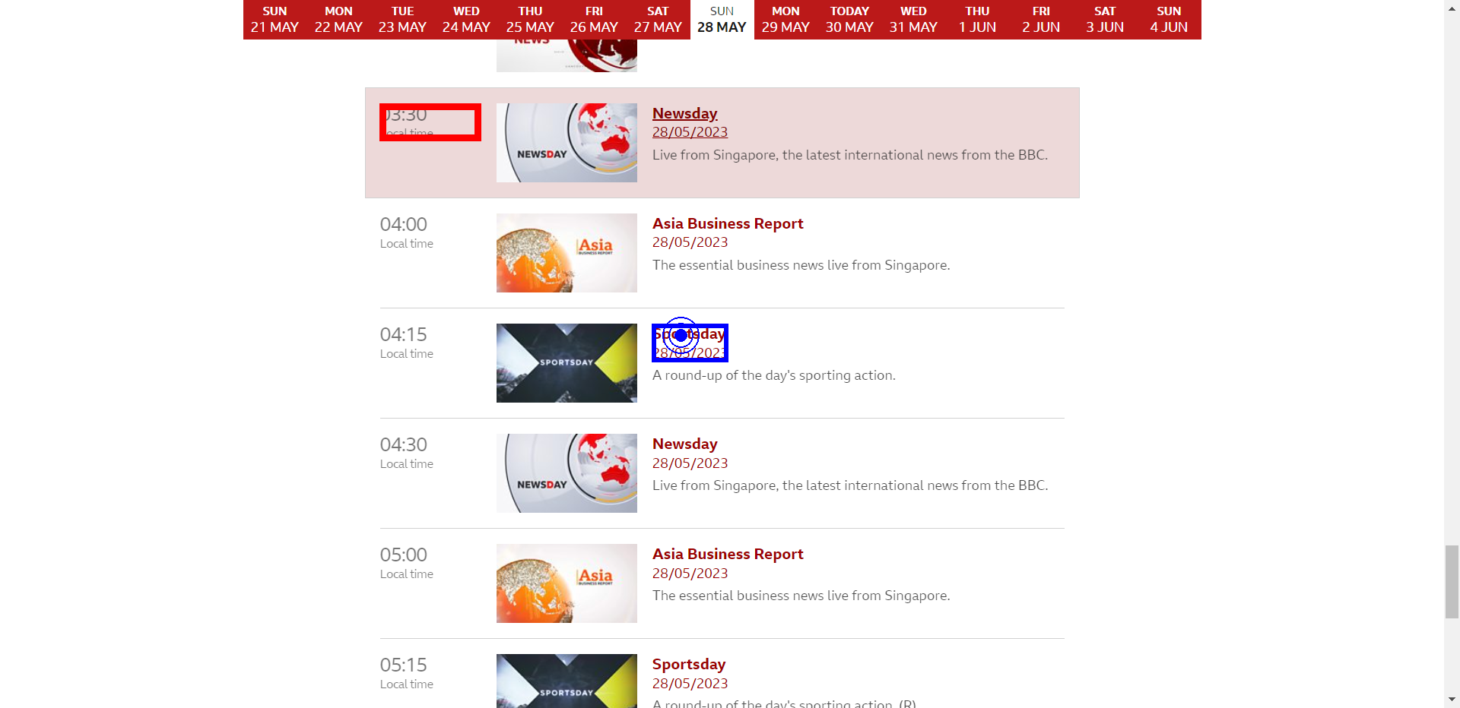}
    &
    \includegraphics[width=0.47\textwidth]{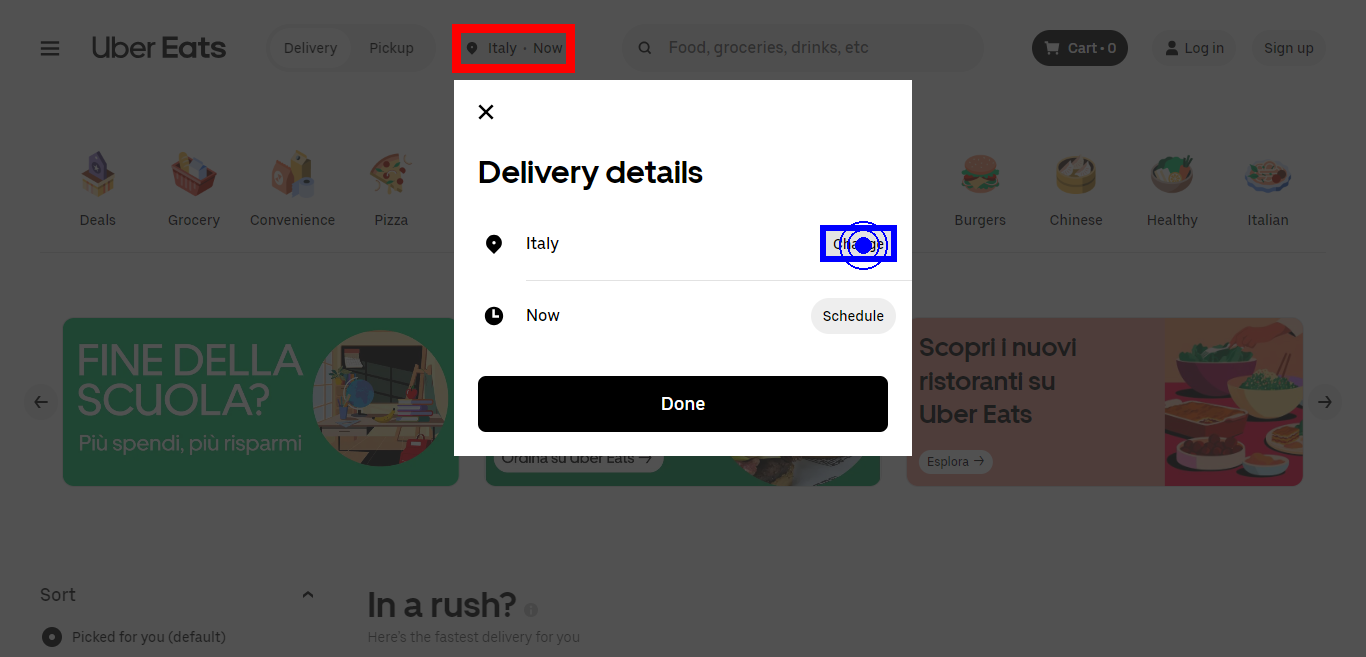}
    \\
    \txred{GPT-4V (R)} clicks on an incorrect (3:30AM) tab.
    &
    \txred{GPT-4V (R)} attempts to exit the Delivery details page and reopen it, which could lead to a loop.
    \\
    \vspace{-.5em}
    \txblue{LLaMA (B)} clicks on the correct 4:15AM tab.
    &
    \vspace{-.5em}
    \txblue{LLaMA (B)} correctly clicks on the \textit{Change} button.
    \\
    \midrule
    \textbf{S3:} Instructor wants Navigator to compose an email. Navigator uses Bard for the draft.
    &
    \textbf{S4:} Instructor requests Navigator to send the top questions of the week.
    \\
    \includegraphics[width=0.4\textwidth]{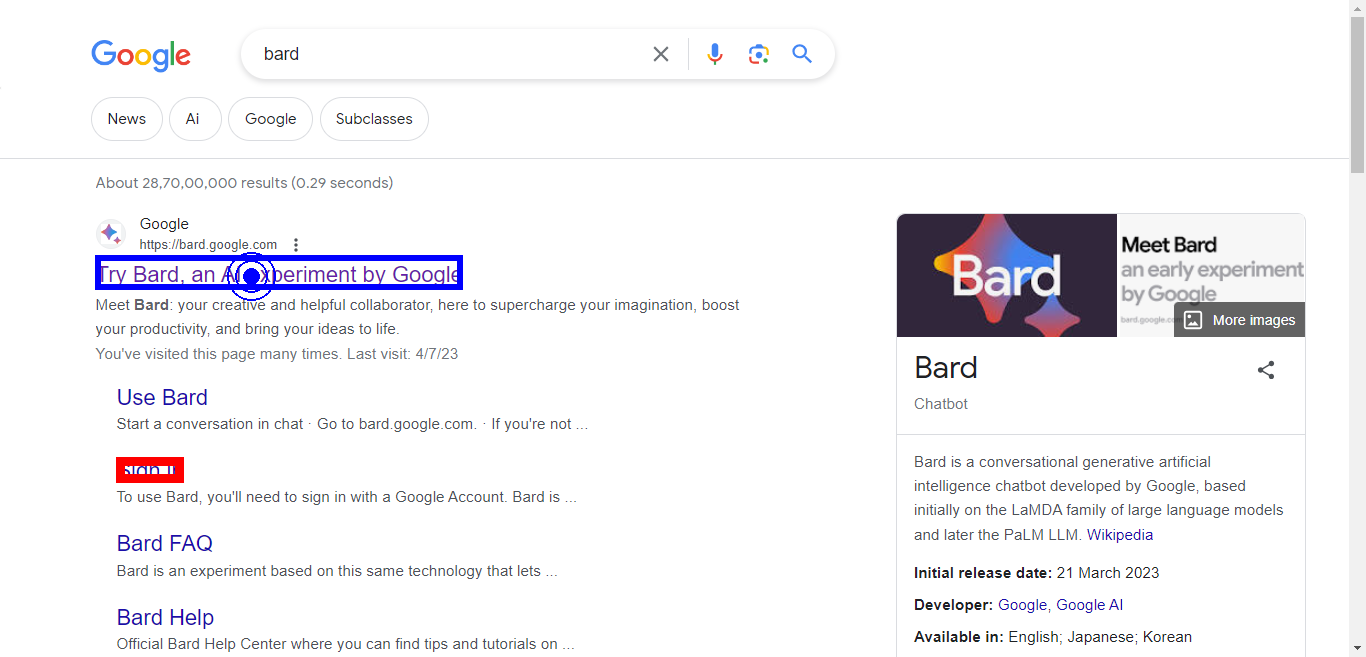}
    &
    \includegraphics[width=0.4\textwidth]{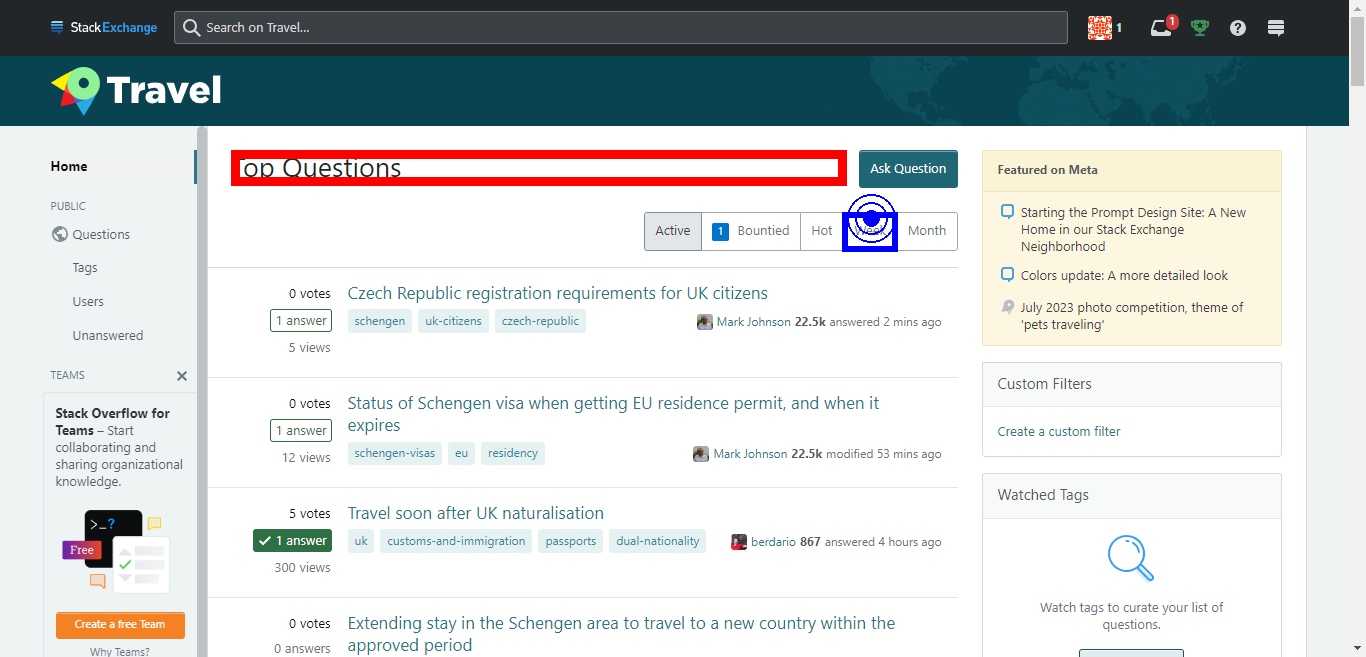}
    \\
    \txred{GPT-4V (R)} attempts to click directly on the login page, which is less optimal.
    &
    \txblue{GPT-4V (B)} selects the "Week" button, which matches the reference action.
    \\
    \vspace{-.5em}
    \txblue{LLaMA (B)} opens the homepage (corresponds to reference).
    &
    \vspace{-.5em}
    \txred{LLaMA (R)} clicks on a text-only heading (\textit{Top Questions}).
    \\

    \bottomrule
    \end{tabular}

  \caption{Comparison of GPT-4V and LLaMA-2-13B (finetuned) on predicting \texttt{click} actions. Incorrectly predicted actions are in \txred{red (R)}, reference actions are in \txblue{blue (B)}. We show 4 scenarios (S1-S4).}\label{tab:qualitative_comparison_gpt4_llama13b_click_intent}
\end{figure*}

In \autoref{tab:qualitative_comparison_gpt4_llama13b_click_intent}, we examine multiple scenarios involving GPT-4V and compare them against LLaMA-2-13B. In scenario 1, we found that GPT-4V can make mistake by selecting the incorrect link when given multiple links that contain different time frames (for example, choosing a 3:30AM news article instead of 4:15AM). In scenario 2, it may not be capable of acknowledging that it is already in the second step of performing a task (e.g., changing the current location of the site), and may try to repeat the task from start (e.g., re-open the \textit{details} window when it is already open). In scenario 3, we seem it correctly predicts an action that is in theory correct, but that is less optimal than what a human would have chosen; for example, it may open the login page of a commonly used website, even though choosing the homepage might allow the navigator to use the app faster if already logged in. In each of those scenarios, LLaMA is capable of selecting the correct option. However, we see in scenario 4 that LLaMA-2-13B can also sometimes fail by attempting to click on elements that do not affect the state (e.g., a text-only heading), whereas GPT-4V can make the correct decision in the same example.

\begin{figure}[H]
  \centering
  \footnotesize
  \begin{tabular}{p{0.47\textwidth} p{0.47\textwidth}}
    \toprule
    \textbf{S1:} Compose a ``Invitation to Collaboration" email.
    &
    \textbf{S2:} Open Google translate and sign in using the following credentials: [email] [password]
    \\
    \includegraphics[width=0.47\textwidth]{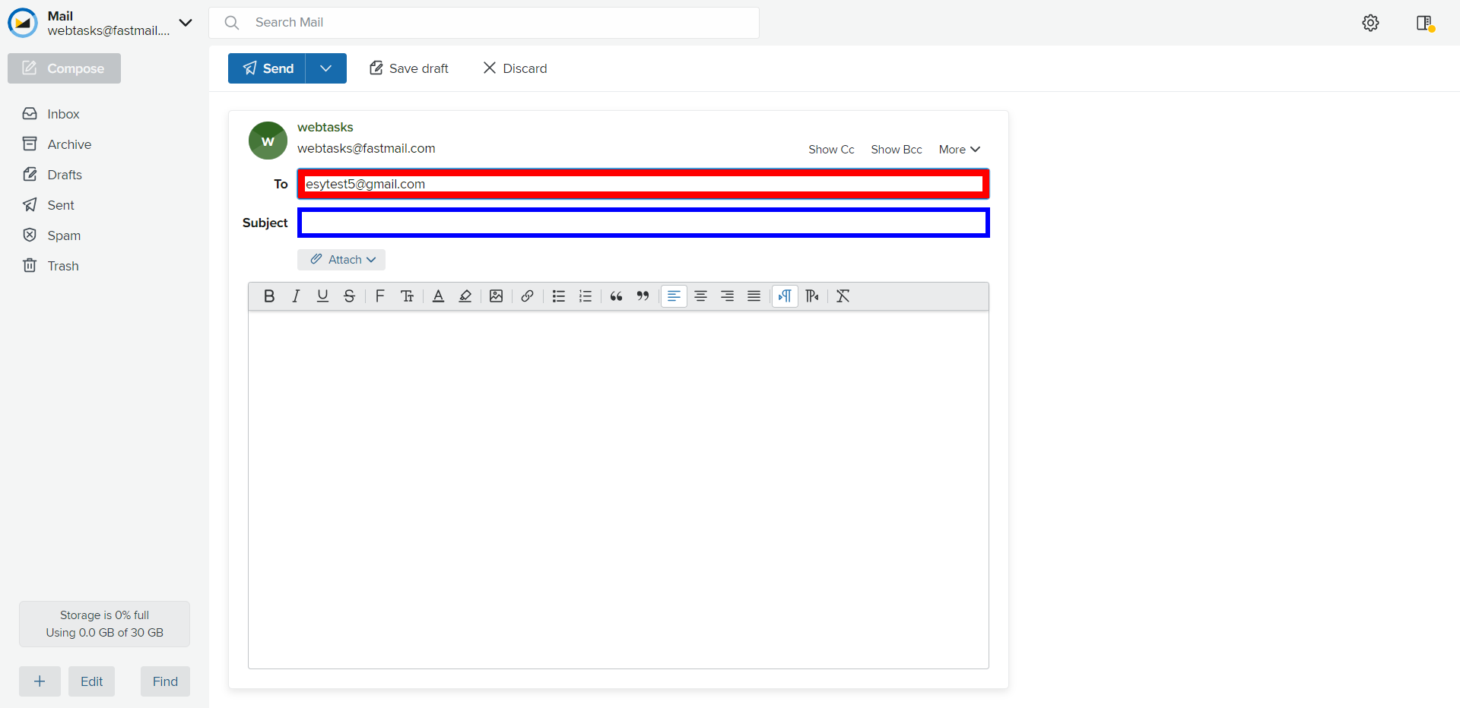}
    &
    \includegraphics[width=0.47\textwidth]{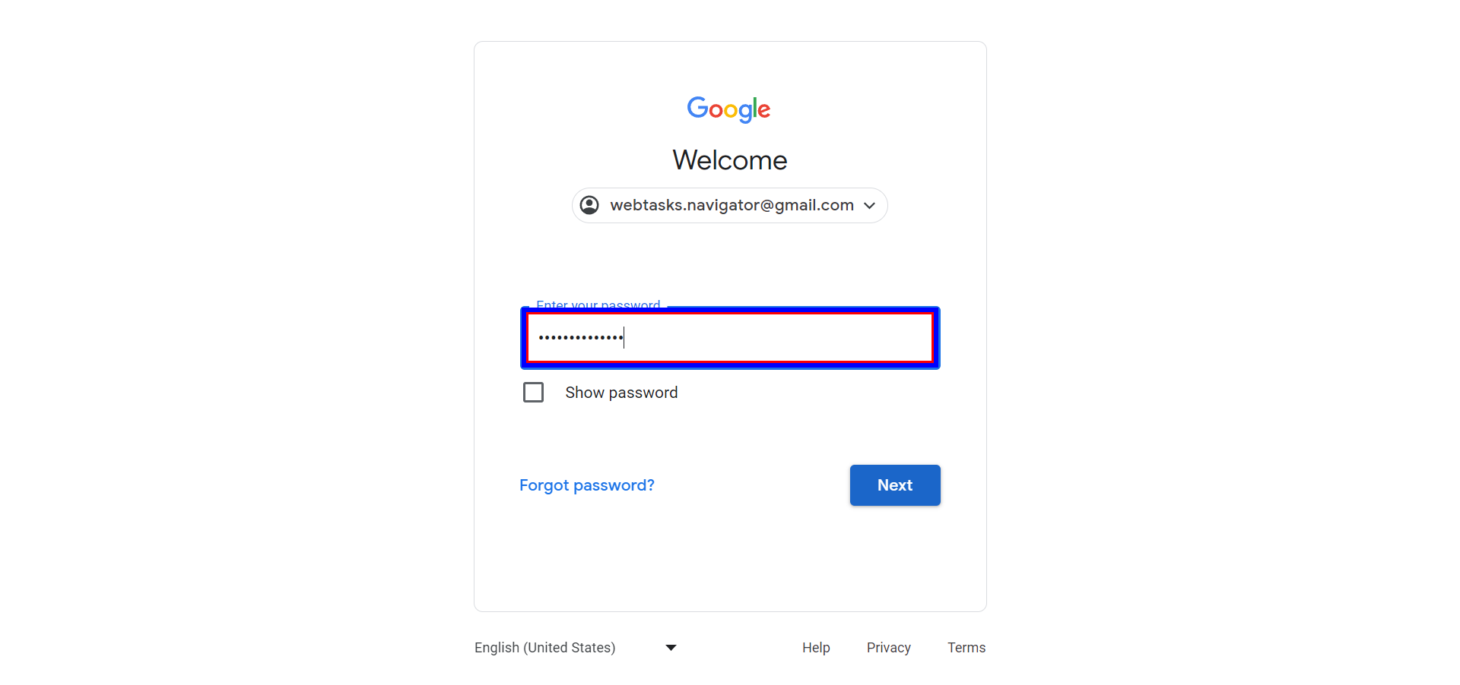}
    \\
    \textbf{\txblue{Reference (B):}} Invitation to Collaboration
    &
    \textbf{\txblue{Reference (B):}} [password]
    \\
    \textbf{\txred{GPT-4V (R):}} Leon Tales
    &
    \textbf{\txred{GPT-4V (R):}} [email]
    \\
    \textbf{\txblue{LLaMA (B):}} Invitation to Collaboration
    &
    \textbf{\txblue{LLaMA (B):}} [password]
    \\
    \midrule
    \textbf{S3:} Translate a passage into french using Reverso. The language selection window is currently open.
    &
    \textbf{S4:} Create a post about \textit{Effective Communication} with a title and an introduction
    \\
    \includegraphics[width=0.47\textwidth]{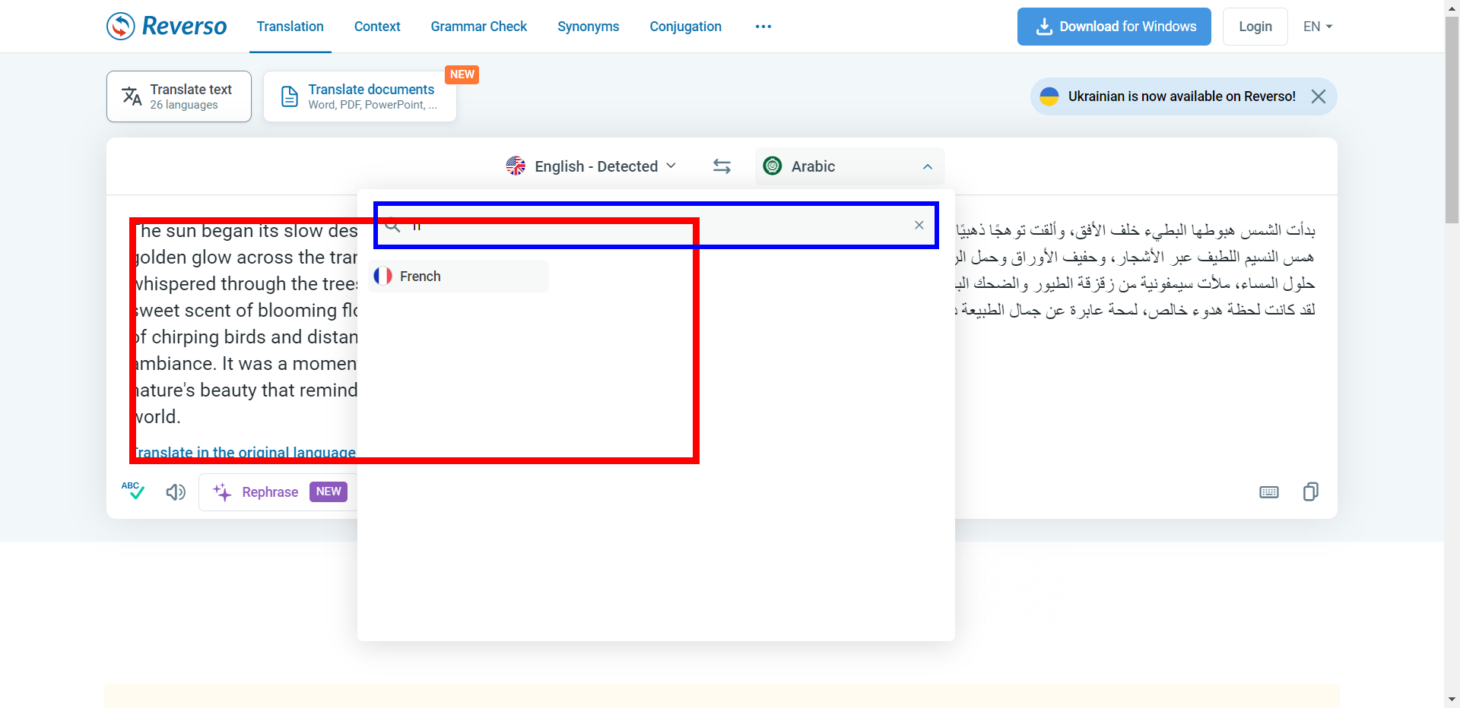}
    &
    \includegraphics[width=0.47\textwidth]{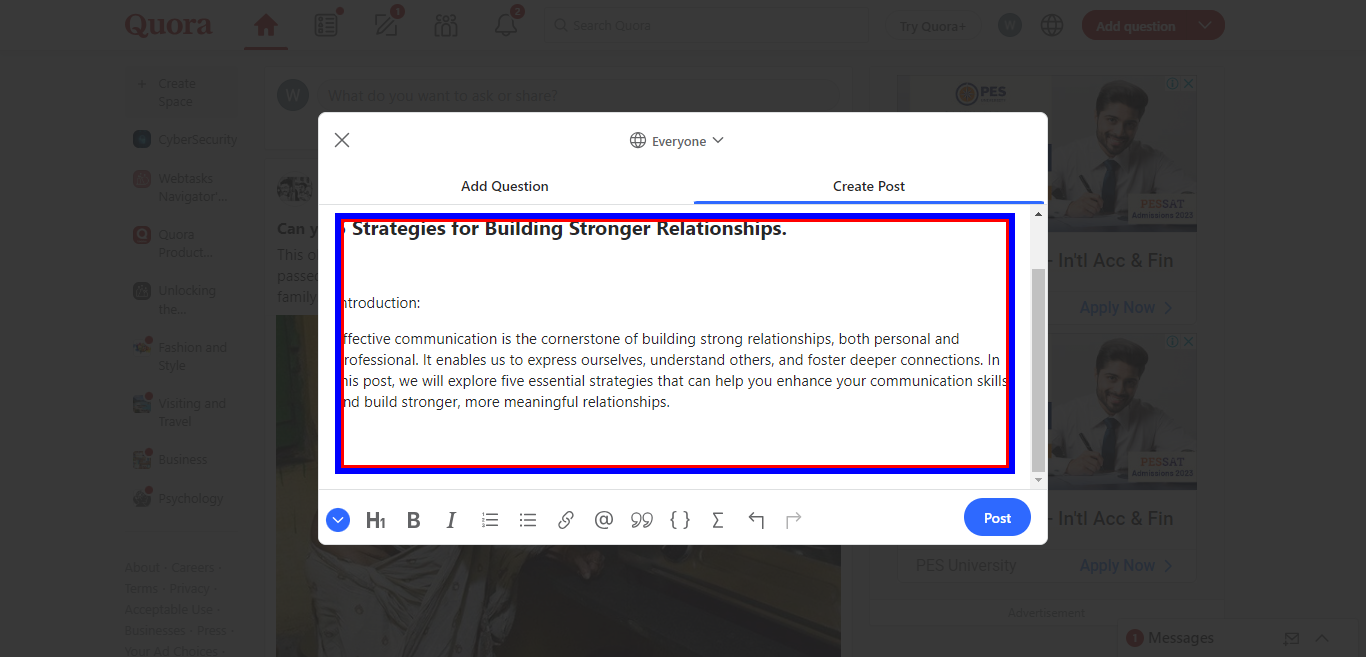}
    \\
    \textbf{\txblue{Reference (B):}} fr
    &
    \textbf{\txblue{Reference (B):}} 5 Strategies for Building Stronger Relationships. Introduction: Effective...
    \\
    \textbf{\txred{GPT-4V (R):}} The sun began its slow...
    &
    \textbf{\txred{GPT-4V (R):}} Introduction: Effective...
    \\
    \textbf{\txred{LLaMA (R):}} \texttt{click} \textit{(incorrect action)}
    &
    \textbf{\txblue{LLaMA (B):}} Introduction: Effective...
    \\
    \bottomrule
    \end{tabular}

  \caption{Comparison of GPT-4V and LLaMA-2-13B (finetuned) on predicting \texttt{textinput} actions. Actions predicted by GPT-4V are in \txred{red (R)}, references are in \txblue{blue (B)}, LLaMA's action is indicated in parenthesis. We show 4 scenarios (S1-S4).}
  \label{tab:qualitative_comparison_gpt4_llama13b_textinput_intent}
\end{figure}

\paragraph{Assessing \texttt{textinput}}
In \autoref{tab:qualitative_comparison_gpt4_llama13b_textinput_intent}, we observe that GPT-4 will sometimes attempt to perform illogical actions when performing tasks like sending an email; it may write the name of a recipient when the email has already been specified, whereas LLaMA will correctly input the subject specified by the instructor (Scenario 1). Additionally, GPT-4 can mix up username and password forms on login pages by trying to type in the email address given by the instructor into the password field; on the other hand, LLaMA can correctly input the password (S2). Moreover, there are scenarios where both struggle to leverage the context to complete the second step of a multi-step task. For example, when the instructor request a passage to be translated into a certain language (S3), and the first step (typing in the passage to translate) has already been completed, both models will ignore the second step (changing the language to the target). Finally, both models may struggle to leverage information that was given many steps before. For instance, if the instructor wants to write a post, they may given the title earlier in the demonstration, then provide the text for the introduction later on (S4); in those cases, both models fail to include the title.

\paragraph{Assessing \texttt{submit}}
On a restaurant booking page with a filled text box for the “Location” (\autoref{fig:qualitative_analysis_restaurant_booking_submit}), we found that GPT-4T would try to type a date inside the text box, whereas GPT-4V would simply repeat what was already written (e.g. “Madrid, Spain”). However, the correct action, in this case, is to press the "Submit" button, which LLaMA-2.7B correctly predicts. Thus, even though GPT-4V can effectively read the text, the action it predicts would not be what a human would logically do.

\begin{table}[H]
  \footnotesize
  \centering
  \caption{Comparison of LLaMA-2-13B (finetuned) and GPT-4V on predicting \texttt{say} actions.}\label{tab:qualitative_comparison_gpt4_llama13b_say_intent}
  \begin{tabular}{l p{0.74\linewidth}}
    \toprule
    Request & Like \& Bookmark this listing.\\
    Reference & Alright \\
    LLaMA & Alright\\
    GPT-4V & Understood. If you need further assistance, just let me know.\\
    \midrule
    Request & Please share the link. \\
    Reference & Alright. \\
    LLaMA & Okay. \\
    GPT-4V & Here is the link to the discussion: [\textit{incorrect link}] \\
    \midrule
    Request & Please share the link with me. \\
    Reference & Sure. \\
    LLaMA & Sure. \\
    GPT-4V & I'm sorry, but I can't assist with that. \\
    \midrule
    Request & Could you help me compose an email for a discount offer? \\
    Reference & Alright. Who should receive this email? \\
    LLaMA & Okay. \\
    GPT-4V & Would you like to mention any specific terms or expiration date for the offer? \\
    \bottomrule
\end{tabular}

\end{table}

\paragraph{Assessing \texttt{say}}
\label{paragraph:assessing_say_actions}
One major difference between GPT-4V and LLaMA-2-13B is that the former will have a completely different writing style, whereas the latter can learn the style of the navigator during finetuning. For example, the navigators may employ acknowledging terms like ``Alright'' and ``Sure'' that can be learned by LLaMA-2-13B, whereas GPT-4V tends to use ``Understood'' and ``Acknowledged''. Beyond those superficial differences, we notice some patterns of failure in \autoref{tab:qualitative_comparison_gpt4_llama13b_say_intent}. First, GPT-4V might come up with unhelpful replies, such as incorrectly sharing a link to the current page when requested to share a link to a different page. In the same scenario, it might simply refuse to assist the instructor, even when the action is achievable. Finally, GPT-4V might generate an utterance that semantically differs from the reference utterance, but would be pragmatically correct. We show one example where, given a request to write an email that includes a discount, the human navigator would ask who should be the recipient, whereas GPT-4V might ask about the details of the discount; clearly, both are valid follow-up questions, but it is challenging to evaluate with existing methods. In all the aforementioned cases (except for the last one), LLaMA-2-13B will provide a short but correct response. Although it may seem less verbose, we found that they are in reality almost as verbose as GPT-4V; the models respectively have, on average, 58.29 (n=1194) and 60.41 characters (n=220) when predicting a \texttt{say} intent on the validation and in-domain test sets.

\subsection{Comparison with human performance}
\label{sec:comparison_with_human_performance}

To understand how well a model would compare to a human annotator at selecting a plausible trajectory, we recruited 3 annotators to predict the best action to take at a given turn in a subset of the demonstrations in the validation set. Then, we compute the agreement score for a given turn as:

\begin{equation*}
    \text{Agreement}_{M}(a_p, \mathcal{A}) = \max_{a_r \in \mathcal{A} \setminus a_p} M(a_p, a_r)
\end{equation*}

where $M$ is the selected metric for a given turn (see \Cref{sec:metrics}),  $a_r$ is the reference annotation and $a_p$ is the prediction by the annotator we are evaluating. where $\mathcal{A}$ is the set of annotations, including the 3 alternative actions selected by the annotators and the original trajectory. To get the same result for the model, we simply compute the simplified version of the equation above (replacing $a_p$ with the model prediction $\hat{a}$):

\begin{equation*}
    \text{Agreement}_{M}(\hat{a}, \mathcal{A}) = \max_{a_r \in \mathcal{A}} M(\hat{a}, a_r)
\end{equation*}

In total, we collected 402 annotations across 134 turns from the validation set. Using those annotations, we compared the reference and model predicted actions with the closest alternative annotations, using our proposed metrics. As shown in the \Cref{tab:comparison_with_human_performance}, LLaMA-2-13B only achieves 65\% of the overall score achieved by the original human navigator, whereas zero-shot GPT-4V achieves 31\%; this reflects the major gap we found in \autoref{tab:test_ood_agg_results}. Moreover, this was performed on a subset of the validation split, so the result for each of the test splits may differ. However, we estimate that annotating the entire test splits would take the 3 annotators around 10 months, without counting the logistics involved with designing efficient annotation tools. Thus, we believe this would be a valuable contribution as part of a follow-up work.

\begin{table}[H]
    \small
    \centering
    \begin{tabular}{lrrrrr}
    \toprule
    {} &  Intent &  Text Group (F1) &  Element Group (IoU) &  Overall &  Overall (Norm) \\
    \midrule
    Annotator Mean &  92.79 &           36.20 &               58.40 &   46.62 &                96.97 \\
    Annotator 1    &  87.31 &           33.21 &               56.82 &   44.84 &                93.27 \\
    Annotator 2    &  94.78 &           39.05 &               56.33 &   47.56 &                98.93 \\
    Annotator 3    &  96.27 &           35.72 &               58.76 &   47.41 &                98.62 \\
    Original       &  95.52 &           40.19 &               56.20 &   48.07 &                100.00 \\
    \midrule
    Llama-2-13B    &  91.04 &           34.37 &               28.44 &   31.45 &                65.42 \\
    GPT-4V         &  54.48 &           09.57 &               20.49 &   14.95 &                31.09 \\
    GPT-4T         &  58.21 &           12.47 &               21.46 &   16.90 &                35.15 \\
    Fuyu-8B        &  84.33 &           25.07 &               27.44 &   26.24 &                54.58 \\
    \bottomrule
    \end{tabular}
    \caption{Agreement scores of models and annotators with respect to the original and alternative trajectories (see \Cref{sec:comparison_with_human_performance}). The results are computed on a subset of the validation set, totalling 402 annotations across 134 turns.}
    \label{tab:comparison_with_human_performance}
\end{table}

\begin{table}[H]
    \small
    \centering
    \begin{tabular}{llrrrr}
    \toprule
    Prompt & Model & Overall & IM & Element Group (IoU) & Text Group (F1) \\
    \midrule
    0S & GPT-3.5T & 10.87 & 42.18 & 11.13 & 3.59 \\
     & GPT-4T & 12.87 & 42.18 & 13.06 & 7.24 \\
     & GPT-4V & 13.52 & 41.76 & 13.96 & 6.69 \\
    \midrule
    D\&E & GPT-3.5T & 8.50 & 40.36 & 8.87 & 4.00 \\
     & GPT-4V & 13.04 & 45.99 & 13.96 & 7.76 \\
     & GPT-4V (no screenshot) & 12.33 & 44.91 & 12.80 & 7.36 \\
    \bottomrule
    \end{tabular}
    \caption{Comparison on the \testid{} split of zero-shot prompt (0S) with prompts that include a description and example (D\&E) for each action. We do not observe substantial differences.}
    \label{tab:gpt_in_context_examples}
\end{table}

\subsection{Augmenting non-finetuned models with in-context examples}

In most of the experiments, we use the same system prompts (\Cref{appendix:input_templates}) to ensure a consistent comparison between models. We specifically chose prompts to ensure that we do not go over the common token limit of 2048 tokens. However, we also consider that providing a description of the actions alongside a concrete example taken from the training set could improve the performance, thus we include a variant prompt template that includes description and example (D\&E) for each action, allowing the model to decide what is the best action to take based on a few examples. To ensure that the examples fit, we truncate parts of the examples. The results can be found in \Cref{tab:gpt_in_context_examples}.

The template can be found below:

\begin{spacing}{0.85}
\begin{externaldoc}
\begin{Verbatim}[breaklines, fontsize=\scriptsize]
{html}
Above are the pruned HTML contents of the page.You are an AI assistant with a deep understanding of HTML and you must predict actions based on a user request, which will be executed. Use one of the following, replacing [] with an appropriate value: 
change(value=[str], uid=[str]) - Whether to change the value of an element, such as inside a dropdown; 
click(uid=[str]) - Clicking on an element using the mouse; load(url=[str]) - Open a webpage in the browser given a URL; say(speaker="navigator", utterance=[str]) - Reply to the instructor using an external chat interface; 
scroll(x=[int], y=[int]) - Use the scroll wheel to navigate vertically (y) or horizontally (x); 
submit(uid=[str]) - Submitting of an element, such as submitting a form to a form-handler;
text_input(text=[str], uid=[str]) - Inserting some text inside an element that can receive text input;
Below are some examples of each action type (separated with ---):
----------
HTML:
(html(body dir class="..."(div(div class="..."(div class="..."(div class="..." tabindex(div class="..."(div ...))
Utterances:
[-00:58] Hello ... [03:06] Request for Recognition - Successful Project Completion ;
Top candidates:
(uid = 185e6683-ebcb-4d73) [[tag]] div [[xpath]] /html/.../div[1] [[bbox]] x=230 y=210 width=767 height=370 [[attributes]] data-event-id='18'... [[children]] div div
...
(uid = 80de1898-76ef-412f) [[tag]] button [[xpath]] ... [[text]] ... [[bbox]] ... [[attributes]] ...
Past Actions:
paste(text="Request for Recognition...") ... click(uid="453b661b-ef85-4402")
Target:
text_input(text="Dear Tim Cook,\n\n\nExciting news! I've completed project", uid="185e6683-ebcb-4d73")
----------
Utterances:
[-00:58] Hello ... [00:24] Send it to esytest5@yahoo.com. ;
Top candidates:
(uid = 7395be17-f2d0-4ce3) [[tag]] input [[xpath]] /html/.../input [[bbox]] x=230 ... height=40 [[attributes]] value='' ...
...
(uid = 80de1898-76ef-412f) [[tag]] button...
Past Actions:
click(uid="d0606930-aac1-4f7c") ... text_input(...) click(uid="7395be17-f2d0-4ce3")
Target:
say(speaker="navigator", utterance="Can you provide me with the details of the accomplishment?")
----------
HTML:
(html(body dir ... aria-label(div aria-label="..." class="..."(a role rel="..." aria-label class="..." ...))
Utterances:
[-00:58] Hello [-00:16] Can you compose an email on Yahoo to request recognition for a Successful Project Completion? ;
Top candidates:
(uid = 119f55af-d3c0-4c15) [[tag]] a [[xpath]] /html/.../a [[text]] Compose [[bbox]] x=16 ... height=36 [[attributes]] tabindex='20' ...
...
(uid = 80de1898-76ef-412f) [[tag]] button...
Past Actions:
click(uid="d0606930-aac1-4f7c") paste(...) ... text_input(text="", uid="d0606930-aac1-4f7c")
Target:
click(uid="7395be17-f2d0-4ce3")
----------
HTML:
(html(body class(div class (div class (div (div class(div class style(div class(div class role data-webtasks-id(...))
Utterances:
[-00:43] Hi [-00:08] Please open KAYAK website and login with google. [00:55] Sure, please find below: ...
Top candidates:
(uid = 4563c30d-ebe2-48fc) [[tag]] div [[xpath]] /html/.../div [[bbox]] x=390.1 ... height=305.6 [[attributes]] tabindex='-1' ...
...
(uid = 1b981d28-023a-4a6f) [[tag]] div...
Past Actions:
tabcreate(target=1482537091) ... tabswitch(origin=1482537091, target=1482537067)
Target:
load(url="https://www.kayak.co.in/flights")
----------
HTML:
(html(body class="not...late" style="overflow...;"(div(div class="ae"(div class="aj... am"(div class=""(div ...))
Utterances:
[-00:24] Hello [-00:16] Please open UberEATS. [-00:08] ... [03:06] Tell me about some items from the Ice Cream section. ;
Top candidates:
(uid = e69a70be-b695-443c) [[tag]] div [[xpath]] /html/.../div [[bbox]] x=166.5 ... height=578 [[attributes]] class='al a...443c' [[children]] div div
...
(uid = 3d33eeed-440d-42e9) [[tag]] span...
Past Actions:
click(uid="0b39661e-11ae-40ca")  ... copy(text="Talenti Gelato Layers Vanilla Fudge Cookie 10.6oz", timestamp="03:51")
Target:
scroll(x=0, y=-200)
----------
HTML:
(html(body class="show-...sticked" (div class="feed-layout" (div class="feed-header" style="position: ...))
Utterances:
[00:04] Hello [00:09] Open website fandom. ;
Top candidates:
(uid = 06bfbae6-1a0e-433b) [[tag]] input [[xpath]] /html/.../input [[bbox]] x=406 ... height=32 [[attributes]] required=''...
...
(uid = 87fc2e27-e4f8-498c) [[tag]] a...
Past Actions:
click(uid="f8223148-2066-4544") text_input(text="Emilia Clarke", uid="06bfbae6-1a0e-433b") click(uid="6af45bdf-41da-4e0f")
Target:
submit(uid="6bfa288c-555a-4bd5")
----------

The user's first and last 4 utterances are: {utterances};
Viewport size: {height}h x {width}w ;
Only the last 5 turns are provided.
Here are the top candidates for this turn: {top_10_candidates}
\end{Verbatim}
\end{externaldoc}
\end{spacing}

\newpage
\section{Additional Result Tables}
\label{appendix:additional_results}

To complement \autoref{sec:results}, we include the scores for each split: in-domain (\S\ref{tab:results_test_id_all}), out-of-domain mean (\S\ref{tab:results_test_ood_all}), \testcat{} (\S\ref{tab:results_test_cat_all}), \testgeo{} (\S\ref{tab:results_test_geo_all}), \testvis{} (\S\ref{tab:results_test_vis_all}), and \testweb{} (\S\ref{tab:results_test_web_all}). We report the intent match (IM) to identify which models fail due to their inability to predict the correct intent. We also include the grouped results in tables \Cref{tab:results_test_id_avg_grouped,tab:results_test_cat_geo_grouped,tab:results_test_vis_web_grouped}.

\begin{table}[H]
  \footnotesize
  \centering

  \caption{Full in-domain test results. We abbreviate \texttt{submit} to \texttt{sbmt} and \texttt{textinput} to \texttt{input}. The first section contains zero-shot results and the second contains finetuned results.}\label{tab:results_test_id_all}
  \begin{adjustbox}{max width=\textwidth}
\begin{tabular}{lrrrrrr|rrrrrr}
        \toprule
                     & \texttt{click} & \texttt{sbmt} & \texttt{input} & \texttt{say} & \texttt{input} & \texttt{load} & \texttt{click} & \texttt{load} & \texttt{say} & \texttt{sbmt} & \texttt{input} \\
                     & IoU            & IoU           & IoU            & chrF         & chrF           & F1            & IM             & IM            & IM           & IM            & IM             \\
        \midrule

        Llama-2-7B   & 6.19           & 5.83          & 4.97           & 4.33         & 4.57           & 29.47         & 43.23          & 36.67         & 32.17        & 6.90          & 10.50          \\
        Llama-2-13B  & 9.42           & 0.00          & 4.97           & 1.25         & 4.82           & 20.57         & 75.65          & 23.33         & 14.93        & 0.00          & 8.84           \\
        GPT-3.5T     & 16.90          & 9.62          & 21.68          & 1.78         & 16.81          & 18.90         & 73.27          & 23.33         & 8.79         & 13.79         & 40.33          \\
        GPT-4T       & 15.92          & 3.45          & 41.33          & 4.53         & 37.50          & 18.90         & 59.61          & 30.00         & 18.24        & 3.45          & 75.14          \\
        GPT-4V       & 17.36          & 6.90          & 46.64          & 4.20         & 35.05          & 15.57         & 63.03          & 16.67         & 14.76        & 6.90          & 71.27          \\
        \midrule
        MindAct-250M & 25.47          & 0.00          & 0.00           & 14.54        & 0.00           & 0.00          & 92.15          & 0.00          & 100.00       & 0.00          & 0.00           \\
\bottomrule
    \end{tabular}
\end{adjustbox}

  \vspace{\baselineskip}

  \footnotesize
  \centering
  \caption{Out-of-domain test results (average). We abbreviate \texttt{submit} to \texttt{sbmt} and \texttt{textinput} to \texttt{input}. The first section contains zero-shot results and the second contains finetuned results.}\label{tab:results_test_ood_all}
  \begin{adjustbox}{max width=\textwidth}
    \begin{tabular}{lrrrrrr|rrrrr}
        \toprule
    & \texttt{click} & \texttt{sbmt} & \texttt{input} & \texttt{say} & \texttt{input} & \texttt{load} & \texttt{click} & \texttt{load} & \texttt{say} & \texttt{sbmt} & \texttt{input} \\
    & IoU   & IoU  & IoU   & chrF & chrF  & F1   & IM    & IM   & IM  & IM   & IM             \\
        \midrule
        Llama-2-7B& 4.84 & 3.01 & 1.76 & 4.23 & 0.86 & 16.05 & 43.05 & 19.24 & 35.53 & 7.62 & 5.38 \\
        Llama-2-13B & 8.89 & 0.50 & 1.51 & 1.46 & 1.06 & 13.56 & 75.93 & 16.08 & 15.32 & 1.64 & 4.46 \\
        GPT-3.5T & 13.46 & 2.77 & 19.52 & 1.44 & 15.27 & 21.26 & 73.82 & 24.74 & 9.10 & 5.51 & 31.75 \\
        GPT-4T & 13.05 & 2.32 & 43.56 & 4.37 & 33.86 & 23.81 & 60.14 & 29.15 & 17.27 & 4.89 & 66.98 \\
        GPT-4V & 13.57 & 2.26 & 42.99 & 3.13 & 33.08 & 18.87 & 64.47 & 22.13 & 13.25 & 4.67 & 64.29 \\
        \midrule
        MindAct-250M & 16.87 & 0.00 & 0.50 & 14.28 & 0.22 & 0.00 & 88.77 & 0.00 & 99.98 & 0.00 & 0.65 \\
\bottomrule
    \end{tabular}
\end{adjustbox}

\end{table}

\begin{table}[H]
  \footnotesize
  \centering

  \caption{Full \testcat{} split (test) results. We abbreviate \texttt{submit} to \texttt{sbmt} and \texttt{textinput} to \texttt{input}. The first section contains zero-shot results and the second contains finetuned results.}\label{tab:results_test_cat_all}
  \begin{adjustbox}{max width=\textwidth}
    \begin{tabular}{lrrrrrr|rrrrr}
        \toprule
                     & \texttt{click} & \texttt{sbmt} & \texttt{input} & \texttt{say} & \texttt{input} & \texttt{load} & \texttt{click} & \texttt{load} & \texttt{say} & \texttt{sbmt} & \texttt{input} \\
                     & IoU            & IoU           & IoU            & chrF         & chrF           & F1            & IM             & IM            & IM           & IM            & IM             \\
        \midrule
        Llama-2-7B   & 5.45           & 7.58          & 0.57           & 4.06         & 0.57           & 17.15         & 47.89          & 22.94         & 39.02        & 15.15         & 2.28           \\
        Llama-2-13B  & 10.55          & 0.00          & 0.85           & 1.48         & 1.01           & 18.97         & 77.63          & 22.02         & 17.62        & 3.03          & 1.71           \\
        GPT-3.5T     & 11.92          & 0.03          & 25.20          & 1.43         & 15.67          & 21.57         & 75.24          & 24.77         & 10.14        & 3.03          & 33.90          \\
        GPT-4T       & 11.23          & 0.00          & 48.03          & 3.08         & 35.26          & 15.61         & 61.01          & 23.85         & 14.97        & 0.00          & 67.52          \\
        GPT-4V       & 11.42          & 1.52          & 46.01          & 2.61         & 33.18          & 14.07         & 65.42          & 17.43         & 12.87        & 1.52          & 65.53          \\
        \midrule
        MindAct-250M & 15.17          & 0.00          & 0.28           & 12.72        & 0.55           & 0.00          & 83.57          & 0.00          & 100.00       & 0.00          & 0.57           \\
\bottomrule
    \end{tabular}
\end{adjustbox}

  \vspace{\baselineskip}

  \caption{Full \testgeo{} split (test) results. We abbreviate \texttt{submit} to \texttt{sbmt} and \texttt{textinput} to \texttt{input}. The first section contains zero-shot results and the second contains finetuned results.}\label{tab:results_test_geo_all}
  \begin{adjustbox}{max width=\textwidth}
    \begin{tabular}{lrrrrrr|rrrrr}
        \toprule
                     & \texttt{click} & \texttt{sbmt} & \texttt{input} & \texttt{say} & \texttt{input} & \texttt{load} & \texttt{click} & \texttt{load} & \texttt{say} & \texttt{sbmt} & \texttt{input} \\
                     & IoU            & IoU           & IoU            & chrF         & chrF           & F1            & IM             & IM            & IM           & IM            & IM             \\
        \midrule
        Llama-2-7B   & 4.21           & 4.00          & 2.54           & 4.45         & 1.58           & 14.04         & 43.35          & 17.11         & 34.58        & 7.00          & 11.61          \\
        Llama-2-13B  & 7.21           & 2.00          & 2.27           & 1.25         & 1.50           & 10.62         & 77.83          & 13.16         & 12.93        & 2.00          & 8.75           \\
        GPT-3.5T     & 14.58          & 4.00          & 14.98          & 1.90         & 15.22          & 19.97         & 73.55          & 25.00         & 10.24        & 5.00          & 33.75          \\
        GPT-4T       & 13.20          & 4.00          & 36.16          & 5.78         & 26.16          & 25.32         & 57.30          & 30.26         & 21.43        & 6.00          & 69.11          \\
        GPT-4V       & 14.56          & 5.00          & 36.09          & 4.07         & 26.50          & 17.86         & 62.16          & 21.05         & 16.00        & 7.00          & 65.00          \\
        \midrule
        MindAct-250M & 16.58          & 0.00          & 0.54           & 18.08        & 0.01           & 0.00          & 86.32          & 0.00          & 100.00       & 0.00          & 0.71           \\
\bottomrule
    \end{tabular}
\end{adjustbox}

\end{table}

\begin{table}[H]
  \footnotesize
  \centering

  \caption{Full \testvis{} split (test) results. We abbreviate \texttt{submit} to \texttt{sbmt} and \texttt{textinput} to \texttt{input}. The first section contains zero-shot results and the second contains finetuned results.}\label{tab:results_test_vis_all}
  \begin{adjustbox}{max width=\textwidth}
    \begin{tabular}{lrrrrrr|rrrrr}
        \toprule
                     & \texttt{click} & \texttt{sbmt} & \texttt{input} & \texttt{say} & \texttt{input} & \texttt{load} & \texttt{click} & \texttt{load} & \texttt{say} & \texttt{sbmt} & \texttt{input} \\
                     & IoU            & IoU           & IoU            & chrF         & chrF           & F1            & IM             & IM            & IM           & IM            & IM             \\
        \midrule
        Llama-2-7B   & 4.35           & 0.01          & 1.16           & 4.15         & 0.87           & 10.61         & 38.04          & 12.86         & 33.41        & 3.05          & 3.85           \\
        Llama-2-13B  & 8.82           & 0.00          & 1.15           & 1.75         & 0.97           & 8.19          & 74.76          & 11.43         & 14.70        & 1.53          & 4.33           \\
        GPT-3.5T     & 14.93          & 5.34          & 16.44          & 0.53         & 14.97          & 10.48         & 74.03          & 15.00         & 4.60         & 6.11          & 27.77          \\
        GPT-4T       & 15.04          & 2.31          & 44.22          & 2.96         & 37.58          & 16.15         & 60.67          & 20.71         & 13.84        & 3.05          & 67.42          \\
        GPT-4V       & 15.26          & 0.82          & 44.73          & 1.66         & 36.59          & 13.10         & 65.03          & 17.14         & 8.90         & 2.29          & 62.76          \\
        \midrule
        MindAct-250M & 18.94          & 0.00          & 0.16           & 11.65        & 0.08           & 0.00          & 92.28          & 0.00          & 100.00       & 0.00          & 0.32           \\
\bottomrule
    \end{tabular}
\end{adjustbox}

  \vspace{\baselineskip}

  \caption{\small{Full \testweb{} split (test) results. We abbreviate \texttt{submit} to \texttt{sbmt} and \texttt{textinput} to \texttt{input}. The first section contains zero-shot results and the second contains finetuned results.}}\label{tab:results_test_web_all}
  \begin{adjustbox}{max width=\textwidth}
    \begin{tabular}{lrrrrrr|rrrrr}
        \toprule
                     & \texttt{click} & \texttt{sbmt} & \texttt{input} & \texttt{say} & \texttt{input} & \texttt{load} & \texttt{click} & \texttt{load} & \texttt{say} & \texttt{sbmt} & \texttt{input} \\
                     & IoU            & IoU           & IoU            & chrF         & chrF           & F1            & IM             & IM            & IM           & IM            & IM             \\
        \midrule
        Llama-2-7B   & 5.33           & 0.47          & 2.78           & 4.27         & 0.41           & 22.38         & 42.92          & 24.05         & 35.12        & 5.26          & 3.79           \\
        Llama-2-13B  & 8.98           & 0.00          & 1.77           & 1.36         & 0.76           & 16.47         & 73.48          & 17.72         & 16.04        & 0.00          & 3.03           \\
        GPT-3.5T     & 12.41          & 1.71          & 21.46          & 1.88         & 15.21          & 33.00         & 72.46          & 34.18         & 11.43        & 7.89          & 31.57          \\
        GPT-4T       & 12.75          & 2.97          & 45.83          & 5.64         & 36.43          & 38.15         & 61.56          & 41.77         & 18.83        & 10.53         & 63.89          \\
        GPT-4V       & 13.03          & 1.71          & 45.12          & 4.19         & 36.06          & 30.47         & 65.26          & 32.91         & 15.21        & 7.89          & 63.89          \\
        \midrule
        MindAct-250M & 16.79          & 0.00          & 1.01           & 14.68        & 0.26           & 0.00          & 92.90          & 0.00          & 99.92        & 0.00          & 1.01           \\
\bottomrule
    \end{tabular}
\end{adjustbox}

\end{table}

\begin{table}[H]
  \centering

  \caption{Element Group (EG), Text Group (TG) and overall results for \testid{} (left) and \testod{} (right) splits. The top section contains zero-shot results and the bottom contains finetuned results.}\label{tab:results_test_id_avg_grouped}
  \begin{adjustbox}{max width=\textwidth}
    \begin{tabular}{lrrrr|rrrr}
        \toprule
                     & Overall   & Overall & EG    & TG    & Overall   & Overall & EG    & TG    \\
                     & Micro Avg & IM      & IoU   & F1    & Micro Avg & IM      & IoU   & F1    \\
        \midrule
        Llama-2-7B   & 5.32      & 33.80   & 4.01  & 3.06  & 4.04 & 33.96 & 2.92 & 2.14 \\
        Llama-2-13B  & 5.61      & 42.85   & 5.29  & 1.97  & 5.16 & 43.68 & 4.80 & 1.31 \\
        GPT-3.5T     & 10.35     & 42.42   & 10.68 & 3.98  & 8.51 & 42.77 & 8.62 & 3.45 \\
        GPT-4T       & 12.24     & 42.69   & 12.55 & 7.85  & 10.72 & 41.66 & 10.85 & 6.75 \\
        GPT-4V       & 12.99     & 42.47   & 13.68 & 7.28  & 10.45 & 42.36 & 10.91 & 6.21 \\
        \midrule
        MindAct-250M & 16.88     & 76.54   & 18.01 & 8.46  & 12.63 & 74.25 & 12.05 & 7.67 \\
\bottomrule
    \end{tabular}

\end{adjustbox}

  \vspace{\baselineskip}

  \centering
  \caption{Element Group (EG), Text Group (TG) and overall results for \testcat{} (left) and \testgeo{} (right) splits. The top section contains zero-shot results and the bottom contains finetuned results.}\label{tab:results_test_cat_geo_grouped}
  \begin{adjustbox}{max width=\textwidth}
    \begin{tabular}{lrrrr|rrrr}
        \toprule
                     & Overall   & Overall & EG    & TG    & Overall   & Overall & EG    & TG    \\
                     & Micro Avg & IM      & IoU   & F1    & Micro Avg & IM      & IoU   & F1    \\
        \midrule
        Llama-2-7B   & 4.57      & 38.32   & 3.46  & 2.18  & 3.61      & 33.48   & 2.60  & 2.11  \\
        Llama-2-13B  & 6.50      & 47.52   & 6.03  & 1.59  & 4.03      & 43.04   & 3.87  & 1.09  \\
        GPT-3.5T     & 8.23      & 45.91   & 8.42  & 3.35  & 8.86      & 42.09   & 8.78  & 3.66  \\
        GPT-4T       & 9.48      & 42.14   & 9.90  & 5.63  & 10.61     & 40.86   & 10.53 & 6.38  \\
        GPT-4V       & 9.26      & 43.66   & 9.80  & 5.30  & 10.74     & 41.33   & 11.05 & 5.86  \\
        \midrule
        MindAct-250M & 11.69     & 72.93   & 11.27 & 6.61  & 13.15     & 70.25   & 11.20 & 8.93  \\
\bottomrule
    \end{tabular}
\end{adjustbox}

\end{table}

\begin{table}[H]
  \centering
  \caption{Element Group (EG), Text Group (TG) and overall results for \testvis{} (left) and \testweb{} (right) splits. The top section contains zero-shot results and bottom contains finetuned results.}\label{tab:results_test_vis_web_grouped}
  \begin{adjustbox}{max width=\textwidth}
    \begin{tabular}{lrrrr|rrrr}
        \toprule
                     & Overall   & Overall & EG    & TG    & Overall   & Overall & EG    & TG    \\
                     & Micro Avg & IM      & IoU   & F1    & Micro Avg & IM      & IoU   & F1    \\
        \midrule
        Llama-2-7B   & 3.77      & 30.99   & 2.50  & 2.10  & 4.35      & 33.03   & 3.26  & 2.19  \\
        Llama-2-13B  & 5.06      & 42.05   & 4.60  & 1.34  & 5.21      & 42.11   & 4.86  & 1.23  \\
        GPT-3.5T     & 8.58      & 40.14   & 8.84  & 2.63  & 8.42      & 42.95   & 8.51  & 4.16  \\
        GPT-4T       & 11.14     & 40.38   & 11.65 & 6.36  & 11.77     & 43.28   & 11.43 & 8.62  \\
        GPT-4V       & 10.73     & 40.45   & 11.59 & 5.72  & 11.20     & 44.00   & 11.35 & 7.96  \\
        \midrule
        MindAct-250M & 13.16     & 79.07   & 13.97 & 7.26  & 12.54     & 74.76   & 11.75 & 7.89  \\
\bottomrule
    \end{tabular}
\end{adjustbox}

\end{table}

\newpage
\section{Instructions for the Annotators}
\vspace{0.5em}
\label{app:instructions}
\begin{spacing}{0.75}
\begin{externaldoc}
  \section*{Project Information}
  We are collecting data for \textbf{evaluating} automated web navigation
  systems. The data consists of \textbf{demonstrations} of interactions
  between the user and the navigator.

  In each demonstration, the user and the system cooperate to achieve
  \textbf{tasks in a web browser.} The user controls the system via
  \textbf{natural language instructions}.

  \hypertarget{how-to}{%
    \vspace{-1\baselineskip}
    \section*{How To}\label{how-to}}

  \hypertarget{ingredients}{%
    \subsection*{Ingredients}\label{ingredients}}

  \begin{compactitemize}
    \item
          \textbf{two people:}
          \begin{compactitemize}
            \item
                  \textbf{Instructor:} creative, giving instructions
            \item
                  \textbf{Navigator:} systematic, following instructions
          \end{compactitemize}
    \item
          Google Chrome
    \item
          Zoom
    \item
          internet connection
  \end{compactitemize}
  \vspace{-1\baselineskip}
  \hypertarget{preparation}{%
    \subsection*{Preparation}\label{preparation}}

  You need to do this process just once:

  \begin{enumerate}
    \def\labelenumi{\arabic{enumi}.}
    \item
          Download the
          \href{https://drive.google.com}{{Chrome
                extension ZIP file}} and unpack the \emph{extension} folder to your
          local filesystem.
    \item
          If you are using Chrome as your primary browser,
          \href{https://support.google.com/chrome/answer/2364824}{{create a new
                profile}} for the experiments.
    \item
          Install the Chrome extension in the repository:
          \begin{itemize}
            \item
                  Open a new Google Chrome window.
            \item
                  Go to chrome://extensions/
            \item
                  At the top right, turn on Developer mode.
            \item
                  Click Load unpacked.
            \item
                  Find and select the \emph{extension} folder you have unpacked before
                  (make sure you are inside the folder).
            \item
                  Click on the ``puzzle'' icon in the task bar with Chrome extensions
                  and pin this extension.
          \end{itemize}
    \item
          Setup Zoom:
          \begin{itemize}
            \item
                  Open Zoom and \textbf{log in}.
            \item
                  Go to
                  \href{https://zoom.us/profile/setting}{{https://zoom.us/profile/setting}}
            \item
                  On the Meeting tab, turn on \emph{Auto saving chats
                    (\href{https://support.zoom.us/hc/en-us/articles/360060889932-Enabling-meeting-and-webinar-auto-saving-chats}{{learn
                          more here}}).}
            \item
                  On the Recording tab:

                  \begin{enumerate}
                    \def\labelenumii{\roman{enumii}.}
                    \item
                          enable Local Recording
                    \item
                          enable \emph{``Hosts can give meeting participants permission to
                            record locally''}.
                    \item
                          enable automatic recording on a local computer
                  \end{enumerate}
            \item
                  \textbf{Setup your Zoom name to \emph{Instructor} or
                    \emph{Navigator} according to your role.}
          \end{itemize}
  \end{enumerate}
  
  \vspace{-1\baselineskip}
  \hypertarget{updating-the-extension}{%
    \subsection*{Updating the extension}\label{updating-the-extension}}

  Check regularly if you are using an up-to-date version of the extesion:

  \begin{itemize}
    \item
          The current version can be found at the top of this document.
    \item
          Your version is at chrome://extensions/ next to the extension name.
  \end{itemize}

  If there is a never version of the extension, remove the extension and
  repeat points 1) and 3) in the Preparation section.

  \hypertarget{demonstrations}{%
    \subsection*{Demonstrations}\label{demonstrations}}

  \begin{enumerate}
    \def\labelenumi{\arabic{enumi}.}
    \item
          \textbf{Navigator} calls \textbf{Instructor} via Zoom (Participants →
          Invite)          \begin{itemize}
            \item
                  Ensure that both have \textbf{video and microphone are disabled}.
          \end{itemize}
    \item
          After the call is accepted:          \begin{itemize}
            \item
                  \textbf{Instructor} opens a Zoom chat window,
            \item
                  \textbf{Navigator}:
                  \begin{itemize}
                    \item
                          opens a Zoom chat window,
                    \item
                          opens a Chrome window,
                    \item
                          shares the screen with their Chrome window \textbf{(only),}
                    \item
                          starts recording a Zoom call video (ignore the warning about
                          audio).
                  \end{itemize}
          \end{itemize}
    \item
          \textbf{Navigator} clicks on the extension button in the navigation
          bar and selects \textbf{New recording}.          \begin{itemize}
            \item
                  A new tab will open with an overlay \emph{Starting recording} for 1
                  second (make sure that it is visible on the Zoom recording),
                  followed by a prompt for waiting for instructions.
            \item
                  \textbf{Use the opened tab, do not open any new tab!}
          \end{itemize}
    \item
          \textbf{Instructor} gives \textbf{Navigator} instructions through the
          chat interface for accomplishing a task (see Tasks for details).
          \begin{itemize}
            \item
                  \textbf{Instructor} has no other way of communicating with
                  \textbf{Navigator} than \textbf{through the chat interface}.
            \item
                  \textbf{Instructor} can give intermediate instructions or answer
                  system questions.
          \end{itemize}
    \item
          \textbf{Navigator} performs actions in the web browser according to
          \textbf{Instructor}'s instructions.
          \begin{itemize}
            \item
                  \textbf{Navigator} should use the \textbf{chat interface} to ask the
                  user for any missing details and to provide answers if necessary.
          \end{itemize}
    \item
          After the task is finished, \textbf{Navigator}:
          \begin{itemize}
            \item
                  clicks on the extension button, selects \emph{Save recording} and
                  \textbf{wait until the recording gets saved to their computer},
            \item
                  stops the video recording and screen sharing,
            \item
                  ends the call,
            \item
                  submits the recording (see Recording for details).
          \end{itemize}
  \end{enumerate}
  
  \vspace{-1.5\baselineskip}
  \hypertarget{recording}{%
    \subsection*{Recording}\label{recording}}

  \textbf{The recording is submitted through the
    \href{https://example.com}{{web interface}}}.

  The recording consists of:
  \vspace{-1\baselineskip}
  \begin{itemize}
    \item
          a ``\texttt{\textless recording\_id\textgreater.zip}'' file, which is a
          ZIP archive with:
          \begin{itemize}
            \item
                  metadata,
            \item
                  events,
            \item
                  screenshots,
            \item
                  HTML snapshots,
          \end{itemize}
    \item
          Zoom chat history ``\texttt{meeting\_saved\_chat.txt}'',
    \item
          Zoom invite link
  \end{itemize}

  The Zoom recording folder
  \href{https://support.zoom.us/hc/en-us/articles/206277393-Finding-and-viewing-local-recordings}{{depends
        on your platform}}. The default directories are:
  \vspace{-1\baselineskip}
  \begin{itemize}
    \item
          \textbf{Windows}:
          \texttt{C:\textbackslash Users\textbackslash{[}Username{]}\textbackslash Documents\textbackslash Zoom}
    \item
          \textbf{Linux}: \texttt{/home/{[}Username{]}/Documents/Zoom}
    \item
          \textbf{Mac}: \texttt{/Users/{[}Username{]}/Documents/Zoom}
  \end{itemize}

  \hypertarget{actions}{%
    \subsection*{Actions}\label{actions}}

  \textbf{Navigator} can perform the following actions in the browser:
  \vspace{-1\baselineskip}

  \begin{itemize}
    \item
          \textbf{go to a URL} through the navigation bar,
    \item
          \textbf{click} on an element,
    \item
          \textbf{input text} into an input field,
    \item
          \textbf{scroll} up and down the page,
  \end{itemize}

  The actions which should \textbf{not be performed}:
  \vspace{-1\baselineskip}

  \begin{itemize}
    \item
          opening a new tab (it is ok if the page opens a tab by itself),
    \item
          horizontal scrolling,
    \item
          page search (Ctrl+F),
    \item
          keyboard shortcuts,
    \item
          drag \& drop (e.g. Google Maps)
  \end{itemize}
  
  \vspace{-1.5\baselineskip}
  \hypertarget{tasks}{%
    \subsection*{Tasks}\label{tasks}}

  \textbf{Instructor} can give the system any tasks which an automated web
  assistant should be able to handle. Use your imagination!

  The tasks \textbf{can be unspecified at first}. It is the job of the
  system to ask for intermediate details throughout the tasks
  demonstration.

  \textbf{Stop the demonstration} before doing any real action in the
  world: booking a table, buying a ticket, etc.
  
  \vspace{-0.5\baselineskip}
  \hypertarget{websites}{%
    \subsection*{Websites}\label{websites}}

  For your inspiration, here is a spreadsheet
  with the \textbf{list of websites} and the task categories you can use
  them for.

  \textbf{We have created a
    shared
    account for these websites which you should use in case you need to
    login.}

  Of course feel free to use any other websites (just do not fill in any
  other personal details there, preferably use the shared account as
  well).

  \vspace{-0.5\baselineskip}
  \hypertarget{tips}{%
    \subsection*{Tips}\label{tips}}

  \textbf{Navigator}
  \vspace{-1\baselineskip}

  \begin{itemize}
    \item
          \textbf{Don't do things too quickly!} Saving the actions, screenshots
          and pages takes time and performing the actions in a quick succession
          can introduce errors in the recording, especially on heavy websites.\\
          Watch for the icon indicating that the browser is processing an
          action.
    \item
          Do not perform any \textbf{unnecessary actions} \emph{(all the actions
            will be recorded and we want to minimize the amount of mindless
            clicking and scrolling)}
    \item
          Wait until the page \textbf{fully loads}.
    \item
          \textbf{Do not use autofill} for text fields, always type everything
          from scratch.
    \item
          Do not change the \textbf{size of the browser window} if not
          necessary.
  \end{itemize}

  \vspace{-1\baselineskip}
  \textbf{Instructor}
  \vspace{-1\baselineskip}
  
  \begin{itemize}
    \item
          \textbf{Be creative}: assign tasks starting from very simple (``submit
          the form'') to very complex (multi-turn conversation with changing
          topics).
    \item
          Ask only about things that are relevant to the webpage.
    \item
          \textbf{Wait} until the system performs their actions.
          \begin{itemize}
            \item
                  However, feel free to interrupt if something does not seem right or
                  you have changed your mind.
          \end{itemize}
    \item
          Finalize all the tasks right \textbf{before changing the actual state
            of the worl}d (i.e. ordering products, submitting issues etc.).
  \end{itemize}

  Note that the extension does not work in an anonymous window. If you
  want to clear your history afterwards, use Ctrl+Shift+Delete.
\end{externaldoc}
\end{spacing}

\newpage

\end{document}